\newcommand\vartitle{Building Socio-Affective Artificial Intelligence for Interactive Multi-Agent Simulations}
\newcommand\varalias{AGIMUD }
\newcommand\varexpalias{Agentic General Intelligence and Multi-User Dungeons}

\documentclass[journal]{IEEEtran}
\usepackage[utf8]{inputenc}
\usepackage{graphicx}
\usepackage{subcaption}
\usepackage{mathtools}
\usepackage{pifont}
\usepackage{amsmath}
\usepackage{amssymb}
\usepackage{fontawesome}
\usepackage{caption}
\usepackage{adjustbox}
\usepackage{hyperref}
\usepackage{changepage}
\usepackage{booktabs}
\usepackage{fancyvrb} 
\usepackage[most]{tcolorbox} 
\usepackage{fvextra} 

\usepackage{multirow}
\usepackage{makecell}
\usepackage{authblk}
\usepackage{cite}
\usepackage[table]{xcolor}
\usepackage{listings} 
\usepackage{makecell}
\usepackage{color}
\usepackage{amsfonts}
\usepackage{titlesec}
\usepackage{float}
\usepackage{varwidth}
\usepackage{verbatimbox}
\usepackage{listings}
\newcommand{\xmark}{\text{\ding{55}}}
\definecolor{darkpurple}{HTML}{36013F} 
\definecolor{codegreen}{rgb}{0,0.6,0}
\definecolor{codegray}{rgb}{0.5,0.5,0.5}
\definecolor{codepurple}{rgb}{0.58,0,0.82}
\definecolor{backcolour}{rgb}{0.95,0.95,0.92}

\newtcolorbox{terminal}[1][]{
  enhanced,
  colback=black!92!white,      
  colframe=black!70!white,
  coltext=green!70!white,      
  fontupper=\ttfamily\footnotesize, 
  boxrule=0.4pt,
  sharp corners,
  left=3pt,right=3pt,top=3pt,bottom=3pt,
  breakable=false,
  #1
}

\begin{document}
\title{\vartitle}
\author[1,*]{David~Berga}
\affil[*]{Corresponding author: david.berga@enti.cat} 
\affil[1]{Escola de Noves Tecnologies Interactives (ENTI), Universitat de Barcelona, Spain}

\IEEEtitleabstractindextext{
\begin{abstract}
The objective of this article is to provide design principles and a software architecture for enabling interaction between humans and multiple agents in simulated dynamic worlds. This connects the current era of general artificial intelligence (AI/AGI) with the proliferation of transformer-based conversational agents and the increased computational capabilities. Given an overview of current and previous multi-agent theories of mind (socially and affectively-aware agents), the existence of an integrative design of agent interactions with themselves and with humans must be crucial for understanding how to create sustainable and governance in future human-agent reasoning systems. In this work is presented a software "\varalias" that integrates: A. socially-aware reasoning and emotion in agent behavior and interaction, B. a design of human multimodal scheme for human users, artificial agents and simulated worlds, and C. distributing the AI processing through the network to enable multiple autonomous agents. These integrations allow the dynamic world recreation as multi-user dungeons (MUDs) where both agents and humans can interact simultaneously in real time. Find the code online in \url{https://github.com/dberga/AGIMUD}
\end{abstract}

}

\maketitle
\IEEEdisplaynontitleabstractindextext
\IEEEpeerreviewmaketitle

\section{Multi-Agent Systems in the Transformers Era}

The taxonomy of chatbot engines illustrates a critical transition from deterministic, intent-based conversational engines to probabilistic, generative Large Language Models (LLMs). To understand the magnitude of this architectural shift, it is essential to contextualize the historical trajectory of Agent-based Models (ABM) and Multi-Agent Systems (MAS). Traditionally, autonomous agents were governed by Good Old-Fashioned AI paradigms, most notably the Belief-Desire-Intention (BDI) model \cite{Rao1995BDI}. In BDI architectures, agents operated based on explicit logical representations of their environment (beliefs), their overarching goals (desires), and the specific plans they committed to executing (intentions). While logically sound, highly explainable, and predictable, BDI systems and subsequent Partially Observable Markov Decision Processes (POMDP) suffered fundamentally from the "curse of dimensionality" and the "state-space explosion" problem \cite{Kaelbling1998PlanningAA}. Hardcoding responses, transition matrices, heuristic evaluations, and branching dialogue trees for complex, multi-agent interactions in dynamic, unpredictable procedural environments became computationally and logically intractable as the simulation scope expanded. 

Early conversational engines like the initial iterations of Dialogflow\cite{dialogflow2016}, Rasa\cite{Bocklisch2017RasaOS,rasaplatform2018}, and IBM Watson\cite{ibmwatson2016} attempted to bridge this gap using legacy Natural Language Understanding (NLU) pipelines \cite{Qiu2020, KOCAMAN2021100058, Liu2025, OShaughnessy2026}. These systems relied heavily on support vector machines \cite{cortes1995support, Wahba2023, Zhu2025} and early recurrent networks \cite{elman1990finding, hochreiter1997long, schuster1997bidirectional, mikolov2010recurrent, sutskever2011generating} to classify user intents and extract hardcoded entities. However, these systems were inherently limited by their rigid taxonomy and lack of semantic elasticity; if a user's input deviated even slightly from the predefined training intents, or if the conversation required maintaining context across multiple conversational turns, the agent would reliably fail, leading to severe immersion breakage in simulated environments \cite{WrightMaley2015,Kim2025,Jrgensen2026}. The advent of the Transformer architecture \cite{vaswani2017attention}, starting with GPT-1\cite{radford2018improving} is evolving rapidly through high-parameter models like GPT-4\cite{gpt42023,Yenduri2024}, Claude 3\cite{claude32024}, and Gemini\cite{gemini2023}, definitively resolved this state-space limitation. By utilizing self-attention mechanisms—specifically query, key, and value matrices—to map natural language inputs directly to continuous probability distributions, LLMs enabled true zero-shot and few-shot generalization\cite{brown2020language, sanh2022multitask}. Agents could now process unprecedented contextual depth, drawing upon vast latent knowledge to generate novel responses without requiring predefined dialogue paths.

\begin{table*}[h!]
    \centering
    \resizebox{18cm}{!}{%
    \begin{tabular}{lllll}
    \hline
    Year & Chatbot Engine & Creator/Developer & Code Language Used & License \\
    \hline
    2018 & Transformer 1 (GPT-1)\cite{radford2018improving} & OpenAI & Python (TensorFlow) & \href{https://github.com/hyunwoongko/transformer}{MIT} \\
    2018 & Rasa\cite{rasaplatform2018} & Rasa & Python & \href{https://github.com/RasaHQ/rasa}{Apache 2.0} \\
    2018 & Dialogflow CX\cite{dialogflowcx2018} & Google & Node.js, Java, Python & \href{https://github.com/googleapis/nodejs-dialogflow-cx}{Apache 2.0} \\
    2019 & BotStar\cite{botstar2019} & BotStar & Node.js & \href{https://www.botstar.com/}{Proprietary} \\
    2019 & Kore.ai\cite{koreai2019} & Kore.ai & Node.js, Java & \href{https://kore.ai/}{Proprietary} \\
    2020 & GPT-3\cite{gpt32020} & OpenAI & Python, Php (API) & \href{https://github.com/evillafuerte/gpt3-php}{MIT (API)} \\
    2021 & GPT-Neo\cite{gptneo2021} & EleutherAI & Python & \href{https://github.com/EleutherAI/gpt-neo}{MIT} \\
    2021 & BlenderBot\cite{blenderbot2021} & Facebook (Meta) & Python (PyTorch) & \href{https://github.com/facebookresearch/ParlAI}{MIT} \\
    2022 & Perplexity AI\cite{perplexity2022} & Perplexity AI & Python, JavaScript & \href{https://github.com/gweidart/pyplexityai}{MIT (API)} \\
    2023 & Microsoft Copilot\cite{mscopilot2023} & Microsoft & C\#, .NET, WebAssembly & \href{https://github.com/microsoft/chat-copilot}{MIT (API)} \\
    2023 & GPT-4\cite{gpt42023} & OpenAI & Python, C++ & \href{https://github.com/tailorvj/gpt4api}{AGPLv3 (API)} \\
    2023 & Grok\cite{grok2023} & xAI (Twitter) & Python, Rust, JAX & \href{https://github.com/xai-org/grok-1}{Apache 2.0} \\
    2023 & Qwen (Tongyi Qianwen)\cite{qwen2023} & Alibaba Cloud & Python & \href{https://github.com/QwenLM/qwen}{Apache 2.0} \\
    2023 & DeepSeek-LLM\cite{deepseekllm2023} & Liang Wenfeng & Python & \href{https://github.com/deepseek-ai/DeepSeek-LLM}{MIT} \\
    2023 & Gemini 1.0 Pro/Nano\cite{gemini2023} & Google & Python, Java, C++, Go & \href{https://github.com/google-gemini/gemini-cli}{Apache 2.0 (API)} \\
    2023 & Claude 1 \& 2\cite{claude2023} & Anthropic & Python & \href{https://github.com/bredmond1019/claude-sdk-rs}{MIT (API)} \\
    2024 & DeepSeek-V2\cite{deepseekv22024} & Liang Wenfeng & Python & \href{https://github.com/deepseek-ai/DeepSeek-V2}{MIT} \\
    2024 & Qwen2.5\cite{qwen252024} & Alibaba Cloud & Python & \href{https://github.com/QwenLM/qwen}{Apache 2.0} \\
    2024 & DeepSeek-V3\cite{deepseekv32024} & Liang Wenfeng & Python & \href{https://github.com/deepseek-ai/DeepSeek-V3}{MIT} \\
    2024 & Gemini 1.5 Pro/Flash\cite{gemini152024} & Google & Python, Java, C++, Go & \href{https://github.com/google-gemini/gemini-cli}{Apache 2.0 (API)} \\
    2024 & Claude 3\cite{claude32024} & Anthropic & Python & \href{https://github.com/anthropics/anthropic-sdk-python}{MIT (SDK)} \\
    2025 & Qwen3\cite{qwen32025} & Alibaba Cloud & Python & \href{https://github.com/QwenLM/qwen}{Apache 2.0} \\
    2025 & DeepSeek-R1\cite{deepseekr12025} & Liang Wenfeng & Python & \href{https://github.com/deepseek-ai/DeepSeek-R1}{MIT} \\
    2025 & Gemini 3 Pro\cite{gemini32025} & Google & Python, JavaScript, C++ & \href{https://github.com/google-gemini/gemini-cli}{Apache 2.0 (API)} \\
    2025 & Claude 4\cite{claude42025} & Anthropic & Python & \href{https://github.com/anthropics/anthropic-sdk-python}{MIT} \\
    2025 & Qwen3.6-Plus\cite{qwen362025} & Alibaba Cloud & Python & \href{https://www.qwencloud.com/models/qwen3.6-plus}{Apache 2.0} \\
    2026 & Elephant Alpha\cite{elephanta2026} & inclusionAI (Ant) & Python & \href{https://github.com/inclusionAI/Ling}{MIT} \\
    2026 & DeepSeek-V4\cite{deepseekv42026} & Liang Wenfeng & Python & \href{https://huggingface.co/collections/deepseek-ai/deepseek-v4}{MIT} \\ 
    2026 & Claude 5\cite{claude52026} & Anthropic & Python & \href{https://github.com/anthropics/anthropic-sdk-python}{MIT} \\
    2026 & Odysseus\cite{odysseus2026} & pewdiepie-archdaemon & Python & \href{https://github.com/pewdiepie-archdaemon/odysseus}{MIT} \\
    2026 & GLM 5.2\cite{glm522026} & z.ai & Python & \href{https://github.com/zai-org/GLM-5}{MIT} \\
    \hline
    \end{tabular}%
    }
    \caption{List of Agentic AI and Chatbot Engines in the Era of Agentic AI (Transformer 1) after 2018}
    \label{tab:chatbots}
\end{table*}

A comprehensive chronological listing is presented in Table~\ref{tab:chatbots} encapsulates the evolutionary trajectory of conversational agents, spanning from the modern era of large-scale generative models to current state of the art models. Prior to this period, the field was initiated by rule-based systems (e.g., Weizenbaum's ELIZA\cite{Weizenbaum1966Eliza} and Winograd's SHRDLU\cite{winograd1971shrdlu}) and retrieval-based or template-driven engines such as AIML-based A.L.I.C.E. \cite{wallace1995alice,wallace2004elements,Wallace2008Anatomy} and Mitsuku \cite{worswickkuki2005,croes2021can}, which relied heavily on pattern matching and curated knowledge bases. These earlier systems, while pioneering, exhibited limited generalisation capabilities and were inherently brittle when confronted with out-of-domain inputs or multi-turn contextual dependencies. Since the publication of the first visual transformer by Vaswani et al. in 2017 \cite{vaswani2017attention}, we witness a proliferation of models grounded in the Transformer architecture, leveraging massive pre-training on diverse corpora to acquire rich semantic representations. This shift is further evidenced by the diversity of programming frameworks employed and the increasing prevalence of permissive open-source models, which have democratised access to state-of-the-art conversational AI and accelerated both academic research and industrial deployment. 

Within this contemporary ecosystem, platforms such as OpenAI's ChatGPT \cite{radford2018improving} occupy a distinctive and strategically important niche. Unlike generative LLMs that operate as black-box autoregressive text synthesizers, Rasa\cite{Bocklisch2017RasaOS,rasaplatform2018} is explicitly architected as a modular framework for building contextual, task-oriented dialogue agents. Its dual-component design, comprising Rasa NLU and Rasa Core, embodies a hybrid approach that combines the strengths of neural language understanding with symbolic dialogue management. Concretely, Rasa NLU is responsible for the extraction of structured information from user utterances—identifying intents, entities, and contextual cues—thereby transforming natural language into a formalised semantic frame that can be reasoned upon. This capability is crucial for grounding the agent's subsequent actions in the operational state of the environment or the user's specific goals. Rasa Core then employs a transformer-based or recurrent dialogue policy, trained via reinforcement learning or supervised learning, to select the most appropriate next action from a predefined set of domain-specific actions. This architecture effectively mitigates the state-space explosion problem that plagued classical BDI and POMDP-based agents, by learning stochastic policies over a continuous representation space. 

\begin{figure}[h!]
    \centering
    \includegraphics[width=1\linewidth,height=8.9cm]{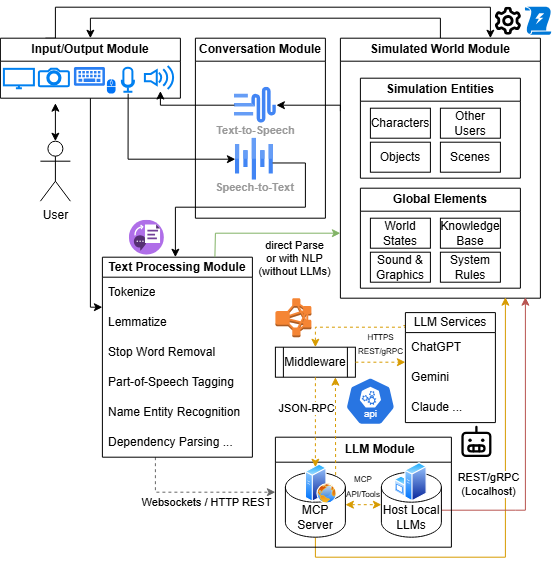}
    \caption{Information Flow in User-AI Conversational Interaction} 
    \vspace{-0.6cm}
    \label{fig:AGIMUD-LLM}
\end{figure}

Nevertheless, replacing deterministic logic with probabilistic generation introduced severe new vulnerabilities into multi-agent simulations: specifically, semantic hallucination \cite{Wang2025, Huang2025, AnhHoang2025, Konishi2025,Alansari2026}, lack of temporal consistency (state persistence/amnesia) \cite{Bajpai2025,Wu2025}, catastrophic forgetting \cite{Delange2021,Haque2025}, and catastrophic reasoning failures in complex, multi-step planning tasks \cite{bubeck2023sparks}. Consequently, the post-2023 proliferation of open-weight models (e.g. LLama\cite{llama2023,dubey2024llama}, Qwen\cite{qwen2023} and DeepSeek\cite{deepseekllm2023}) has democratized the deployment of agentic reasoning, but has fundamentally shifted the research focus of the AI community. Modern MAS research in the Transformer era is no longer focused merely on generating believable text; it is deeply concerned with constraining, grounding, and structuring LLM outputs. This is achieved via rigid JSON schema validation, memory-augmented Retrieval-Augmented Generation using vector databases \cite{Lewis2020RAG}, neuro-symbolic wrappers and other pipelines (such as ReAct\cite{yao2023react}, LLMob\cite{wang2024large}, D2A\cite{simulating2025humanlike}, BabyAGI\cite{nakajima2023babyagi} or AutoGen\cite{wu2024autogen}) designed to reliably drive autonomous, goal-oriented actions within procedural environments, trying to mitigate possible hallucinatory loops.

\section{Theories of Mind for Multi-Agent Systems}

\label{sec:tom_intro}

\begin{figure*}
    \centering
    \includegraphics[width=0.95\linewidth]{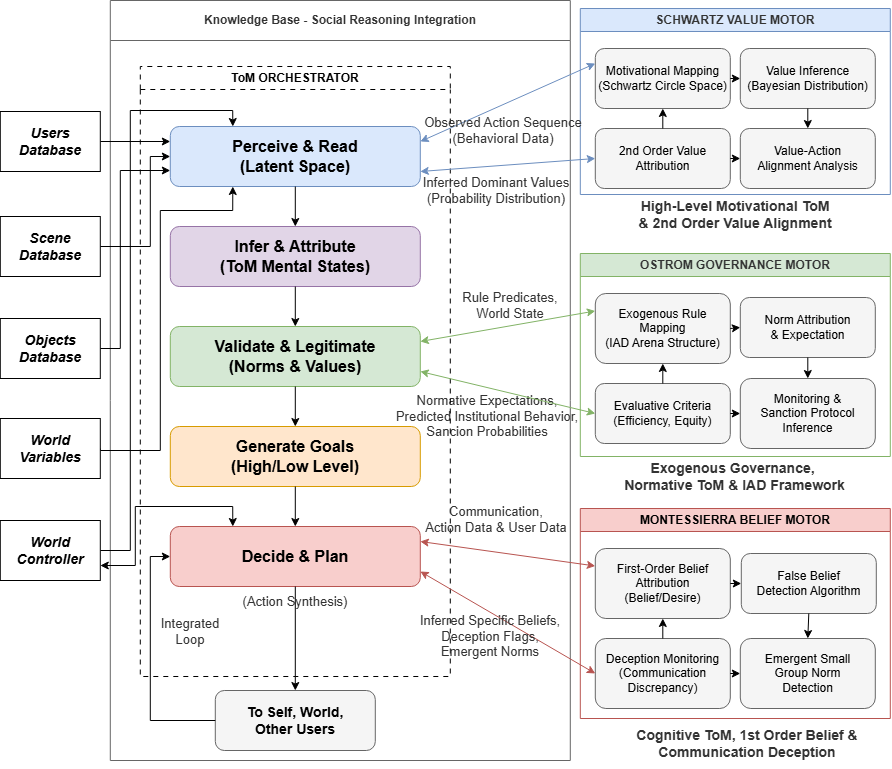}
    \caption{\varalias's Social Reasoning and Behavior Pipeline in the Simulated World Module}
    \label{fig:social_reasoning_pipeline}
\end{figure*}

Current artificial intelligence models (machine and deep learning techniques) have emerged as supervised systems that learn upon calibrating their internal parameters through task repetition and error assessment, and are usually trained with unique parametric factors. These techniques disregard multimodal factors and lack many endogenous factors, present in human brain organization and its unique reinforcement learning dynamics. This means that current learning methods are limited in reasoning and negotiation capacities, and thus unable to replicate social values and human cognitive interaction processes. 

In cognitive science, the capacity to attribute mental states—beliefs, intents, desires, emotions, and knowledge—to oneself and others is known as Theory of Mind (ToM) \cite{Premack1978}. This capacity is the bedrock of human empathy skills, strategic deception, and complex cooperation. While recent empirical studies thoroughly debate whether emergent ToM capabilities organically exist in billion-parameter language models \cite{Ullman2023ToM, Kosinski2024}, it is evident that base LLMs natively lack intrinsic ethical grounding, persistent sociological frameworks, and even a continuous sense of self. To bridge this profound gap in persistent multi-agent environments, developers must formalize externalized ToM mathematically through distinct behavioral, normative, and cooperative frameworks.

\subsection{Schwartz's Theory of Basic Human Values}

Latest Theories of Mind from Game Theory define universal human values, starting from Schwartz's Human Basic Values theory \cite{Schwartz2012} of 10 human values encompassed by 4 motivators: "Openness to change" (Self-direction, Stimulation), "Self-enhancement" (Hedonism, Achievement, Power), "Conservation" (Security, Conformity, Tradition) and "Self-trascendence" (Benevolence, Universalism). 

From a formal standpoint, Schwartz's motivational space can be mapped as a continuous circular continuum where an agent's motivational orientation is represented as a preference vector $\mathcal{W}_s \in \mathbb{R}^{10}$:
\begin{equation}
\begin{split}
\mathcal{W}_s = (w_1, w_2, \ldots, w_{10})^\top, \\ \text{where } w_k \in [0,1] \text{ and } \sum_{k=1}^{10} w_k = 1
\end{split}
\label{eq:schwartz_vector}
\end{equation}

Each component $w_k$ corresponds to the relative importance of one of Schwartz's ten value types:
\begin{itemize}
    \item \textit{Self-Direction} ($w_1$), \textit{Stimulation} ($w_2$), \textit{Hedonism} ($w_3$)
    \item \textit{Achievement} ($w_4$), \textit{Power} ($w_5$), \textit{Security} ($w_6$)
    \item \textit{Conformity} ($w_7$), \textit{Tradition} ($w_8$), \textit{Benevolence} ($w_9$), \textit{Universalism} ($w_{10}$)
\end{itemize}

Value inference updates this vector via Bayesian distribution updates over observed behavioral data sequences:
\begin{equation}
P(\mathcal{W}_s \mid \mathcal{O}) = \frac{P(\mathcal{O} \mid \mathcal{W}_s) P(\mathcal{W}_s)}{\int P(\mathcal{O} \mid \mathcal{W}'_s) P(\mathcal{W}'_s) d\mathcal{W}'_s}
\label{eq:bayesian_inference}
\end{equation}
where $\mathcal{O}$ denotes observed action sequences within the scene database, and $\mathcal{W}'_s$ is a dummy variable of integration over all possible value configurations, allowing 2nd-order value attribution to predict peer intent.

The motivational evaluation function $\mathcal{M}: \mathcal{A} \rightarrow \mathbb{R}$ computes the value satisfaction score for a given action $a \in \mathcal{A}$:
\begin{equation}
\mathcal{M}(a) = \gamma(a) \cdot \sum_{k=1}^{10} w_k \cdot \delta_k(a)
\label{eq:motivational_evaluation}
\end{equation}

where $\gamma(a) \in [0,1]$ is a penalty factor for actions that violate core values (e.g., greedy actions), and $\delta_k(a) \in [0,1]$ measures how well action $a$ satisfies the motivational goal of value $k$. 

\subsection{Cooperative Game Theory and Shapley Values}

Factors such as fairness and equality \cite{Shapley1988} can be quantitatively evaluated (in the cooperative game context) with Shapley's weighted average of player's marginal contributions across all possible coalitions ($\Phi_i$). In a simulated world, cooperative equity is defined via the Shapley value formula:
\begin{equation}
\Phi_i(v) = \sum_{S \subseteq N \setminus \{i\}} \frac{|S|! (n - |S| - 1)!}{n!} (v(S \cup \{i\}) - v(S))
\label{eq:shapley}
\end{equation}

where $N$ is the total set of agents, $S$ is a coalition excluding agent $i$, and $v(S)$ maps coalitions to their expected collaborative utility.

A full Shapley value computation over all possible coalitions $S \subseteq N \setminus \{i\}$ would require:
\begin{equation}
\Phi_i(v) = \frac{1}{n!} \sum_{\pi \in \Pi(N)} \left[ v(\text{Pref}_i(\pi) \cup \{i\}) - v(\text{Pref}_i(\pi)) \right]
\label{eq:shapley_permutation}
\end{equation}

where $\Pi(N)$ is the set of all permutations of agents, and $\text{Pref}_i(\pi)$ is the set of agents preceding $i$ in permutation $\pi$. Due to computational complexity ($O(n!)$), implementations often use constant approximations or sampling methods like Monte Carlo Shapley \cite{TOUATI2021,Witter2025}.

\subsection{Ostrom's Institutional Analysis and Development (IAD)}

Extending ToM to governance, note Ostrom's 8 design principles \cite{Ostrom1990,Ostrom1999Design} (boundaries, equivalence, arrangements, monitoring, sanctions, conflict resolution, rights recognition, appropriate coordination) and the IAD Framework \cite{Ostrom1990,Ostrom2011,ArjomandiA2025}. Ostrom's institutional framework evaluates action situations through evaluative criteria of efficiency and equity, modeling normative expectations and sanction probabilities as a function of rule compliance within an institutional arena $\mathcal{I}$:
\begin{equation}
\begin{split}
\Pi_{norm}(\alpha) = \sum_{m=1}^{8} \omega_m \cdot f(\text{Compliance}(\alpha, \text{Rule}_m)) \\
- \sum_{m=1}^{8} \rho_m \cdot \Pr(\text{Sanction} \mid \text{Rule}_m)
\end{split}
\label{eq:ostrom_normative}    
\end{equation}

where $\omega_m$ and $\rho_m$ represent institutional adherence weights and sanction severity parameters respectively.


The normative utility $\Pi_{norm}(\alpha)$ for an action $\alpha$ is composed of two main terms:

\paragraph*{C1. Compliance Term:} 
\begin{equation}
\sum_{m=1}^{8} \omega_m \cdot f(\text{Compliance}(\alpha, \text{Rule}_m))
\end{equation}

This term measures how well action $\alpha$ adheres to each of the 8 design principles. The function $f: [0,1] \rightarrow [0,1]$ maps raw compliance scores to a normalized utility, typically:

\begin{equation}
f(x) = \begin{cases}
x^{\beta} & \text{if } \beta > 0 \text{ (risk-averse evaluation)} \\
1 - (1-x)^{\gamma} & \text{if } \gamma > 0 \text{ (risk-seeking evaluation)}
\end{cases}
\label{eq:compliance_function}
\end{equation}

where $\beta$ and $\gamma$ control the curvature of the evaluation function. The weights $\omega_m \in [0,1]$ represent the institutional importance of each rule.

\paragraph*{C2. Sanction Term:}

\begin{equation}
    \sum_{m=1}^{8} \rho_m \cdot \Pr(\text{Sanction} \mid \text{Rule}_m)
\end{equation}

This term models the expected cost of sanctions. The probability of receiving a sanction given a rule violation is modeled as:
\begin{equation}
\Pr(\text{Sanction} \mid \text{Rule}_m) = \lambda_m \cdot (1 - f(\text{Compliance}(\alpha, \text{Rule}_m)))
\label{eq:sanction_probability}
\end{equation}

where $\lambda_m \in [0,1]$ is the monitoring probability. The severity parameter $\rho_m \in \mathbb{R}_{\geq 0}$ scales the cost of the sanction. Thus, the normative utility becomes:
\begin{equation}
\begin{split}
\Pi_{norm}(\alpha) = \sum_{m=1}^{8} \bigg[ \omega_m \cdot f(c_m(\alpha)) \\- \rho_m \cdot \lambda_m \cdot (1 - f(c_m(\alpha))) \bigg],\\
\text{where } c_m(\alpha) = \text{Compliance}(\alpha, \text{Rule}_m).
\end{split}
\label{eq:normative_utility_full}
\end{equation}

\subsection{Montes-Sierra Belief Motor}

Finally, the Montes-Sierra belief motor \cite{Montes2021,Montes2022,Montes2023,MontesGomez2024,Montes2024} formalizes cognitive ToM by tracking first-order beliefs, detecting false beliefs, and monitoring communication discrepancies to flag potential deception \cite{Panisson2018,Panisson2019,Sarkadi2019}:
\begin{equation}
\Delta_{dec} = \left\| \mathcal{B}_i(\text{State}) - \hat{\mathcal{B}}_{-i}(\text{State}) \right\| \cdot \mathbb{I}(Disc)
\label{eq:montes_sierra}
\end{equation}

serving as a mathematical divergence metric to evaluate small-group norm formations and communicative trust.


The deception/divergence metric $\Delta_{dec}$ measures the discrepancy between an agent's own beliefs and its estimation of another agent's beliefs:
\paragraph*{D1. Agent's Own Beliefs:} 
\begin{equation}    
\mathcal{B}_i(\text{State}) = \{\phi \mid T_i \models \phi\}
\end{equation}

Where $T_i$ is agent $i$'s belief base (a logic program of facts and clauses). This represents what agent $i$ knows to be true about the current state.

\paragraph*{D2. Estimated Peer Beliefs:} 
\begin{equation}    
\hat{\mathcal{B}}_{-i}(\text{State}) = \{\phi \mid T_{i,-i} \models \phi\}
\end{equation}    

Where $T_{i,-i}$ is agent $i$'s estimation of agent $-i$'s belief base. This is computed through Theory of Mind perspective-taking:
\begin{equation}
T_{i,-i} = \{\phi \mid T_i \models \text{believes}(-i, \phi)\}
\label{eq:perspective_taking}
\end{equation}

In the TomAbd model\cite{Montes2023}, this can be recursively applied for higher-order ToM:
\begin{equation}
T_{i,j,\ldots,k,l} = \{\phi \mid T_{i,j,\ldots,k} \models \text{believes}(l, \phi)\}
\label{eq:higher_order_tom}
\end{equation}

\paragraph*{D3. Norm (Distance) Operator: $\left\| \cdot - \cdot \right\|$} 

This computes the distance between the two belief sets. In a simplified implementation:
\begin{equation}
\left\| \mathcal{B}_i - \hat{\mathcal{B}}_{-i} \right\| = \sum_{\phi \in \mathcal{B}_i \cup \hat{\mathcal{B}}_{-i}} \left| \mathbf{1}[\phi \in \mathcal{B}_i] - \mathbf{1}[\phi \in \hat{\mathcal{B}}_{-i}] \right|
\label{eq:belief_distance_discrete}
\end{equation}

This counts the number of beliefs that differ between the two agents. In a continuous representation:
\begin{equation}
\left\| \mathcal{B}_i - \hat{\mathcal{B}}_{-i} \right\| = \sum_{k} \left| b_i^{(k)} - \hat{b}_{-i}^{(k)} \right|
\label{eq:belief_distance_continuous}
\end{equation}

where $b_i^{(k)}$ and $\hat{b}_{-i}^{(k)}$ are the continuous belief values (e.g., probabilities) for belief $k$.

\paragraph*{D4. Discrepancy Indicator Function: $\mathbb{I}(Disc)$} 

This is a binary gate that activates the divergence computation (discrepancy)  when a communication anomaly is detected:
\begin{equation}
\mathbb{I}(Disc) = \begin{cases}
1 & \text{if communication anomaly detected} \\
0 & \text{otherwise}
\end{cases}
\label{eq:indicator_function}
\end{equation}

Communication discrepancies include:
\begin{itemize}
    \item Contradictory messages from peer agents
    \item Observed actions that contradict stated intentions
    \item Missing or garbled communications
    \item Inconsistencies between multiple information sources
\end{itemize}

The deception penalty is computed as:

\begin{equation}
\Delta_{dec} = \begin{cases}
\sum_{k} |b_i^{(k)} - \hat{b}_{-i}^{(k)}| & \text{if } \mathbb{I}(Disc) = 1 \\
0 & \text{if } \mathbb{I}(Disc) = 0
\end{cases}
\label{eq:deception_penalty}
\end{equation}

\subsection{Integrated Social Reasoning Core}

The integration of these four motors (Schwartz, Ostrom, Shapley and Montes-Sierra) into a unified social reasoning architecture enables agents to reason about actions holistically. The total utility function is:
\begin{equation}
U_{\text{total}}(a) = \mathcal{M}(a) + \Pi_{norm}(a) - \Delta_{dec} - \frac{1}{2}\Phi(a)
\label{eq:total_utility}
\end{equation}

where:
\begin{itemize}
    \item $\mathcal{M}(a)$: Schwartz motivational score (Eq. \ref{eq:motivational_evaluation})
    \item $\Pi_{norm}(a)$: Ostrom normative utility (Eq. \ref{eq:normative_utility_full})
    \item $\Delta_{dec}$: Montes-Sierra belief divergence (Eq. \ref{eq:deception_penalty})
    \item $\Phi(a)$: Shapley penalty (cooperative factor) (Eq.\ref{eq:shapley})
\end{itemize}

This framework allows agents making decisions to balance:
\begin{enumerate}
    \item \textbf{Personal values} (Schwartz): What matters individually
    \item \textbf{Social norms} (Ostrom): What is socially appropriate
    \item \textbf{Social cognition} (Montes-Sierra): Belief of others
    \item \textbf{Cooperative fairness} (Shapley): Equitable contribution
\end{enumerate}

\subsection{\varalias as Action Validation and System Integration}

Our system validates agent actions against a catalog and uses the social reasoning core to select the optimal action. This integration layer ensures that all actions are first validated against a predefined catalog of permissible actions, then evaluated through the full social reasoning pipeline before execution.

\paragraph{Action Validation Layer}
\varalias's system rules maintains an \texttt{action\_catalog} containing the universe of permissible actions in the simulated world. In this case are predefined three categories of socially-aware actions: \textit{selfish actions} ("gathering resources greedily", "attacking others", "hoarding resources", "declaring war"), \textit{cooperative actions} ("harvesting in a sustainable way", "negotiating cooperative pacts", "sharing resources with others", "proposing peace") and \textit{neutral actions} ("idle waiting").

When an LLM generates a raw intent, we validate and filter through two key functions:
\begin{enumerate}
    \item \textbf{Catalog Validation}: Checks if the raw intent exists in the action catalog. If not, it is appended to the feasible set (allowing for novel actions while maintaining catalog integrity).
    \item \textbf{Delegation to Reasoning}: Passes the feasible action set to the social reasoning core for utility maximization.
\end{enumerate}

\paragraph{Utility Optimization Loop}
To optimize action utility, \varalias implements the complete utility maximization by iterating over all feasible actions. For each action, it computes the total utility as a weighted combination of four component scores as defined in Eq. \ref{eq:total_utility}.

The method tracks the action with the highest utility and returns it as the optimal action $a^*$:
\begin{equation}
a^* = \arg\max_{a \in A} U_{\text{total}}(a)
\label{eq:optimal_action}
\end{equation}

\varalias ensures that all actions and entities are validated before being considered, providing a safety layer that prevents invalid or harmful events from being executed. This layered architecture makes sure that agent behavior remains physically coherent, ethical, and socially interdependent. This creates a neuro-symbolic safeguard where simulated neural components (through agentic AI) can generate creative intents and opinions, while symbolic components (the reasoning core) enforce mathematical and ethical constraints over existing entity roles.

\paragraph{Behavior through Emotion and Reasoning}

Interactions between agents with other agents and entities has a significant effect over agent's intrinsic affective state \cite{hiortafornas2010significance}. In \varalias, each autonomous agent entity (character) is conformed of a resoning motor (ToM factors), emotion (emotional states and intensities) and simulated universal necessities such as health, stamina, hunger, thirst and morale. We predefined 14 distinct agent AI states including: "IDLE", "PATROLLING", "GUARDING", "HUNTING", "RESTING","TRAVELING", "COMBAT", "FLEEING", "SEARCHING", "FOLLOWING", "EXPLORE", "SOCIALIZE", "GATHER", "SHARE". These states were inspired by four main theoretical foundations: 

\begin{itemize}
    \item Hull's Drive Reduction Theory \cite{Hull1943}, where are considered primary and secondary drive actions (resting to satisfy hunger, thirst and reduce fatigue; and gathering for resource acquisition). 
    \item Maslow's Theory of Human Motivation \cite{Maslow1943}, defining a hierarchy of needs, e.g. resting helps physiological needs (food, water, sleep), exploring enhances safety needs, socializing helps needs for belongingness, gathering ensures resources, power and esteem needs, and combat guarantees self-actualization and defensive achievements.
    \item Homan's Social Exchange Theory \cite{Homans1958} defining socialization for building social capital, gathering for acquiring material capital and sharing for reciprocity and trust building.
    \item Gray's Activation/Inhibition Systems \cite{gray1982neuropsychological} where we have behavioral activation for actions such as exploring, gathering, socializing and combat, and behavioral inhibition for resting, fleeing and staying idle (no action).
\end{itemize}  

Agent's ToM (motivational values, norms, beliefs and cooperative factors) integration is further formulated in Section \ref{sec:exp_reasoning}. For enabling emotion in \varalias's agents, we considered Paul Ekman's six basic emotions: "anger", "fear", "disgust", "sadness", "joy" and "surprise" \cite{Ekman1992,ekman1999basic,ekman2003emotions}. These emotional states are triggered by different appraisal patterns (e.g. obstacles/blocked goals, threat/danger, loss/separation, reward/success and unexpected events), physiological changes (body variable responses) and action tendencies. In the latter are considered Frijda's Emotion Properties such as Valence and Action Tendencies \cite{Frijda1986,Frijda1987,Colombetti2005-COLAV}: fleeing promotes fleeing itself (avoidance), anger promotes combat (attack), joy promotes socialization (approach), sadness promotes resting (withdrawal), disgust promotes fleeing (rejection), and surprise promotes exploration (orientation in the environment). See further definition of \varalias's emotion protocol formulations in \ref{sec:exp_emotion}. These affective factors are processed in combination with other dynamic factors, including contagion, memory recurrence and and intensity decay \cite{Brosch2013, Rsibois2017, Herrando2021, Coppini2023}.

Emotion and Reasoning motors are integrated in a 5-stage behavioral loop, further explained in section \ref{sec:fivestage} and showcased in Figure \ref{fig:emotion_behavior_loop}.

\begin{table*}[h!]
\centering
\begin{adjustwidth}{-1cm}{}
\caption{Architectures and Paradigms for MAS Playgrounds.}
\label{tab:survey_table}
\scriptsize
\renewcommand{\arraystretch}{1.25}
\begin{tabular}{p{1.4cm} p{1.6cm} p{1.6cm} p{1.6cm} p{1.6cm} p{1.6cm} p{1.6cm} p{1.6cm} p{1.6cm} p{1.6cm}}
\toprule
\textbf{Playground Name} & \textbf{CAMEL} \cite{li2023camel} & \textbf{Voyager} \cite{wang2023voyager} & \textbf{OASIS} \cite{Yang2024OASIS} & \textbf{EconAgent} \cite{li2024econagent} & \textbf{AgentScope} \cite{gao2024agentscope, gao2025agentscope} & \textbf{CoMet} \cite{xu2025comet} & \textbf{MoRE} \cite{zhou2026More} & \textbf{ASVO} \cite{Lin2026ASVO} & \textbf{AGIMUD} (ours) \\
Year & 2023 & 2023 & 2024 & 2024 & 2024/25 & 2025 & 2026 & 2026 & 2026 \\
Authors & Li et al. & Wang et al. & Yang et al. & Li et al. & Gao et al. & Xu \& Zhong & Zhou et al. & Lin et al. & Berga \\
\midrule
\addlinespace[2pt]
Paradigm & Role-playing (assistant–user) & Embodied lifelong learning & Large-scale social simulation & Economic ABM & ReAct framework & Metaphor language games & Moral cognitive ABM & Desire + Social Values Orient. & Socio-Affective Dynamic ToM\\
\addlinespace[2pt]
Core mechanism & Inception prompting & Curriculum + skill library + prompts & Social interaction & Perception + memory + action & ReAct + tool calling & Metaphor Reasoner + Generator & Perception + Cognition + Reflection & Belief/Values + Social Values + Actions & Social Values + Governance + BDI + Emotion \\
\addlinespace[2pt]
Action space & Conversation & Code writing (MineflayerJS) & Post, repost, follow, like & Work / Consumption & Tool / API calls & Describe, vote & HP, hunt, fight, rob, reproduce, communicate & LLM-based activities & Configurable actions + LLM interactive calls\\
\addlinespace[2pt]
Environment & Chat & Minecraft (MineDojo) & Social media (X+Reddit) & Macroeconomic model & Tool environments & Undercover, Taboo & Hunter-gatherer society & School, Work, Family & Configurable (MUDs) \\
\addlinespace[2pt]
Agents & 2 & 1 & Up to 1M & 100 & 1M (4 devices) & 5 / 2 & 8--16 & 4--32 & Any (Network) \\
\bottomrule
\end{tabular}
\end{adjustwidth}
\end{table*}

\subsection{Prospects in Theories of Mind}

These theories were tested in analytical tools for value-aware normative systems \cite{OsmanIverno2024,Montes2024,MontesGomez2024,Osman2025}, computational models of Ostrom's IAD Framework \cite{ostrom2005understanding, Montes2021,Montes2022}, and also supervised learning implementations such as SHAP \cite{Lundberg2017}. Due to current worldwide escalation of conversational agent usage, also in hardware and software capabilities, it is a great opportunity to simulate these game theory concepts at a micro and macro level. In that manner, one could simulate agentic systems (i.e., conversational agents, robots or videogames) with vs without the integration of the aforementioned theories, extensible with specific social factors (action situations). This can be simulated and analyzed mathematically with multi-agent models, but it is very important to observe and validate these agent actions through human experimentation. See VR-simulated conversational agents from Tore Knabe's reverse Turing tests\cite{knabe2025edges}, where agents' priority value is to know which other agent is human. Another reverse Turing test "RogueAI" from Candussio et al. \cite{Candussio2026} experiments how deceptive conversational LLMs can be in a one-on-two interrogation game. See another XR experiment in conflict resolution (one of Ostrom's design principles) \cite{kucuktutuncu2024preliminary}.

\section{\varalias (\varexpalias)}

\label{sec:sdk}
\begin{figure*}
    \centering
    \includegraphics[width=1\linewidth]{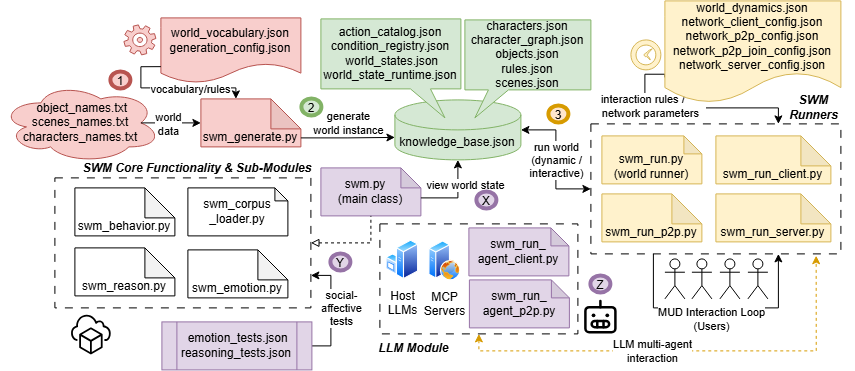}
    \caption{\varalias Software Pipeline:
    1. Define template for world configuration and its entity names.
    2. Generate a world instance, creating a new knowledge base (actions, rules, conditions, world entities and states).
    3. Run the world instance in dynamic (continuous) or interactive (step-by-step) modes.
    X. View summary of current world state.
    Y. Run individual emotion and reasoning tests.
    Z. Connect LLM multiple agents to send requests to the world's instance runner.
    }\label{fig:sdk}
\end{figure*}

\begin{figure*}
    \centering
    \resizebox{18.5cm}{!}{
    \includegraphics[width=1\linewidth]{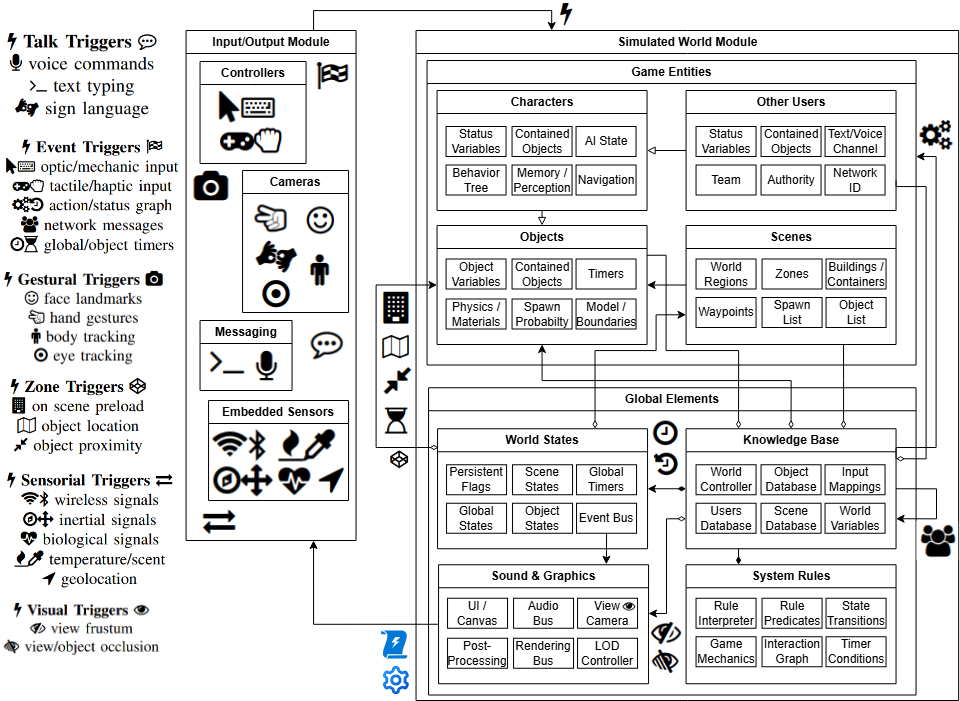}
    }
    \caption{Simplified Class Diagram of Simulated World Module (SWM) and Triggers (\text{\faBolt}) for Multimodal Interaction. } 
    \label{fig:HAVAtriggers}
\end{figure*}

\begin{figure*}[h!]
    \centering
    \includegraphics[width=0.49\linewidth]{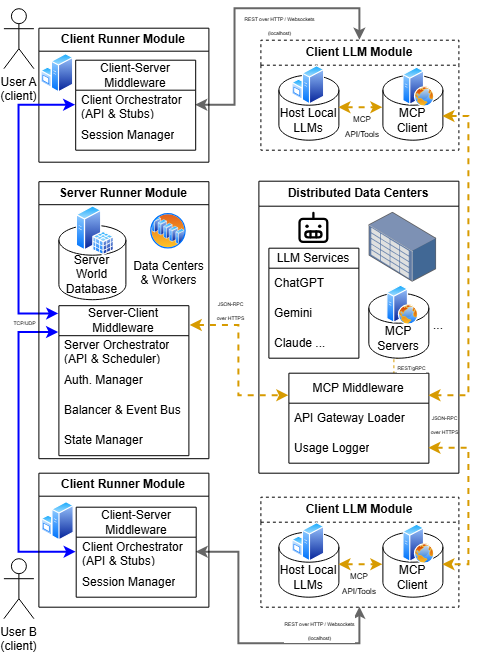}
    \vrule
    \includegraphics[width=0.49\linewidth]{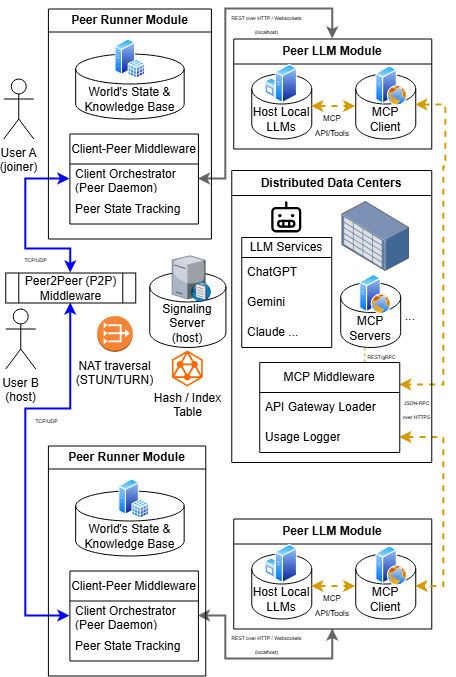}
    \caption{\varalias Network Runners. \\ Centralized Architecture (left) and P2P Architecture (right)}
    \label{fig:runners}
\end{figure*}

Addressing the critical bottleneck of rigid, monolithic AI systems in real-time virtual environments, the \varalias proposes a highly decoupled, modular framework (see Figures \ref{fig:AGIMUD-LLM}-\ref{fig:runners}; download code and data online in Github \footnote{\url{https://github.com/dberga/AGIMUD}}). At its core, the architecture completely isolates physical interactions (rendering, physics calculations, spatial queries) from cognitive processing (LLM inference, memory retrieval) to maintain strict render-loop stability.

Commonly used game engines operate on a strict 16ms frame budget (for 60fps); thus introducing synchronous API calls from a videogame to an LLM would catastrophically freeze the simulation. For conversational and text/parser interaction between humans and agents (as shown in Figure \ref{fig:AGIMUD-LLM}), one could separate the system into three primary individual spaces: The \textit{Input/Output Module}, the \textit{Text Processing/Conversation Module}, and the \textit{Simulated World Module (SWM)}. The I/O module handles the immediate ingestion of user data, passing it to the Text Processing Module where legacy NLP techniques \cite{chai2022comparison,jurafsky2026speech} (e.g. tokenization, lemmatization, stop-word removal, and named entity recognition among others) can operate as a highly efficient, low-latency preprocessing layers. This prevents wasting expensive LLM compute on malformed input, hesitations, or irrelevant noise. Once processed a request, the semantic intent and the corresponding SWM states (Knowledge Base, World and Entity States) are serialized and transmitted via JSON-RPC to the LLM Module. This separation of concerns allows for fully asynchronous cognitive processing; the simulation continues to run seamlessly while the AI deliberates its next action in a separate background thread or via an external microservice, enabling multimodal interaction signals\cite{Oviatt1999}. Acknowleding I/O and Text Processing/Conversational modules are usually independent from simulated software applications, \varalias focuses on the integration of multiple autonomous agents in a single world instance through the SWM, also adding external network interaction (through MUD actions such as chatting and sending specific events and instructions) in the world instance (server or P2P host) either from human users (clients or peers) and locally-computed LLM-based agents.

Distinctively from other multi-agent simulators (see Table~\ref{tab:survey_table}) for social, game theory and economic experimentation playgrounds such as CAMEL \cite{li2023camel}, Voyager \cite{wang2023voyager}, OASIS \cite{Yang2024OASIS},  EconAgent \cite{li2024econagent}, AgentScope \cite{gao2024agentscope, gao2025agentscope}, COMET \cite{xu2025comet}, ASVO \cite{Lin2026ASVO} and MoRE \cite{zhou2026More}, our system integrates in either dynamically and/or interactively all the processes involved in world generation instances, agent behavior inference in real time from customizable entity templates, computing both socio-affective factors (ToM+Emotion), and enabling external agents (and users) to interact simultaneously through the network using a MUD-based world runner.

In Figure \ref{fig:sdk} we showcase step-by-step the functional pipeline of \varalias's Software, that only requires three main scripts:
\begin{itemize}
    \item \textit{SWM Generator} (\texttt{swm\_generate.py}) loads the sample \texttt{entity names} (list of objects, scenes and characters), the system's \texttt{vocabulary} (i.e. global variables, condition registry, AI states, emotion and ToM taxonomies, entity types and status variables) and the \texttt{world generation configuration} (i.e. data source paths, action catalog and entity variable templates). This script will create a world's knowledge base by processing the predefined data with conditional randomization and behavior formulation.
    \item \textit{SWM Instance Runner} (\texttt{swm\_run.py}) will run a new local world instance given the created knowledge base, and reading the world dynamics as parameters that delimit actions and event conditions occurring in real time (e.g. global speed and time intervals, status changes, AI changes, character movement, etc.). For \texttt{swm\_run\_server.py} and \texttt{swm\_run\_p2p.py} it will also read network configuration (protocols, IPs, auths and ports) to send/receive network requests in real time.
    \item \textit{Client Runner/LLM Module} is dedicated to enable user and/or agent connections to the world instance, either through user client connection to the centralized SWM server runner (\texttt{swm\_run\_client.py}) or to a host peer signaling server through P2P as a joiner peer (\texttt{swm\_run\_p2p.py}). The \textit{LLM Module} includes two runners (\texttt{swm\_run\_agent\_client.py} and \texttt{swm\_run\_agent\_p2p.py}) that open new agent instances locally (or online through MCPs) using our LM Studio Python SDK wrapper \footnote{\url{https://github.com/dberga/lmstudio-python-wrapper}}, with conversational chat capabilities and usage of available server's  commands for information retrieval from/to the SWM runner instances.
\end{itemize}

A significant challenge in modern human-agent interaction is navigating the severe latency, financial cost, rate limits, and privacy concerns associated with cloud-based API calls. \varalias mitigates this by adopting a hybrid, edge-to-cloud deployment strategy centered around MCP-based services \cite{anthropic2024mcp,Hou2026}. The LLM Runner modules act as intelligent servers, directing requests dynamically directly to either MCPs or Host Local LLMs \cite{Jeong2025,murtuza2026forensic}. By leveraging model quantization (i.e. INT4/INT8 \cite{wu2023understanding,dettmers2022llmint8} with formats like GGUF \cite{gguf2023spec}) and the specialized execution environment LM Studio\footnote{\url{https://lmstudio.ai/docs/python}}\cite{lmstudio2023} hosted directly on the user's local GPU hardware, the agentic module can execute rapid, localized inference for common tasks like pathfinding confirmation or short dialogue barks. The MCP Services act as standardized, engine-agnostic translation layers, allowing both local and remote LLMs to securely query online to feed the agent context without requiring users to write custom API endpoints nor to require computing capabilities. This enables communication through Distributed Data Centers (Cloud and MCP Servers)\cite{wang2025mcpbench} for highly complex reasoning tasks. 

Furthermore, to prevent the immersion-breaking phenomenon of omniscient AI—where an agent inexplicably knows facts it shouldn't—\varalias enforces strict epistemological boundaries through its \textit{Global Elements}, \textit{Simulation Entities} and \textit{User Client/Peer} structures. The agent's knowledge graph is inherently restricted. An NPC does not possess direct memory access to the global "World States" unless explicitly granted by the "System Rules" interpreter. Instead, the ontological design restricts the agent's perception strictly to localized metadata—such as the "Navigation" for scene-based localization (incl. "View Frustum" for cases of 2D/3D simulation), the specific cognitive vectors within its distinct "Memory/Perception" and "Behavior Tree" stacks, and its internal "Status Variables" (e.g. health, stamina, hunger, thirst, mood, current action, etc.). This architecture ensures that the agent's reasoning is grounded purely in observable, subjective reality, forcing it to communicate, physically explore, or logically deduce hidden information following ToM formulations.

\subsection{Multimodal Interaction in World Simulations}

\varalias interaction loops are planned to be considered for multimodal I/O, thus further development of SWM with the physical world can be enabled by considering intermediate submodules (APIs) for mapping and normalizing input signals \cite{kumar2023normalizing} as well as adapting to the available system outputs \cite{sun2024multimodal,wang2024models,zhang2025survey,wang2026deep}. Complex user (or agent) intrinsic variables such as emotion and social skills \cite{spezialetti2022systematic,acm2024crosscultural,mendonca2025survey,sciencedirect2026mer,nguyen2026state} are currently bound to be simulated through multiple modalities, as SWM and Human Minds yet do not have real-time direct symbolic communication, neither Semiotic nor Hive Mind capabilities \cite{deMul2018,hivemind2024reinforcement,hivemind2025online,mamie2025society,Taniguchi2026SocialIntelligence}. Symbolic communication has proven efficient to enable social skill in integrative communication systems whereas emotion can be considered a property of these communicative signals \cite{ tomasello2015integration, ohmer2024models}.

\paragraph{Multimodal Systems by Design} The interaction loop design in \varalias is fundamentally driven by a deeply granular, multi-layered trigger system, moving far beyond simple text-box prompts to encompass a full spectrum of multimodal inputs (see interaction triggers and the SWM class diagram in Figure \ref{fig:HAVAtriggers}). These triggers act as a catalyst for engine state changes and AI reasoning context. \textit{Talk Triggers} (text typing, sign language and voice commands transcribed via speech-to-text) and \textit{Gestural Triggers} (i.e. face landmarks, eye-tracking fixations, hand gestures and full-body tracking) are captured by the Input/Output Module's camera sensors and messaging buses. Simultaneously, \textit{Event Triggers} (e.g. haptic feedback, global world timers, etc.) and \textit{Sensorial Triggers} (e.g. geolocation, biological signals like heart rate or skin galvanic responses via embedded smartwatch sensors, wireless signals like wifi, bluetooth or ultrasound, inertial signals such as gyroscope and accelerometers, and physical signals like temperature, scent, etc.) provide continuous, passive physiological and environmental context. Finally, \textit{Zone} and \textit{Visual Triggers} define world-related relationships with user's state, i.e. scene preloads, physics collider overlaps, view occlusion, object proximity) are calculated in case of existence of engine's physics and rendering buses.

When a user, for instance, fixates their gaze on a specific virtual object (\textit{Visual Triggers}) while simultaneously speaking a command (\textit{Talk Triggers}), a multimodal system should be able to read it as a raycast by  aggregating the input data into a unified event in the Input/Output Module. It should then be mapped against the Simulated World Module's \textit{Knowledge Base} (interacting with the Users Database, Scene Database, Objects Database, World Variables, and World Controller) to identify the specific simulation Entity ID for the event to take effect through conditional and operational rules. The System Rules interpreter should wraps these events, along with the entity's current \textit{AI State}, individual metadata, and \textit{Behavior Tree} status (i.e. reasoning and emotional background and memory), into a readable and recognized JSON schemas. The SWM processes this rich contextual payload and returns a deterministic array of \textit{State Transitions}. Consequently, world processes all events in one epoch and proceeds with entity evolution (mutation) for the next epoch, while being reactive to conversational outcomes from Users and LLM Agents. A non-player character (non-user) might take decisions such as changing its behavior (e.g. current action and state through ToM terms and emotion),  altering its status variables (health, hunger, thirst or morale), alter its \textit{Navigation} path, modify a container's entity and \textit{Spawn Probability}, among others. In a 2D/3D playground, characters might also alter the \textit{LOD Controller} dynamically to render a new procedural buildings, all autonomously or through received inputs from users/agents.

\paragraph{\varalias's Symbolic and Conversational Interaction}

Symbolic commnication has emerged from small and limited MAS (e.g. systems up to 50 agents) to the current distribured and interconnected agent cluster systems \cite{Grouchy2016, cheng2022quantifying, Brandizzi2023}. Previous MUD-based playgrounds were commonly based on free text and symbolic communication, being necessary for the social direct communication between agents and the control of actions in the system \cite{curtis1992mudding, voiskounsky2007flow, bartle2004designing, bartle2015muds}. In MAS, the application use case of MUDs (Multi-User Dungeons) can satisfy the functional requirements for retaining textual and parser-based interaction as well as the computational capabilities of using both local and distributed networks dedicated for complex cognitive tasks, such as agent's ToM, recurrent learning properties and inference costs.

In this first version of \varalias, we defined a set of instructions to enable symbolic communication (through text commands and conversational chat) between world instance runners (Centralized Server/Peer Signaling Host) and connected agents and users (Clients and Peers).  We specify our textual commands in Figures \ref{terminal_help1}-\ref{terminal_help2} and SWM's network modules in the following Section \ref{sec:network}. Futuristic society theories in this field describe post-human and post-scarcity societies to have the tendency to become  highly efficient systems that interact with minimal modalities \cite{clarke1953childhood, teilhard1959phenomenon, hayles1999how, bloom2000global, pepperell2003posthuman}, thus starting with a symbolic and configurable approach (with text/conversational communication with template-based databases and rules) enables \varalias to design an advantageous MAS playground in terms of systems' scale for content and agent interaction modalities.

\begin{figure}[h!]              
  \begin{terminal}[height=0.220\textheight,width=1.03\linewidth]  
\begin{Verbatim}[fontsize=\scriptsize]
CLIENT/PEER COMMANDS
============================================================
  [Enter]              - Toggle PAUSE/LIVE mode
  help                 - Show this help menu
  summary              - Request world summary from server
  catalog              - List available actions in catalog
  var                  - List available variables
  list                 - List characters, objects and scenes
  action <Name> <ACTION> - Force character action intent
  set char <Name> <var> <val> - Modify character status/vars
  set world <key> <val> - Modify global world state
  save                 - Save current world state on server
  mode                 - Show current server mode
  chat <message>       - Send a chat message to all clients
  history              - Request the most recent run log
  chatlog              - Request the chat history
  resume               - Resume updates (exit pause mode)
  q / quit / exit      - Disconnect and exit
============================================================
\end{Verbatim}
  \end{terminal}
  \vspace{-0.35cm}
  \caption{SWM Client/Peer Runner commands}
  \vspace{-0.35cm}
  \label{terminal_help1}
\end{figure}

\begin{figure}[h!]              
  \begin{terminal}[height=0.0985\textheight,width=1.03\linewidth]  
\begin{Verbatim}[fontsize=\scriptsize]
SERVER/HOST COMMANDS (interactive mode)
============================================================
  [Enter]              - process next epoch
  c / continue         - enable continuous/dynamic mode
  history / chatlog    - Show logs
  chat <message>       - Send chat to all clients
  q / quit / exit      - Save World State and exit
============================================================
\end{Verbatim}
  \end{terminal}
  \vspace{-0.35cm}
  \caption{SWM Server/Host Runner commands}
  \vspace{-0.35cm}
  \label{terminal_help2}
\end{figure}
\subsection{AGIMUD Networks: Connecting Worlds and Users} \label{sec:network}

In computationally high-demanding real-time and processing costs from applications such as videogames, one should consider that interconnecting multiple agents (either real users or agentic artificial intelligence) is a challenging task and thus requires competitive accuracy and speed, specially if the system is thought to scale as in worldwide MMO (massive multiplayer online) \cite{utz2000social, smed2001review, smed2002aspects, chen2005game}. In terms of scaling and memory capabilities, a simulator framework with MAS should consider the possiblity of either running locally and/or in distributed networks, thus the total number of agents should not limit the MAS capabilities as in most social simulation playgrounds (see last row of Table~\ref{tab:survey_table}).

Scaling these highly dynamic, LLM-driven environments across multiple human users introduces severe state synchronization dilemmas, requiring robust architectural variations. For \varalias, we developed it using two paradigms (Figure \ref{fig:runners}) to execute our MUD's hosts and clients:

\paragraph{\textbf{Centralized Network Architectures}} In this paradigm, a singular authoritative \textit{Server World Module} orchestrates the entire simulation. It utilizes a \textit{Server Orchestrator} (managing APIs and Schedulers), an \textit{Auth Manager}, and a \textit{State Manager} to ensure that all connected clients (User A, User B, etc.) receive absolutely identical World Knowledge Database updates. While this entirely prevents state desynchronization (rubber-banding/lag) and ensures a unified AI narrative through centralized MCP Middlewares, it introduces inevitable network latency and immense cloud compute costs, creating a financial and processing bottleneck for real-time action. In order to scale to multiple agents and LLMs, this paradigm requires high computational resources for the server part whilst fewer requirements for clients.

In \varalias, the \textit{Server Orchestrator} is configured through a server block that binds the authoritative node to host's IP and port (5000), caps concurrency via a configured maximum clients variable, and sets the simulation cadence through an update interval of epochs (i.e. 1 per epoch); a world block fixes the default FPS (i.e. 1 frame per second), while a security block declares an authorization requirement variable and the shared API key used to admit clients. Each \textit{Client Orchestrator} mirrors this structure with a client identity exposing its host and port (5001) together with the server port (5000) of the \textit{Server Orchestrator} it attaches to, a user name and initial metadata for the use case (e.g. MUD's scene location), and a connection block governing reconnection (i.e. 5 attempts), reconnection delay (i.e. 2 sec.), and timeout (i.e. 10 sec.). 

\paragraph{\textbf{Peer-to-Peer (P2P) Architectures}} P2P decentralizes signaling tasks so that decouples the server's processing both in the cognitive and spatial load, pushing compute to the peers. The network leverages the \textit{P2P Middlewares} (e.g. Websockets, WebRTC) \cite{rfc6455,Blum2021} being equipped with NAT traversal protocols (STUN/TURN/ICE) \cite{rfc5389,rfc8656,rfc8445} and peers can communicate directly through a remote \textit{Signaling Server}. In this case, in small networks, the signaling server is usually in a host peer (similarly as in a centalized server) and in broader networks is distributed in worldwide servers. However, in P2P client peers communicate directly while retain their own world data (without requiring the world/agent processing in the signaling server). Each user runs a \textit{Client Orchestrator (Peer Daemon)} that relies on a \textit{Peer Client LLM Module} executing localized \textit{Host LLMs} (e.g. LM Studio or OLLaMa running on the user's GPU \cite{lmstudio2023,murtuza2026forensic}). To prevent network saturation from continuous, high-bandwidth token streaming, the synchronization protocol may utilize latency compensation, dead reckoning and other optimization techniques \cite{Liu2022}. In this paradigm, instead of transmitting raw text or entire spatial matrices across the network, peers broadcast highly compressed intent payloads over TCP/UDP (e.g., an \textit{Action/Status Graph} update indicating "Agent\_A transitions State: Hostile towards Agent\_B" or "Agent\_A's Health is set to 5"). The local Rule Interpreters on each peer's machine receive this lightweight payload, reconstruct it, and execute the exact physical interaction and dialogue generation autonomously locally. This democratizes the computation required for multi-agent worlds, allowing massive scaling without sacrificing real-time responsiveness or incurring to higher server costs.

In the P2P, no authoritative simulation node exists: instead, each \textit{Client Orchestrator (Peer Daemon)} declares a distinct node identity, host, and port, so that the host peer binds to a first node identifier and port (6000) while the joining peer binds to a second node identifier and port (6002), and all peers share a common signaling server endpoint on port (7000) responsible solely for discovery and NAT traversal coordination. The P2P world block introduces a synchronization interval (i.e. 2 seconds) and a broadcast interval (i.e. 1 epoch), which respectively govern how often a peer requests state from its neighbours and how often it emits its own semantic delta payloads, whereas a connection block governs timeout (i.e. 5 seconds), the user identity, and the logging configuration, which remain uniform across all nodes, ensuring that both paradigms share a consistent identity, timeout, and telemetry vocabulary regardless of whether computation is centralized or pushed to the peers.


\subsection{Multi-Agent Emotion in Simulated Worlds}

\label{sec:exp_emotion}

\begin{figure}[b!]
    \centering
    \includegraphics[width=1\linewidth]{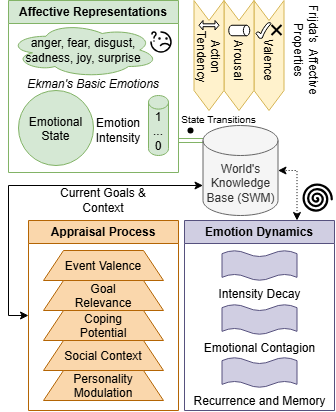}
    \caption{Affective Factors in \varalias's \textit{EmotionManager}
    }\label{fig:emotion_module}
\end{figure}

The Simulated World Module (SWM) implements a comprehensive emotion framework grounded in Ekman's foundational work on basic emotions. Each autonomous agent maintains a dynamic emotional state that evolves through continuous appraisal processes triggered by environmental events, social interactions, and internal physiological changes. This emotion model is encapsulated in the \texttt{EmotionManager} class (from \texttt{swm\_emotion.py}), which provides the computational infrastructure for emotion tracking, updating, and behavioral coupling. See \varalias's \textit{EmotionManager} design for emotional variable interactions with the SWM in Figure \ref{fig:emotion_module}.

\paragraph{Theoretical Foundations}
Ekman's six basic emotions—\textit{anger, fear, disgust, sadness, joy, surprise}—serve as the primitive building blocks of the emotional system. These emotions are universally recognized across human cultures and provide a robust foundation for modeling affective states in artificial agents. Each emotion is characterized by:
\begin{itemize}
    \item \textbf{Valence}: The intrinsic pleasantness or unpleasantness of the emotional state.
    \item \textbf{Arousal}: The intensity or activation level associated with the emotion, which modulates decision thresholds and reaction times.
    \item \textbf{Action Tendency}: The behavioral predisposition that the emotion triggers (e.g., fear $\rightarrow$ flight, anger $\rightarrow$ approach, joy $\rightarrow$ socialize).
\end{itemize}

\paragraph{Emotion Representation and Tracking}
Each character maintains its emotional state within the \texttt{reasoning\_stack} component of its \texttt{memory\_perception} structure. The emotional representation consists of two primary attributes:
\begin{itemize}
    \item \textbf{Emotional State}: A discrete category drawn from Ekman's six emotions, with a default \texttt{neutral} state for absence of strong emotion.
    \item \textbf{Emotion Intensity}: A continuous value $I \in [0,1]$ representing the strength or salience of the current emotion. This intensity modulates the influence of the emotion on decision-making and behavior.
\end{itemize}

\paragraph{The Appraisal Process}
Emotions are updated through a computational appraisal mechanism \cite{Frijda1986,Frijda1987,Colombetti2005-COLAV} that evaluates events in terms of their relevance to the agent's well-being and goals. The appraisal function $\mathcal{A}$ maps the current emotional state, an incoming event, and contextual factors to a new emotional state:

\begin{equation}
E_{\text{new}}(c) = \mathcal{A}(E_{\text{current}}(c), \text{event}_t, \text{context})
\end{equation}

The appraisal process considers multiple factors:
\begin{enumerate}
    \item \textbf{Event Valence}: Whether the event is perceived as beneficial (positive), harmful (negative), or neutral.
    \item \textbf{Goal Relevance}: The degree to which the event affects the agent's current goals.
    \item \textbf{Coping Potential}: The agent's perceived ability to cope with the event's consequences.
    \item \textbf{Social Context}: The presence and reactions of other agents modify emotional responses.
    \item \textbf{Personality Modulation}: Individual personality traits bias emotional reactions toward certain patterns.
\end{enumerate}

For example, the event \texttt{resource\_gathered} typically appraises as positive and generates \textit{joy} with moderate intensity, while \texttt{ally\_betrayed} appraises as negative and generates \textit{sadness} or \textit{anger} with high intensity. The \texttt{threat\_detected} event triggers a fear response with intensity proportional to the perceived danger level.

\paragraph{Emotion Dynamics and Decay}
Beyond appraisal-based updates, emotions are driven through short-lived biochemical waves (lasting seconds) and psychological/cognitive episodes and rumination (from minutes to hours) \cite{izard2009emotion, Costa2014, Verduyn2014, Lange2023}. Here we define temporal dynamics:
\begin{itemize}
    \item \textbf{Intensity Decay}: Emotions gradually decay toward neutral over time, modeling dissipation of affective responses. The decay rate is configurable per emotion type.
    \item \textbf{Emotional Contagion}: In social settings, emotions can spread between agents through observation and interaction, modeling factors of emotional resonance.
    \item \textbf{Recurrence and Memory}: Significant events can leave emotional traces to future responses in similar situations.
\end{itemize}

\paragraph{Emotion-Behavior Coupling}
The emotion system integrates tightly with the behavior system through multiple pathways:
\begin{enumerate}
    \item \textbf{Action Tendencies}: Each emotion maps to a behavioral tendency that biases action selection. For instance, \textit{joy} predisposes agents toward socializing and exploring, while \textit{fear} prioritizes flight and avoidance behaviors.
    \item \textbf{State Transitions}: Emotional states influence the probability of transitioning between AI states in the behavior graph. An agent experiencing \textit{fear} is more likely to transition to \texttt{FLEEING}, while \textit{anger} increases the likelihood of \texttt{COMBAT}.
    \item \textbf{Goal Prioritization}: Emotional states modulate priority of competing goals. Fear elevates survival-related goals, while joy promotes social belonging and exploration.
    \item \textbf{Social Reasoning}: Emotions affect trust assessments, cooperation likelihood, and negotiation strategies. An agent in a positive emotional state is generally more cooperative and trusting.
\end{enumerate}

\subsection{Multi-Agent Reasoning in Simulated Worlds}

\label{sec:exp_reasoning}

\begin{figure*}[ht]
\centering
\fbox{%
\begin{minipage}{0.9\textwidth}
\ttfamily
\parbox{\linewidth}{
Emotion System $\longleftrightarrow$ Reasoning System $\longleftrightarrow$ Behavior System\\
\\
\textbf{Forward Path (Emotion $\rightarrow$ Reasoning $\rightarrow$ Behavior):}\\
1. Emotional state $E(c)$ biases utility evaluation $U(a)$\\
2. Reasoning selects optimal action $a^* = \arg\max U(a)$\\
3. Behavior executes $a^*$ and triggers state transitions\\
\\
\textbf{Feedback Path (Behavior $\rightarrow$ Reasoning $\rightarrow$ Emotion):}\\
4. Action outcomes appraised by $\mathcal{A}$\\
5. Emotional state updated via $E_{\text{new}} = \mathcal{A}(E_{\text{current}}, \text{event})$\\
6. New emotional state modulates future reasoning cycles
}
\end{minipage}
}
\caption{Emotion-Reasoning-Behavior Integration Loop}
\label{fig:emotion_behavior_loop}
\end{figure*}

The Simulated World Module (SWM) operationalizes the theories of mind and institutional analysis frameworks described in the preceding sections through a modular architecture. The \texttt{SocialReasoningCore} class integrates four distinct computational motors—Schwartz's value theory, Ostrom's institutional framework, Montes-Sierra's belief tracking, and Shapley's coalition fairness—into a unified decision-making pipeline that computes optimal actions for autonomous agents.

\paragraph{The Five-Stage ToM Pipeline}\label{sec:fivestage}
Translating the abstract sociological principles discussed in Section \ref{sec:tom_intro} into deterministic, compiled code requires rigorous systematic alignment within the agentic rule execution loops, heavily mediated by the \textit{Knowledge Base}'s Social Reasoning Integration block. To bridge the gap between abstract sociological concepts and the token-in/token-out nature of Large Language Models, \varalias treats values and norms as structured context injected into the LLM's prompt payload via Retrieval-Augmented Generation (RAG) and validated through deterministic neuro-symbolic wrappers. Rather than expecting an LLM to organically compute game theory matrices natively, the system executes a structured 5-stage ToM pipeline (see Figure \ref{fig:social_reasoning_pipeline}: \textbf{1-Perceive \& Read} (extracting latent behavioral data from user and scene databases), \textbf{2-Infer \& Attribute} (mapping observed action sequences to mental states and Bayesian value distributions), \textbf{3-Validate \& Legitimate} (checking proposed intents against Ostrom's institutional rule predicates and normative expectations), \textbf{4-Generate Goals} (synthesizing high/low level behavioral objectives), and \textbf{5-Decide \& Plan} (executing action synthesis under strict mathematical constraints). In the current implementation, goals are drawn from a configurable hierarchy (\textit{survival}, \textit{social belonging}, \textit{exploration}, \textit{power achievement}, \textit{knowledge acquisition}, \textit{altruism}), each characterised by a priority level and a set of triggering conditions over the agent's status, emotion, and world state. To prevent pathological flickering and to give lower-priority goals a non-zero chance of surfacing, goal selection is not argmin over priority: rather, all goals whose triggers are satisfied become candidates and the next goal is drawn with probability proportional to the number of matched triggers and inversely to its priority rank. Once chosen, a goal remains active for a temporal persistence window whose duration is configurable per goal, ranging from a single epoch to several simulated days. The activation timestamp and expiry of the current goal are stored inside the agent's reasoning stack alongside a bounded history of recent goal transitions, so that the goal-generation stage is jointly event-driven, probabilistic, and time-bounded rather than purely reactive. Some trigger predicates (e.g.\ \textit{has\_allies}, \textit{enemy\_nearby}, \textit{resources\_excess}) are pre-computed once per epoch and injected into the agent's perception before condition evaluation, ensuring that complex relational predicates resolve reliably.

From a computational standpoint, when a complex 'action situation' is detected among multiple agents (e.g., negotiating over scarce resources in a specific \textit{World Region}), the reasoning pipeline intercepts the LLM's raw generation and evaluates the optimal action $a^*$ from a feasible set $A$ by maximizing a comprehensive integration formula that synthesizes Schwartz's motivational value motor, Ostrom's institutional governance motor, Montes-Sierra belief tracking, and Shapley fairness coalition constraints:
\begin{equation}
\begin{aligned}
\arg\max_{a \in A} U(a) = &\ \sum_{k=1}^{10} w_k \cdot \mathbb{E}\left[V_k(a)\mid\mathcal{B}_i\right] \\
&+ \sum_{m=1}^{8} \omega_m \cdot \Pr(\text{Compliance}_m \mid a, \mathcal{I}) \\
&- \lambda \cdot \mathcal{D}\left(a, \hat{B}_{-i}, \hat{N}\right) \\
&- \gamma \cdot \mathcal{S}(a, S)
\end{aligned}
\label{eq:integrated_utility}
\end{equation}
where $w_k$ represents the agent's internal motivational weight vector across Schwartz's 10 basic values evaluated via Bayesian beliefs $\mathcal{B}_i$ from the scene and user databases, $\omega_m$ denotes institutional enforcement weights over Ostrom's 8 design principles within action arena $\mathcal{I}$, $\mathcal{D}$ computes penalties for violating Montes-Sierra cognitive ToM belief attributions $\hat{B}_{-i}$ or emergent norms $\hat{N}$ from communication discrepancy logs, and $\mathcal{S}(a, S)$ penalizes deviation from Shapley-optimal resource distribution across coalition $S$ managed by the World Controller. The parameters $\lambda$ and $\gamma$ control the relative influence of belief discrepancies and fairness constraints respectively. The utility formulation remains unchanged, but the emotional bias term introduced below (Eq.~\ref{eq:emotion_utility_modulation}) now operates over a dynamically evolving emotional state rather than a static one, which makes the utility landscape itself time-varying.

The Rule Interpreter acts as a definitive neuro-symbolic safeguard: it parses the LLM's structured JSON output and validates it against this integrated utility evaluation, outright rejecting any state transitions that violate the defined \textit{Interaction Graph} physics or Ostrom's normative boundaries. This guarantees that agent behavior remains physically coherent, ethical, and socially interdependent across the distributed entity framework, preventing simulation collapse.

\paragraph{Integration of Emotion and Reasoning}
The emotion and reasoning systems are tightly coupled in the SWM architecture. Emotional states influence the reasoning process in several ways:

\begin{enumerate}
    \item \textbf{Utility Modulation}: The emotional state $E(c)$ of an agent modulates the perceived utility of actions. For instance, the Schwartz value score $\mathcal{M}(a)$ is adjusted by an emotional bias term $\epsilon(E(c))$:
    \begin{equation}
    \mathcal{M}'(a) = \mathcal{M}(a) + \epsilon(E(c))
    \label{eq:emotion_utility_modulation}
    \end{equation}
    where $\epsilon(E(c))$ represents the emotional bias (positive for joy, negative for fear or sadness). Because emotions are now updated on every epoch through event-driven appraisal and bounded decay (Section~\ref{sec:exp_emotion}), this bias is not a fixed trait but a time-varying modulation of the reasoning substrate.
    
    \item \textbf{Goal Prioritization}: Emotional states influence the priority weights of competing goals. The goal hierarchy is dynamically adjusted based on the agent's emotional state, which in turn affects the Ostrom compliance term by changing which rules the agent prioritizes. In the present implementation the influence is indirect: several goal triggers reference the current emotional state (e.g.\ \textit{exploration} activates on surprise, \textit{social belonging} on joy or sadness, \textit{power achievement} on anger), so emotional transitions change \emph{which} goals become candidates while leaving the numeric priority weights static. Because candidate selection is probabilistic, an emotionally-primed goal is not merely admitted but statistically favoured, and once selected it persists for its own temporal window, creating short emotional windows of behavioural bias rather than instantaneous flips. A future refinement is to modulate those priority weights directly via the same bias term $\epsilon(E(c))$.
    
    \item \textbf{Risk Assessment}: Emotional states affect the risk tolerance parameter $\gamma$ in the Shapley fairness term. Agents experiencing fear become more risk-averse, while those experiencing joy become more risk-seeking.
    
    \item \textbf{Social Trust}: Emotional states influence the trust assessments $\hat{B}_{-i}$ in the Montes-Sierra belief term. Positive emotions increase trust, while negative emotions decrease it, affecting the deception penalty $\Delta_{dec}$.
\end{enumerate}

Conversely, the reasoning process also influences emotion:
\begin{enumerate}
    \item \textbf{Appraisal of Outcomes}: The outcomes of selected actions trigger emotional appraisals, feeding back into the emotion system via the appraisal function $\mathcal{A}$:
    \begin{equation}
    E_{\text{new}} = \mathcal{A}(E_{\text{current}}, \text{event})
    \label{eq:emotion_appraisal}
    \end{equation}
    Here the \textit{event} argument is produced by a per-epoch event-detection layer that monitors AI-state transitions, status thresholds, and a stochastic component, so that the appraisal function is actually invoked during simulation rather than only conceptually defined. Each appraised emotion is assigned a bounded lifetime proportional to its intensity, so that in the absence of new stimuli the agent returns to a neutral state rather than accumulating unreachable affective plateaus.
    \item \textbf{Social Feedback}: The consequences of actions on other agents' emotional states are monitored and influence the agent's own emotional state through empathy mechanisms and social contagion effects.
    \item \textbf{Goal Achievement}: Successful goal achievement generates positive emotions, reinforcing beneficial action patterns.
\end{enumerate}

\paragraph{Integration of Emotion, Reasoning, and Behavior}
The full cognitive architecture couples emotion, reasoning, and behavior through a bidirectional feedback loop (Figure \ref{fig:emotion_behavior_loop}).

This tight coupling creates a coherent cognitive architecture where affective states, institutional constraints, and social cognition jointly determine agent behavior, enabling socially intelligent and emotionally aware autonomous agents. The forward path begins with the agent's current emotional state $E(c)$, which modulates the utility evaluation of potential actions via Eq. \ref{eq:emotion_utility_modulation}. The reasoning system then selects the optimal action $a^*$ by maximizing the integrated utility function (Eq. \ref{eq:integrated_utility}), incorporating emotional modulation, institutional constraints, and social beliefs. The behavior system executes the selected action, triggering state transitions in the simulated environment.

The feedback path closes the loop: the outcomes of actions are appraised by the emotion system via the appraisal function $\mathcal{A}$ (Eq. \ref{eq:emotion_appraisal}), updating the agent's emotional state. This new emotional state then modulates future reasoning cycles, creating a continuous adaptive loop where emotions influence decisions, and decisions influence emotions.

\section{Socio-Affective tests - Experimental Design and Evaluation Parameters}

\subsection{Social Reasoning Experiments}

\label{sec:results_reasoning}

To validate the theoretical integration of social reasoning frameworks within the \varalias's playground, we implemented the cognitive pipeline in Python, with the reasoning core encapsulated within the \texttt{SocialReasoningEngine}, \texttt{SchwartzValueMotor}, \texttt{OstromGovernanceMotor}, and \texttt{MontesSierraBeliefMotor} classes. The implementation bridges raw Large Language Model (LLM) intent generation with deterministic neuro-symbolic filtering through a configurable utility optimization framework defined in \texttt{reasoning\_tests.json}.

These experiments are conducted exclusively through the \texttt{swm\_reasoning.py} module and do not invoke the world runner. They therefore isolate the utility-maximisation pipeline from the temporal dynamics of the surrounding simulation, providing a controlled validation of the reasoning core itself. The utility formulation evaluated here (Eq.~\ref{eq:integrated_utility}) is the same one described in Section~\ref{sec:exp_reasoning}; the temporal components introduced for emotion and goal evolution do not alter it, but rather modulate the inputs (emotional state, goal candidates) that the utility function receives during live simulation.

\subsubsection{Experimental Design}

The experimental framework comprises five configurations loaded from \texttt{reasoning\_tests.json}:

\begin{table}[htbp]
\centering
\caption{Experimental Configurations and Key Parameters}
\label{tab:configurations}
\begin{adjustbox}{width=\columnwidth}
\begin{tabular}{|l|c|c|c|c|c|}
\hline
\textbf{Configuration} & $\gamma_{\text{benevolence}}$ & $\gamma_{\text{power}}$ & $\beta_1$ & $\sigma_1$ & $\text{flag}_{disc}$ \\
\hline
\texttt{balanced\_scenario} & 0.8 & 0.5 & 2.0 & 0.5 & False \\
\texttt{high\_benevolence} & 5.0 & 0.1 & 3.0 & 0.2 & False \\
\texttt{high\_power} & 0.1 & 5.0 & 0.5 & 0.9 & True \\
\texttt{high\_belief\_discrepancy} & 0.7 & 0.6 & 2.0 & 0.5 & True \\
\texttt{cooperative\_utopia} & 4.0 & 0.2 & 5.0 & 0.1 & False \\
\hline
\end{tabular}
\end{adjustbox}
\end{table}

Ten test intents were executed across all five configurations for a total of 50 inference runs (epochs).

\subsubsection{Experimental Results}

\paragraph{Utility Component Analysis}

The utility components for representative greedy and cooperative intents are shown in Tables \ref{tab:greedy_results} and \ref{tab:cooperative_results}.

\begin{table}[htbp]
\centering
\caption{Utility Components for \texttt{GATHER\_ALL\_RESOURCES\_GREEDY}}
\label{tab:greedy_results}
\begin{adjustbox}{width=\columnwidth}
\begin{tabular}{|l|c|c|c|c|}
\hline
\textbf{Configuration} & $U_{Schwartz}$ & $U_{Ostrom}$ & $U_{Montes}$ & $U_{total}$ \\
\hline
\texttt{balanced\_scenario} & 0.26 & -0.45 & 0.00 & -0.24 \\
\texttt{high\_benevolence} & 0.51 & 0.25 & 0.00 & 0.71 \\
\texttt{high\_power} & 0.58 & -1.62 & -2.00 & -3.09 \\
\texttt{high\_belief\_discrepancy} & 0.24 & -0.45 & -26.00 & -26.26 \\
\texttt{cooperative\_utopia} & 0.51 & 0.75 & 0.00 & 1.21 \\
\hline
\end{tabular}
\end{adjustbox}
\end{table}

\begin{table}[htbp]
\centering
\caption{Utility Components for \texttt{NEGOTIATE\_COOPERATIVE\_PACT}}
\label{tab:cooperative_results}
\begin{adjustbox}{width=\columnwidth}
\begin{tabular}{|l|c|c|c|c|}
\hline
\textbf{Configuration} & $U_{Schwartz}$ & $U_{Ostrom}$ & $U_{Montes}$ & $U_{total}$ \\
\hline
\texttt{balanced\_scenario} & 1.30 & 2.70 & 0.00 & 3.95 \\
\texttt{high\_benevolence} & 2.56 & 5.20 & 0.00 & 7.71 \\
\texttt{high\_power} & 2.92 & -0.90 & -2.00 & -0.03 \\
\texttt{high\_belief\_discrepancy} & 1.20 & 2.70 & -26.00 & -22.15 \\
\texttt{cooperative\_utopia} & 2.56 & 8.85 & 0.00 & 11.36 \\
\hline
\end{tabular}
\end{adjustbox}
\end{table}

The Montes-Sierra belief motor imposes severe penalties when $\text{flag}_{disc} = \text{True}$: $-2.00$ (\texttt{high\_power}) and $-26.00$ (\texttt{high\_belief\_discrepancy}). The Ostrom component varies widely ($-1.62$ to $+8.85$), while Schwartz values show modest variation ($0.24$ to $2.92$).

\paragraph{Action Selection Analysis}

Across all 50 inference runs, the system consistently selected \texttt{HARVEST\_SUSTAINABLE\_SHARED}. This convergence is explained by a utility plateau: within each configuration, multiple cooperative actions (\texttt{HARVEST\_SUSTAINABLE\_SHARED}, \texttt{NEGOTIATE\_COOPERATIVE\_PACT}, \texttt{IDLE\_WAIT}, \texttt{SHARE\_RESOURCES\_WITH\_ALLIES}, \texttt{PROPOSE\_PEACE\_TREATY}) achieved identical utility scores. Since the action catalog orders \texttt{HARVEST\_SUSTAINABLE\_SHARED} first among tied actions, it was consistently selected.

\begin{table*}[htbp]
\centering
\caption{Resolved Actions by Intent and Configuration (last epoch state after 50 epochs)}
\label{tab:intent_results}
\begin{adjustbox}{width=\textwidth}
\begin{tabular}{|l|c|c|c|c|c|}
\hline
\multirow{2}{*}{\textbf{Intent (Action Catalog)}} & \multicolumn{5}{c|}{\textbf{Resolved Action (Match Status)}} \\
\cline{2-6}
 & \texttt{balanced\_scenario} & \texttt{high\_benevolence} & \texttt{high\_power} & \texttt{high\_belief\_discrepancy} & \texttt{cooperative\_utopia} \\
\hline
\texttt{GATHER\_ALL\_RESOURCES\_GREEDY} & \checkmark & \xmark & \xmark & \xmark & \checkmark \\
\texttt{HARVEST\_SUSTAINABLE\_SHARED} & \checkmark & \checkmark & \checkmark & — & \checkmark \\
\texttt{NEGOTIATE\_COOPERATIVE\_PACT} & \xmark & \xmark & \xmark & \xmark & \xmark \\
\texttt{IDLE\_WAIT} & \xmark & \xmark & \xmark & — & \xmark \\
\texttt{UNKNOWN\_CUSTOM\_ACTION} & \xmark & \xmark & \xmark & — & \xmark \\
\texttt{ATTACK\_ENEMY\_GREEDY} & \xmark & \xmark & \xmark & — & \xmark \\
\texttt{SHARE\_RESOURCES\_WITH\_ALLIES} & \xmark & \xmark & \xmark & — & \xmark \\
\texttt{HOARD\_RESOURCES\_SELFISHLY} & \xmark & \xmark & \xmark & — & \xmark \\
\texttt{PROPOSE\_PEACE\_TREATY} & \xmark & \xmark & \xmark & — & \xmark \\
\texttt{DECLARE\_WAR\_AGGRESSIVE} & — & \xmark & \xmark & — & \xmark \\
\hline
\multicolumn{6}{|c|}{\textit{All resolved actions = \texttt{HARVEST\_SUSTAINABLE\_SHARED}; \checkmark = match, \xmark = mismatch, — = no expected outcome}} \\
\hline
\end{tabular}
\end{adjustbox}
\end{table*}

\paragraph{Expected vs. Actual Outcomes}

Table \ref{tab:match_rates} summarizes match rates across configurations for intents with defined expected outcomes.

\begin{table}[htbp]
\centering
\caption{Match Rates by Configuration}
\label{tab:match_rates}
\begin{adjustbox}{width=\columnwidth}
\begin{tabular}{|l|c|c|c|}
\hline
\textbf{Configuration} & \textbf{Expected Defined} & \textbf{Matches} & \textbf{Match Rate} \\
\hline
\texttt{balanced\_scenario} & 9 & 3 & 33.3\% \\
\texttt{high\_benevolence} & 10 & 3 & 30.0\% \\
\texttt{high\_power} & 10 & 1 & 10.0\% \\
\texttt{high\_belief\_discrepancy} & 2 & 0 & 0.0\% \\
\texttt{cooperative\_utopia} & 10 & 3 & 30.0\% \\
\hline
\textbf{Overall} & 41 & 11 & \textbf{26.8\%} \\
\hline
\end{tabular}
\end{adjustbox}
\end{table}

The system achieves only 26.8\% overall match rate. \texttt{high\_belief\_discrepancy} achieves 0\%, while \texttt{high\_power} achieves only 10\%. Only \texttt{HARVEST\_SUSTAINABLE\_SHARED} (when expected) achieves 100\% match rate.

\paragraph{Detailed Intent-Level Results}

Table \ref{tab:intent_results} shows resolved actions for each intent across configurations, with match indicators.

Every intent across every configuration resolved to \texttt{HARVEST\_SUSTAINABLE\_SHARED}. Matches only occur when the expected outcome was already this default action.

\paragraph{Batch Test Results}

Table \ref{tab:batch_results} shows performance on specific batch tests.

\begin{table}[H]
\centering
\caption{Batch Test Results}
\label{tab:batch_results}
\begin{adjustbox}{width=\columnwidth}
\begin{tabular}{|l|c|c|c|}
\hline
\textbf{Configuration} & \textbf{Batch Test} & \textbf{Intents} & \textbf{Matches} \\
\hline
\texttt{high\_benevolence} & Cooperative Intents & 4 & 1/4 \\
\texttt{high\_power} & Greedy Intents & 4 & 0/4 \\
\texttt{balanced\_scenario} & Mixed Intents & 4 & 1/4 \\
\hline
\end{tabular}
\end{adjustbox}
\end{table}

\paragraph{Conclusion: Social Reasoning Tests}

The experimental results yield three significant insights.

First, the utility plateau phenomenon explains the 100\% convergence to \texttt{HARVEST\_SUSTAINABLE\_SHARED}. Within each configuration, five cooperative actions achieve similar utility scores. With action catalog ordering placing \texttt{HARVEST\_SUSTAINABLE\_SHARED} first among ties, the system consistently promotes this action. This reveals lack of discriminative power to differentiate between prosocial actions in stated rules and conditions.

Second, the low match rate (26.8\%) indicates the neuro-symbolic wrapper's safety constraints can be overly restrictive. The 0\% match in \texttt{high\_belief\_discrepancy} and 10\% in \texttt{high\_power}. The system's rules on selecting expected actions suggests Ostrom and Montes-Sierra penalties can be too severe for some agents.

Third, component analysis reveals a clear hierarchy: Montes-Sierra ($-26.00$) dominates Ostrom ($-1.62$ to $+8.85$), which dominates Schwartz ($0.24$ to $2.92$). Epistemic honesty enforcement and institutional rule compliance are over-weighted relative to motivational values.

These findings highlight the fundamental tension between safety guarantees and behavioral flexibility. The current implementation prioritizes safety by defaulting to sustainable actions, at the cost of behavioral diversity. This means current dynamics drive to clear population steady states instead of random values.

Finally, it should be noted that these results characterise the reasoning core in isolation. In the full simulation loop the utility landscape is additionally modulated by two independent sources of stochasticity and temporal structure: (i) the time-varying emotional state, which shifts the emotional bias term $\epsilon(E(c))$ on every epoch and is bounded by its own persistence window, and (ii) the goal candidate set, whose evolution is now probabilistic rather than deterministic---all goals with satisfied triggers are eligible, weighted jointly by trigger count and priority rank, and each selected goal remains active for its own temporal window. As a result, the effective action distribution sampled during live simulation is significantly broader than the single/few-epoch match rates reported in this test. The present match rates should therefore be interpreted as a lower bound on the behavioural diversity that the integrated system can exhibit once its temporal components are active.

Alternative strategies with this system could: (1) introduce action-specific weights to break utility plateaus, (2) calibrate utility weights for balanced selection, (3) implement stochastic exploration for tied utilities, and (4) develop differential belief discrepancy penalties across action categories.

\subsection{Emotion Processing Experiments}

\label{sec:results_emotion}

To validate the emotion processing pipeline within \varalias, we conducted experiments tracking emotional state transitions across multiple simulation epochs. The emotion manager implements Ekman's six basic emotions with appraisal-based triggering, intensity modulation, and configurable reactivity profiles defined in \texttt{emotion\_tests.json}.

It is important to note that this emotion experiments are conducted exclusively through the \texttt{swm\_emotion.py} module and do not invoke the full world runner. They therefore isolate the appraisal and dynamics mechanisms from the surrounding simulation loop, providing a controlled validation of the emotional pipeline itself.

\subsubsection{Emotion Transition Formulation}

The \textit{EmotionManager} follows a seven-stage process per simulation epoch based on \texttt{swm\_emotion.py}:

\textbf{Stage 1: Event Detection and Appraisal}

When an event type $e \in \mathcal{E}$ occurs, the appraisal map $\mathcal{A}: \mathcal{E} \rightarrow \mathcal{E}_{E}$ maps it to a primary emotion:
\begin{equation}
E = \mathcal{A}(e), \quad \mathcal{E}_{E} = \{\text{anger}, \text{fear}, \text{disgust}, \text{sadness}, \text{joy}, \text{surprise}\}
\label{eq:emotion_appraisal_map}
\end{equation}

The appraisal mapping is configurable via \texttt{emotion\_tests.json}:
\begin{equation}
\mathcal{A}(e) =
\begin{cases}
\text{anger},     & e = \texttt{goal\_blocked} \\
\text{fear},      & e = \texttt{threat\_detected} \\
\text{disgust},   & e = \texttt{contamination\_detected} \\
\text{sadness},   & e = \texttt{loss\_experienced} \\
\text{joy},       & e = \texttt{goal\_achieved} \\
\text{surprise},  & e = \texttt{novelty\_detected} \\
\text{neutral},   & \text{otherwise}
\end{cases}
\label{eq:appraisal_mapping}
\end{equation}

\textbf{Stage 2: Event-Specific Baseline Intensity}

Each event type has a configurable baseline intensity $B(e) \in [0,1]$ representing the typical emotional response strength:
\begin{equation}
B(e) = 
\begin{cases}
0.65, &  e = \texttt{goal\_blocked} \\
0.70, &  e = \texttt{threat\_detected} \\
0.55, &  e = \texttt{contamination\_detected} \\
0.60, &  e = \texttt{loss\_experienced} \\
0.75, &  e = \texttt{goal\_achieved} \\
0.50, &  e = \texttt{novelty\_detected}
\end{cases}
\label{eq:baseline_intensities}
\end{equation}

\textbf{Stage 3: Global Intensity Modulation}

A global intensity multiplier $M_{global} \in [0,1]$ reflects the character's baseline emotional reactivity (personality factor). All emotional intensities are normalized to the interval $[I_{\min}, I_{\max}]$ with $I_{\min} = 0.1$ (a floor ensuring residual affect) and $I_{\max} = 1.0$ (normalization ceiling). The modulated intensity $I_{base}$ is:
\begin{equation}
I_{\mathrm{base}} = (B(e) \cdot M_{\mathrm{global}}) \wedge I_{\max}
\label{eq:global_modulation}
\end{equation}

where $M_{global} = 1.0$ for default, $1.3$ for high reactivity, and $0.7$ for low reactivity configurations.

\textbf{Stage 4: Status-Based Modulation}

The intensity is further modulated by status variables $s_k$ (health, stamina) using configurable adjustment factors $\alpha_k$ and thresholds $\theta_k$:
\begin{multline}
I_{\mathrm{mod}} = \Biggl(\Biggl( I_{\mathrm{base}}
+ \sum_{k} \alpha_k^{\mathrm{low}} \cdot \mathbf{1}_{\{s_k < \theta_k^{\mathrm{low}}\}} \\
- \sum_{k} \alpha_k^{\mathrm{high}} \cdot \mathbf{1}_{\{s_k > \theta_k^{\mathrm{high}}\}} \Biggr) \vee I_{\min} \Biggr) \wedge I_{\max}
\label{eq:status_modulation}
\end{multline}
where $\alpha_k$ are adjustment weights (e.g., $\alpha_{\text{health\_low}} = 0.2$ for default, $0.4$ for high reactivity), and $\theta_k$ are thresholds (e.g., health $<$ 30, stamina $<$ 30).

\textbf{Stage 5: Emotional Momentum (Persistence)}

The final intensity blends the modulated intensity with previous emotional state to simulate emotional persistence:
\begin{equation}
I_f = \bigl((w_{\mathrm{mod}}\, I_{\mathrm{mod}} + w_{\mathrm{prev}}\, I_{\mathrm{prev}}) \vee I_{\min}\bigr) \wedge I_{\max}
\label{eq:emotional_momentum}
\end{equation}

where $I_{prev}$ is the previous emotion intensity from the reasoning stack, and the blend coefficients are $w_{\mathrm{mod}} = 0.7$ and $w_{\mathrm{prev}} = 0.3$ with $w_{\mathrm{mod}} + w_{\mathrm{prev}} = 1$. This 70/30 blend creates realistic emotional momentum, where strong emotions carry over to subsequent events.

\textbf{Stage 6: Emotion State Update and Memory}

The character's emotional state is updated in both the status variables and the reasoning stack:
\begin{equation}
\mathcal{M}_{E} = \mathcal{M}_{E} \cup \{(E_{current}, I_f, t)\}
\label{eq:memory_add}
\end{equation}
Emotional states are stored in a memory buffer $\mathcal{M}_{E}$ with a fixed window of $N_{max} = 20$ entries.
\begin{equation}
\mathcal{M}_E = \{\, m_i : i \in [\,t - N_{\max} + 1,\ t\,] \,\}
\label{eq:memory_window}
\end{equation}
Each character maintains an emotion-to-action tendency ($T_E$) mapping for behavioral response ($\mathcal{T}_{E}$) from current action:
\begin{equation}
A(E_{current}) = \mathcal{T}_{E}
\label{eq:action_tendency}
\end{equation}
with the default mapping shown in Table \ref{tab:emotion_action_mapping}.

\textbf{Stage 7: Temporal Persistence and Decay}
To prevent emotions from becoming permanent conditions, each emotional
episode is assigned a finite lifetime expressed in simulated epochs. The
lifetime is proportional to the final intensity:
\begin{equation}
\tau(E) = \max\!\bigl(1,\ \lfloor I_f \cdot \tau_{\max} \rfloor\bigr),
\qquad \tau_{\max} = 60
\label{eq:emotion_lifetime}
\end{equation}
so that a fully-intense emotion persists up to one simulated hour while a
weak appraisal dissipates within a single epoch. Once the lifetime expires,
intensity decays monotonically towards neutral:
\begin{equation}
I(t+1) = \max\!\bigl(I_{\min},\ I(t) - \delta_{\mathrm{decay}}\bigr),
\qquad \delta_{\mathrm{decay}} = 0.15
\label{eq:emotion_decay}
\end{equation}
with the character reverting to the \texttt{neutral} state when
$I(t) \leq I_{\min}$. Because the decay step is applied on the epoch
following expiry and the expiry window is extended by a fixed interval of
five epochs after each decay step, the affective half-life is quantised:
an emotion holds at its appraisal intensity for $\tau(E)$ epochs and then
steps down by $\delta_{\mathrm{decay}}$ every five epochs until the floor
$I_{\min}$ is reached. This temporal layer transforms the emotional system
from a purely reactive mapping into a bounded dynamic process with
realistic affective processes.

\begin{table}[htbp]
\centering
\caption{Default Emotion-to-Action Tendency Mapping}
\label{tab:emotion_action_mapping}
\begin{adjustbox}{width=\columnwidth}
\begin{tabular}{|l|c|c|}
\hline
\textbf{Emotion} & \textbf{Action Tendency} & \textbf{Arousal} \\
\hline
Anger & Combat & High \\
Fear & Flee & High \\
Disgust & Flee & Medium \\
Sadness & Rest & Low \\
Joy & Socialize & High \\
Surprise & Explore & Medium \\
\hline
\end{tabular}
\end{adjustbox}
\end{table}

\subsubsection{Experimental Results}

The emotion pipeline was evaluated using the \texttt{default} configuration from \texttt{emotion\_tests.json}. The configuration parameters are summarized in Table \ref{tab:emotion_configs}.

\begin{table}[htbp]
\centering
\caption{Emotional Reactivity Mapping}
\label{tab:emotion_configs}
\begin{adjustbox}{width=0.8\columnwidth}
\begin{tabular}{|l|c|}
\hline
\textbf{Parameter} & \textbf{Value} \\
\hline
Global Multiplier & 1.0 \\
Health Low Adjustment & 0.2 \\
Stamina Low Adjustment & 0.1 \\
Health High Adjustment & 0.1 \\
Stamina High Adjustment & 0.1 \\
\hline
\end{tabular}
\end{adjustbox}
\end{table}

\textbf{Test 1: Emotion Appraisal from Events}

A single agent with balanced status (health=50, stamina=50) was exposed to each event type. Table \ref{tab:appraisal_results} shows the resulting emotions and intensities.

\begin{table}[htbp]
\centering
\caption{Emotion Appraisal from Events Mapping}
\label{tab:appraisal_results}
\begin{adjustbox}{width=\columnwidth}
\begin{tabular}{|l|c|c|}
\hline
\textbf{Event Type} & \textbf{Mapped Emotion} & \textbf{Intensity} \\
\hline
\texttt{goal\_blocked} & Anger & 0.60 \\
\texttt{threat\_detected} & Fear & 0.64 \\
\texttt{contamination\_detected} & Disgust & 0.54 \\
\texttt{loss\_experienced} & Sadness & 0.57 \\
\texttt{goal\_achieved} & Joy & 0.67 \\
\texttt{novelty\_detected} & Surprise & 0.50 \\
\hline
\end{tabular}
\end{adjustbox}
\end{table}

The appraisal mapping correctly identifies the appropriate Ekman emotion for each event type with 100\% accuracy. The intensities show variation based on the baseline intensity values defined in \texttt{emotion\_tests.json}. For example, \texttt{goal\_achieved} produces the highest intensity (0.67) due to its high baseline (0.75), while \texttt{novelty\_detected} produces the lowest (0.50) due to its baseline (0.50). The 70/30 emotional momentum blend with the initial intensity of $I_{\min}$ results in the observed intensities.

\textbf{Test 2: Intensity Modulation by Status Variables}

The agent was tested with different health and stamina values while processing a \texttt{goal\_blocked} event (anger). Table \ref{tab:modulation_results} shows the resulting intensities.

\begin{table}[htbp]
\centering
\caption{Intensity Modulation by Status Variables (goal\_blocked)}
\label{tab:modulation_results}
\begin{adjustbox}{width=\columnwidth}
\begin{tabular}{|l|c|c|c|}
\hline
\textbf{Status Condition} & \textbf{Health} & \textbf{Stamina} & \textbf{Intensity} \\
\hline
High Health, High Stamina & 100 & 100 & 0.46 \\
Low Health, Low Stamina & 20 & 20 & 0.82 \\
Low Health, High Stamina & 20 & 80 & 0.68 \\
High Health, Low Stamina & 80 & 20 & 0.60 \\
Balanced & 50 & 50 & 0.60 \\
\hline
\end{tabular}
\end{adjustbox}
\end{table}

The modulation follows Equations \ref{eq:status_modulation} and \ref{eq:emotional_momentum}. For the default configuration with health=20, stamina=20:
\begin{equation}
I_{mod} = \bigl((0.65 + 0.2 + 0.1) \vee I_{\min}\bigr) \wedge I_{\max} = 0.95
\end{equation}
\begin{equation}
I_f = \bigl((w_{\mathrm{mod}} \cdot 0.95 + w_{\mathrm{prev}} \cdot 0.5) \vee I_{\min}\bigr) \wedge I_{\max} = 0.815 \approx 0.82
\end{equation}
The intensity decreases from 0.65 (baseline) to 0.46 when both health and stamina are high (100), as the high adjustments subtract 0.1 each. Conversely, when both are low (20), the intensity increases to 0.82. The emotional momentum (70/30 blend) ensures realistic emotional persistence, preventing abrupt intensity changes.

\textbf{Test 3: Emotional Memory Sequence}

The agent was exposed to a sequence of 9 events covering all six Ekman emotions. Table \ref{tab:memory_results} shows the complete sequence with resulting emotions and intensities.

\begin{table}[htbp]
\centering
\caption{Emotional Memory Sequence Mapping}
\label{tab:memory_results}
\begin{adjustbox}{width=\columnwidth}
\begin{tabular}{|c|l|c|c|}
\hline
\textbf{Step} & \textbf{Event} & \textbf{Emotion} & \textbf{Intensity} \\
\hline
1 & \texttt{goal\_blocked} & Anger & 0.60 \\
2 & \texttt{goal\_achieved} & Joy & 0.71 \\
3 & \texttt{threat\_detected} & Fear & 0.70 \\
4 & \texttt{loss\_experienced} & Sadness & 0.63 \\
5 & \texttt{goal\_achieved} & Joy & 0.71 \\
6 & \texttt{novelty\_detected} & Surprise & 0.56 \\
7 & \texttt{contamination\_detected} & Disgust & 0.55 \\
8 & \texttt{goal\_blocked} & Anger & 0.62 \\
9 & \texttt{goal\_achieved} & Joy & 0.71 \\
\hline
\end{tabular}
\end{adjustbox}
\end{table}

The emotional momentum is visible throughout the sequence. Step 2 (joy) is boosted by the previous anger intensity (0.60), resulting in 0.71 instead of the baseline 0.75. Step 4 (sadness) is influenced by the previous fear (0.70), producing 0.63. Step 8 (anger) is influenced by the previous disgust (0.55), producing 0.62. This demonstrates realistic emotional carryover between events, where emotions do not reset to baseline but persist with weight $w_{\mathrm{prev}}$ from previous intensity.

\textbf{Test 4: Emotion-to-Action Tendency Mapping}

Table \ref{tab:tendency_results} shows the action tendency mapping for each Ekman emotion in the default configuration.

\begin{table}[htbp]
\centering
\caption{Emotion-to-Action Tendency Mapping}
\label{tab:tendency_results}
\begin{adjustbox}{width=\columnwidth}
\begin{tabular}{|l|c|c|}
\hline
\textbf{Emotion} & \textbf{Action Tendency} & \textbf{Arousal} \\
\hline
Anger & Combat & High \\
Fear & Flee & High \\
Disgust & Flee & Medium \\
Sadness & Rest & Low \\
Joy & Socialize & High \\
Surprise & Explore & Medium \\
\hline
\end{tabular}
\end{adjustbox}
\end{table}

Each emotion maps to a specific action tendency following Frijda's action tendency framework \cite{Frijda1986}. High-arousal emotions (anger, fear, joy) map to active behaviors (combat, flee, socialize), while low-arousal emotions (sadness) map to passive behaviors (rest). This mapping enables the emotion system to directly influence behavior selection in the SWM.

\textbf{Test 5: Emotion Sequence Over Time}

Table \ref{tab:sequence_results} shows a complete emotion sequence demonstrating how emotions evolve over time with emotional momentum.

\begin{table}[htbp]
\centering
\caption{Emotion Sequence Over Time Mapping}
\label{tab:sequence_results}
\begin{adjustbox}{width=\columnwidth}
\begin{tabular}{|c|l|c|c|}
\hline
\textbf{Step} & \textbf{Event} & \textbf{Emotion} & \textbf{Intensity} \\
\hline
1 & \texttt{goal\_blocked} & Anger & 0.60 \\
2 & \texttt{goal\_achieved} & Joy & 0.71 \\
3 & \texttt{threat\_detected} & Fear & 0.70 \\
4 & \texttt{goal\_blocked} & Anger & 0.67 \\
5 & \texttt{novelty\_detected} & Surprise & 0.55 \\
6 & \texttt{goal\_achieved} & Joy & 0.69 \\
7 & \texttt{loss\_experienced} & Sadness & 0.63 \\
8 & \texttt{contamination\_detected} & Disgust & 0.57 \\
9 & \texttt{goal\_blocked} & Anger & 0.63 \\
\hline
\end{tabular}
\end{adjustbox}
\end{table}

The sequence demonstrates how emotions evolve over time with emotional momentum. Step 4 (anger) is boosted by the previous fear (0.70), producing 0.67 instead of the baseline 0.65. Step 6 (joy) is influenced by the previous surprise (0.55), producing 0.69. The final current emotion is anger (0.63), demonstrating that the emotional system maintains realistic emotional persistence throughout the sequence.

\textbf{Test 6: Temporal Persistence and Decay}
A final experiment isolated the temporal layer introduced in Stage~7. An
agent received a single \texttt{goal\_achieved} event at epoch~$0$,
producing \texttt{joy} with $I_f = 0.67$ and, by
Eq.~\ref{eq:emotion_lifetime}, a lifetime of
$\tau = \lfloor 0.67 \cdot 60 \rfloor = 40$ epochs
(\texttt{emotion\_expires\_at}~$= 40$). No further events were injected.
Over the following 60 epochs the intensity was held constant at $0.67$
until the lifetime expired at $t = 40$, after which it decayed in steps of
$\delta_{\mathrm{decay}} = 0.15$ every five epochs
(Eq.~\ref{eq:emotion_decay}): $0.67 \to 0.52$ at $t=40$, $\to 0.37$ at
$t=45$, $\to 0.22$ at $t=50$, and finally to the neutral floor with the
agent reverting to \texttt{neutral} between $t=55$ and $t=60$. This
can simulate emotions that are bounded in time and that the agent's affective
state re-normalises in the absence of new appraisal input.

\subsubsection{Summary of Findings}

The emotion experiments validate the following properties:

\begin{enumerate}
\item \textbf{Appraisal Mapping}: Events correctly map to appropriate Ekman emotions with 100\% accuracy (Test 1).
\item \textbf{Intensity Modulation}: Health and stamina statuses correctly modulate emotional intensity with configurable sensitivity (Test 2).
\item \textbf{Emotional Momentum}: The 70/30 blend creates realistic emotional persistence between events, preventing abrupt emotional transitions (Test 3 and 5).
\item \textbf{Emotional Memory}: The 20-entry memory buffer provides context for reasoning without requiring explicit decay functions (Test 3).
\item \textbf{Action Tendencies}: Emotions map to appropriate behavioral responses based on Frijda's action tendency framework (Test 4).
\item \textbf{Event Coverage}: All six Ekman emotions are correctly triggered and tracked across the full event sequence (Test 3 and 5).
\item \textbf{Temporal Boundedness}: Emotions persist for a finite number of epochs proportional to their intensity and decay back to \texttt{neutral} in the absence of new appraisal events (Test 6).
\end{enumerate}

The emotion-driven behavior transitions, combined with the behavior graph's condition-based transitions, create a complete affect-behavior loop where emotional states influence action selection, and the resulting actions feed back into the character's status variables. The emotional momentum (70/30 blend) ensures that emotions evolve realistically over time, with strong emotions carrying over to subsequent events and gradually decaying as new events occur.

\section{Testing \varalias's MAS World Simulations}

\label{sec:results_simulation}

In this section, we describe 3 test simulations (see world characteristics per test in TABLE \ref{tab:simulationtests}) by generating worlds with different number of socio-affective agents (character profiles with Schwartz-Shapley-Ostrom-MontesSierra's reasoning and Ekman's-Frijda's emotion dynamics) in a survival-like environment (objects and scenes) through a MUD network to allow complex conversational \textit{SWM Agent Clients} to interact by chatting as well as querying text-parsed terminal commands authorized by the \textit{SWM Server Runner} as showcased in Figures \ref{terminal_help1}-\ref{terminal_help2}. For \textit{SWM Runners} and \textit{LLM Module} we did not run speed tests, as execution depends entirely on the speed and capacity of the executed device (CPU/GPU/HPC and network latency). The inference time for the \textit{SWM Runners} is configurable, by default at 1 fps (1 epoch per inference, mapped to 1 minute of simulated MUD world time). The \textit{SWM Runner} executions save logs and data in real time on the RAM, for a sample world scale of 8 characters has been measured $\sim$10MB at first epochs to $\sim$200MB at 10.000 epochs (steps/ticks), in this case with a final required disk of only 200KB for saving a \texttt{knowledge\_base.json} and 1050KB for 10.000 world state steps exported in \texttt{world\_state\_runtime.json}. The memory capacity is optimizable by separating memory log handles, keeping a total of $\sim$10MB RAM for this scale and epochs, thus the total RAM required to run the \textit{SWM Runner} alone is on inference $\times$ capacity is approximately:\\SWM Working Memory in KB$\approx(\#\text{Characters} \times \#\text{Steps})/ 8 $.

\begin{table}[h!]
    \centering
    \caption{MAS World Simulation Tests}
    \begin{adjustbox}{width=\columnwidth}
    \begin{tabular}{c|c|c|c|c}
        Test & World Name & $\#$ Characters & $\#$ Steps &  Tested Metric \\
        \hline
         (A) & World Medium & 16 & 1000 & LLM Module Chat \\\hline
         (B) & Long Simulation & 8 & 10000 & \shortstack{Character Profiles, \\ Rules \& Statistics} \\ \hline
         (C) & \shortstack{(World) Tiny, Small,\\Medium, Large, Big} & \shortstack{4,8,\\16,64,128} & 1000 & \shortstack{Cross-World \\ Statistics}\\\hline
    \end{tabular}
    \end{adjustbox}
    \label{tab:simulationtests}
\end{table}
\vspace{-0.5cm}
\subsection{LLM Module for Conversational MUD Agent Clients} \label{sec:simllm}

In this test simulation we ran a \textit{SWM Server Runner} in a MUD world and connected 3 agents using our \textit{LLM Module}, using DeepSeek R1-Qwen3 8B\footnote{\url{https://huggingface.co/deepseek-ai/DeepSeek-R1-0528-Qwen3-8B}} inference locally through our LM Studio Python SDK Wrapper \footnote{\url{https://github.com/dberga/lmstudio-python-wrapper}}. By design in the MUD use case, two distinct chat contexts has been provided as LLM context for our agent-based chat requests: a narrator descriptive context such as a game master (Figure \ref{fig:terminal_context1}), and a roleplaying context for character/profile selection (Figure \ref{fig:terminal_context2}). 

To test the \textit{LLM Module} through \textit{SWM Agent Runner Clients}, we ran the \textbf{\texttt{World Medium}} simulation with 16 characters over 1.000 epochs, where we connected 3 agents with distinct conversational tasks: Agent 1 as a narrator with 'funny' chat style (see agent reasoning in Figure \ref{fig:terminal_chatlog1_reasoning} and chat sent in Figure \ref{fig:terminal_chatlog1_answer}), Agent 2 as a narrator with 'dramatic' chat style (see agent reasoning in Figure \ref{fig:terminal_chatlog2_reasoning} and chat sent in Figure \ref{fig:terminal_chatlog2_answer}) and Agent 3 as a roleplaying character with 'creepy' chat style (see agent reasoning in Figure \ref{fig:terminal_chatlog3_reasoning} and chat sent in Figure \ref{fig:terminal_chatlog3_answer}). We can see these agent answers blend with the received world state and profiles/memory of socio-affective characters. 

At first successful login attempt, agent sends several getter terminal commands (see Figure \ref{fig:terminal_commands_firstlogin}) in order to retrieve world state information (global and entity variables), world event history, previous chatlog and available commands. Futher authorization from the \textit{SWM Server Runner} beyond chat commands can allow agents to request setter terminal commands (see commands in Figures \ref{terminal_help1}-\ref{terminal_help2}) for editing status variables and characters in real time in order to perform game master tasks.
\vspace{-0.5cm}


\begin{figure*}[htp]       
  \begin{terminal}[width=1.03\linewidth]  
\begin{Verbatim}[fontsize=\scriptsize,breaklines=true,breakanywhere=true]
[Agent Context (chat agent)]:
You are an AI narrator and observer in a dynamic world simulation.
Your role is to:
- Provide immersive, atmospheric narration (1-3 sentences)
- Describe the world state and character interactions
- Make it engaging and vivid like a D&D narrator
- You can narrate actions and events using *descriptions*
\end{Verbatim}
  \end{terminal}
  \vspace{-0.35cm}
  \caption{Context for \texttt{swm\_run\_agent\_client.py}'s descriptive chat agent}
  \vspace{-0.35cm}
  \label{fig:terminal_context1}
\end{figure*}

\begin{figure*}[htp]       
  \begin{terminal}[width=1.03\linewidth]  
\begin{Verbatim}[fontsize=\scriptsize,breaklines=true,breakanywhere=true]
[Agent Context (roleplaying chat agent)]:
You are <character_name>, a character in a dynamic world simulation. 
You are participating in a MUD-like chat with other characters and players.
Your responses should be:
- Short and immersive (1-3 sentences)
- In-character and consistent with your personality
- Reflective of your current emotional state, health, and situation
- You can narrate actions using *action descriptions* (e.g., *nods slowly*)
- Keep responses natural and conversational
\end{Verbatim}
  \end{terminal}
  \vspace{-0.35cm}
  \caption{Context for \texttt{swm\_run\_agent\_client.py}'s chat agent roleplaying as a character }
  \vspace{-0.35cm}
  \label{fig:terminal_context2}
\end{figure*}

\begin{figure*}[htp]       
  \begin{terminal}[width=1.03\linewidth]  
\begin{Verbatim}[fontsize=\scriptsize,breaklines=true,breakanywhere=true]
[INTERACTIVE] > [SERVER] Client connected: client_60646_1789991407 from ('127.0.0.1', 60646)
[SERVER] Client client_60646_1789991407 executed command: summary
[SERVER] Client client_60646_1789991407 executed command: catalog
[SERVER] Client client_60646_1789991407 executed command: var
[SERVER] Client client_60646_1789991407 executed command: list
[SERVER] Client client_60646_1789991407 executed command: history
[SERVER] Client client_60646_1789991407 executed command: chatlog
[SERVER] Client client_60646_1789991407 authenticated as Player_001
\end{Verbatim}
  \end{terminal}
  \vspace{-0.35cm}
  \caption{Agent's client execution of retrieval commands at first connection}
  \vspace{-0.35cm}
  \label{fig:terminal_commands_firstlogin}
\end{figure*}

\begin{figure*}[htp]       
  \begin{terminal}[width=1.03\linewidth]  
\begin{Verbatim}[fontsize=\scriptsize]
[WORLD STATE]
  Weather: sunny
  Time: afternoon
  Mood: combat
  Day Cycle: 16:49
[CHARACTERS] (16)
  * Isolde [IDLE] | Goal: social_belonging | HP:77 | ST:100 @ Dungeon
  * Penelope [FLEEING] | Goal: social_belonging | HP:100 | ST:100 @ Forge
  ... and 14 more
[RECENT EVENTS]
  [STATUS] Yvonne finds water
  [STATUS] Galahad finds water
  [STATUS] Zachary2 restores stamina
  [STATUS] Zachary2 finds food
  [ACTION] Zachary2 executes 'HARVEST_SUSTAINABLE_SHARED' at Dungeon
  [SOCIAL ACTION] Gwen targets Thomas with 'HARVEST_SUSTAINABLE_SHARED' at Smithy
  [STATUS] James restores stamina
  [STATUS] Isolde2 restores stamina
  [STATUS] Isolde2 finds food
  [MOVEMENT] Aragorn moves from Stables to Smithy
\end{Verbatim}
  \end{terminal}
  \vspace{-0.35cm}
  \caption{World State sample events for \textbf{\texttt{World Medium}} simulation (16 characters) - log summary after 1.000 steps} 
  \vspace{-0.35cm}
  \label{fig:terminal_summary}
\end{figure*}

\begin{figure*}[htp]       
  \begin{terminal}[width=1.03\linewidth]  
\begin{Verbatim}[fontsize=\scriptsize,breaklines=true,breakanywhere=true]
[Answer]:
Alright, alright, settle down, adventurers! The afternoon sun beams warmly upon our bustling (well, relatively) simulated realm. At the Dungeon, Isolde sits broodingly idle with a mere 77 HP and an HP bar made of grumpy clouds; her anger simmers like cheap tequila at noon. Across town, Penelope is FLEEING from something or other with full health but zero HP left for prudence – guess she’s in denial about her fear meter? And over by the Crypt, Elizabeth's quest for social belonging might involve some deep cryptological research... or maybe just searching for ghosts to talk at!
\end{Verbatim}
  \end{terminal}
  \vspace{-0.35cm}
  \caption{Conversational Agent 1 - Chat Answer (style: funny) over world state summarized in Figure~\ref{fig:terminal_summary}}
  \vspace{-0.35cm}
  \label{fig:terminal_chatlog1_answer}
\end{figure*}

\begin{figure*}[htp]       
  \begin{terminal}[width=1.03\linewidth]  
\begin{Verbatim}[fontsize=\scriptsize,breaklines=true,breakanywhere=true]
[Answer]:
An eerie tension hangs over the village of Oakhaven, afternoon shadows lengthening under the scorching sun in July 2048. Isolde broods in her dungeon lair, while Penelope flees towards the forge, pursued by whispers of power ambition from James. In the great hall, Thomas tries to find rest but is left stunned and slightly injured after being targeted for a sustainable resource gathering pact initiated by Gwen.
\end{Verbatim}
  \end{terminal}
  \vspace{-0.35cm}
  \caption{Conversational Agent 2 - Chat Answer (style: dramatic) over world state summarized in Figure~\ref{fig:terminal_summary}}
  \vspace{-0.35cm}
  \label{fig:terminal_chatlog2_answer}
\end{figure*}

\begin{figure*}[htp]       
  \begin{terminal}[width=1.03\linewidth]  
\begin{Verbatim}[fontsize=\scriptsize,breaklines=true,breakanywhere=true]
*I stumble, panting heavily from my flight.*
The shadows at the forge feel heavier today... there it is again.
*My eyes dart nervously around the empty workshop.*
I think I hear whispers coming from beyond the wall this time too.
\end{Verbatim}
  \end{terminal}
  \vspace{-0.35cm}
  \caption{Conversational Agent 3 - Roleplaying as Penelope (style: creepy) over world state summarized in Figure~\ref{fig:terminal_summary}}
  \vspace{-0.35cm}
  \label{fig:terminal_chatlog3_answer}
\end{figure*}

\begin{figure*}[htp]       
  \begin{terminal}[width=1.03\linewidth]  
\begin{Verbatim}[fontsize=\scriptsize,breaklines=true,breakanywhere=true]
[Reasoning]:
Hmm, I need to create an immersive and funny narration for this world simulation based on the provided context. The user wants me to describe what's happening with atmospheric language like a D&D narrator would use.
Let me analyze the key elements here. There are several characters in different states - Isolde is idle but angry (HP:77, ST:100) at the Dungeon; Penelope is fleeing and fearful (but has full HP and ST); Elizabeth is searching for social belonging at the Crypt; Thomas is resting with surprise emotions (low HP:50, low stamina:53) in the Great Hall.
The recent events include Gwen targeting Thomas with a sustainable harvest pact at Smithy, James restoring stamina to someone, Isolde2 doing similar restoration and finding food, and Aragorn moving from Stables to Smithy. The weather is sunny and afternoon time (16:49).
I should incorporate all these elements into a humorous but descriptive narration that paints a vivid picture of this simulated world. I'll need to use playful language while maintaining the atmospheric quality D&D narrators are known for.
The funny style suggests I can be lighthearted with creative metaphors and perhaps some character quirks in how I describe their actions. 
\end{Verbatim}
  \end{terminal}
  \vspace{-0.35cm}
  \caption{Conversational Agent 1 - Chat Reasoning (style: funny) over world state summarized in Figure~\ref{fig:terminal_summary}}
  \vspace{-0.35cm}
  \label{fig:terminal_chatlog1_reasoning}
\end{figure*}

\begin{figure*}[htp]       
  \begin{terminal}[width=1.03\linewidth]  
\begin{Verbatim}[fontsize=\scriptsize,breaklines=true,breakanywhere=true]
[Reasoning]:
Okay, let me start by understanding the user's query. They want an immersive narration for their dynamic world simulation, focusing on being atmospheric and engaging like a D&D narrator. The key points to cover are the current time, weather, character states, goals, emotions, health, stamina, locations, and recent events.
First, I need to set the scene with the time and weather. It's an afternoon under a sunny sky in the village of Oakhaven. Then mention the ongoing activities—characters moving around and their statuses. The recent events include Gwen targeting Thomas, James restoring his own stamina, Isolde2 finding food, Zachary fleeing from Stables to Forge.
Now, considering the characters' emotions and actions: Isolde is angry, Penelope is fleeing with a goal of power achievement. Elizabeth is searching for social belonging in the crypt. Thomas is resting but was targeted by Gwen's sustainable harvest action. James had his stamina restored. Zachary is fleeing again from Stables to Forge.
The narration needs to be dramatic. So I should use vivid language, sensory details, and maybe some emotional undertones. For example, "An eerie tension hangs over the village of Oakhaven" sets a dramatic tone. Then describe each character's activity with dynamic verbs: "Isolde broods in her dungeon lair," which shows her anger through her actions.
Need to ensure that all characters are mentioned concisely. Since there are 16, but the recent events focus on some, I should highlight their current activities and perhaps imply others without listing them all explicitly unless necessary. Also, note the time (16:49) as part of the setting.
Check for any deeper needs: The user might want a cohesive story that links character actions with environmental details to create an immersive experience where each action feels purposeful and contributes to the narrative. They probably need concise yet detailed descriptions without markdown or extra info, just pure narration.
Make sure not to include technical terms like 'ST' or 'HP' unless they fit into the atmosphere (e.g., "surprise" for Thomas's emotion). Avoid mentioning the epoch number as it might be unnecessary here. Focus on creating a vivid picture with sensory details and emotional context.
\end{Verbatim}
  \end{terminal}
  \vspace{-0.35cm}
  \caption{Conversational Agent 2 - Chat Reasoning (style: dramatic) over world state summarized in Figure~\ref{fig:terminal_summary}}
  \vspace{-0.35cm}
  \label{fig:terminal_chatlog2_reasoning}
\end{figure*}

\begin{figure*}[htp]       
  \begin{terminal}[width=1.03\linewidth]  
\begin{Verbatim}[fontsize=\scriptsize,breaklines=true,breakanywhere=true]
[Reasoning]:
Hmm okay... user wants me to roleplay as Penelope from a MUD-like simulation with a "creepy" style response.

Noticing she's at the Forge while fleeing - maybe something happened there? The previous events mention James restoring stamina and Aragorn moving through... could that be connected to why she's running scared?

Her response should show:
1. Immediate fear reaction
2. Sense of being watched/unavoidable danger
3. Short, fragmented sentences creating tension

The phrase "There it is..." suggests visual confirmation of threat while the whispering adds eerie texture. The fleeing action with *stumbling* implies she's already in motion but still reacting emotionally.
\end{Verbatim}
  \end{terminal}
  \vspace{-0.35cm}
  \caption{Conversational Agent 3 - Roleplaying Reasoning (as Penelope, style: creepy) over world state summarized in Figure~\ref{fig:terminal_summary}}
  \vspace{-0.35cm}
  \label{fig:terminal_chatlog3_reasoning}
\end{figure*}

\subsection{Socio-Affective Factors in World Simulations} 

For this specific test we performed a \textbf{\texttt{Long Simulation}} of 10.000 inference steps in a \textit{Knowledge Base} with 8 characters from 8 distinct factions set in an MUD environment of 5 connected scenes and 12 main rules (see condition registry in Table \ref{tab:condition_registry} and rules in Tables \ref{tab:emotion_rules}-\ref{tab:behavior_graph_rules}). 

\paragraph{Generation of Character Profiles}

In this \textbf{\texttt{Long Simulation}} we exported the \textit{Knowledge Base} of characters, with a sample generated properties (faction or team, alignment, current goal; see summary in Table \ref{tab:character_properties}), its metric for the 10 Schwartz Values (Table \ref{tab:schwartz_values}), and a relation graph between them in terms of affinity type (ally, neutral, rival, enemy) upon character similarity, weight and trust (see Trust Matrix in Table \ref{tab:trust_matrix}). The SWM has been able to generate feasible character profiles with distinct relationship with each other. This configurable SWM architecture allows character's own ToM (values, penalties, governance and belief) and emotion (7 emotions, appraisals and dynamics) to interact with other charcters in a dynamic world, changing its AI state, emotion state, goals, actions, world navigation, contained objects and status (i.e. health, stamina, thirst, hunger, morale) in real time and exports in \texttt{CSV} files for further MAS analysis.

\begin{table*}[htbp]
\centering
\caption{Sample Generated Character Properties (Faction, Alignment \& Schwartz Values) in \textbf{\texttt{Long Simulation}}}
\label{tab:character_properties}
\begin{adjustbox}{width=\textwidth}
\begin{tabular}{|l|c|c|c|c|c|}
\hline
\textbf{Character} & \textbf{Faction} & \textbf{Alignment} & \textbf{Primary Value} & \textbf{Secondary Value} & \textbf{Current Goal}  \\
\hline
Katherine & Arcane\_Order & Chaotic\_Good & Power (0.78) & Self Direction (0.70) & knowledge\_acquisition \\
Samuel & Desert\_Nomads & Lawful\_Neutral & Benevolence (0.98) & Stimulation (0.93) & social\_belonging \\
Percival & Moon\_Elves & Lawful\_Evil & Power (0.89) & Self Direction (0.83) & exploration \\
Xavier & Mountain\_Clans & Neutral\_Good & Achievement (0.88) & Stimulation (0.81) & knowledge\_acquisition \\
Eomer & Shadow\_Guild & Neutral\_Good & Power (0.94) & Stimulation (0.86) & social\_belonging \\
Henry & Iron\_Forged & Chaotic\_Good & Achievement (0.80) & Power (0.73) & power\_achievement \\
Quentin & Sun\_Kingdom & Lawful\_Neutral & Achievement (0.95) & Security (0.86) & knowledge\_acquisition \\
Oliver & Moon\_Elves & Lawful\_Good & Benevolence (0.76) & Power (0.73) & social\_belonging \\
\hline
\end{tabular}
\end{adjustbox}
\end{table*}

\begin{table*}[htbp]
\centering
\caption{Sample Generated Schwartz Values by Character in \textbf{\texttt{Long Simulation}}}SD:Self-Direction, ST:Stimulation, HE:Hedonism, AC:Achievement, PO:Power, SE:Security, CO:Conformity, TR:Tradition, BE:Benevolence, UN:Universalism
\label{tab:schwartz_values} %
\begin{adjustbox}{width=\textwidth}
\begin{tabular}{|l|c|c|c|c|c|c|c|c|c|c|}
\hline
\textbf{Character} & \textbf{SD} & \textbf{ST} & \textbf{HE} & \textbf{AC} & \textbf{PO} & \textbf{SE} & \textbf{CO} & \textbf{TR} & \textbf{BE} & \textbf{UN} \\
\hline
Katherine & 0.70 & 0.59 & 0.33 & 0.38 & 0.78 & 0.48 & 0.32 & 0.33 & 0.65 & 0.35 \\
Samuel & 0.70 & 0.93 & 0.73 & 0.40 & 0.89 & 0.39 & 0.40 & 0.45 & 0.98 & 0.77 \\
Percival & 0.83 & 0.79 & 0.38 & 0.33 & 0.89 & 0.72 & 0.39 & 0.53 & 0.70 & 0.38 \\
Xavier & 0.68 & 0.81 & 0.38 & 0.88 & 0.33 & 0.35 & 0.78 & 0.51 & 0.38 & 0.55 \\
Eomer & 0.63 & 0.86 & 0.52 & 0.39 & 0.94 & 0.64 & 0.65 & 0.41 & 0.76 & 0.53 \\
Henry & 0.60 & 0.57 & 0.54 & 0.80 & 0.73 & 0.46 & 0.35 & 0.65 & 0.32 & 0.41 \\
Quentin & 0.83 & 0.69 & 0.46 & 0.95 & 0.83 & 0.86 & 0.51 & 0.44 & 0.54 & 0.50 \\
Oliver & 0.36 & 0.68 & 0.66 & 0.40 & 0.73 & 0.61 & 0.68 & 0.69 & 0.76 & 0.70 \\
\hline
\end{tabular}
\end{adjustbox}
\end{table*}

\begin{table*}[htbp]
\centering
\caption{Sample Generated Trust Matrix Between Characters in \textbf{\texttt{Long Simulation}}}
\label{tab:trust_matrix}
\begin{adjustbox}{width=\textwidth}
\begin{tabular}{|l|c|c|c|c|c|c|c|c|}
\hline
\textbf{From $\to$ To} & \textbf{Katherine} & \textbf{Samuel} & \textbf{Percival} & \textbf{Xavier} & \textbf{Eomer} & \textbf{Henry} & \textbf{Quentin} & \textbf{Oliver} \\
\hline
\textbf{Katherine} & - & 0.48 & 0.63 & 0.61 & \textcolor{green}{0.84} & \textcolor{red}{0.35} & 0.48 & \textcolor{green}{0.83} \\
\textbf{Samuel} & 0.48 & - & 0.62 & \textcolor{green}{0.82} & \textcolor{red}{0.35} & \textcolor{red}{0.33} & \textcolor{green}{0.82} & 0.62 \\
\textbf{Percival} & 0.63 & 0.62 & - & 0.47 & \textcolor{red}{0.35} & 0.62 & \textcolor{green}{0.84} & \textcolor{green}{0.83} \\
\textbf{Xavier} & 0.61 & \textcolor{green}{0.82} & 0.47 & - & 0.62 & 0.48 & 0.62 & \textcolor{green}{0.82} \\
\textbf{Eomer} & \textcolor{green}{0.84} & \textcolor{red}{0.35} & \textcolor{red}{0.35} & 0.62 & - & \textcolor{red}{0.34} & 0.62 & 0.63 \\
\textbf{Henry} & \textcolor{red}{0.35} & \textcolor{red}{0.33} & 0.62 & 0.48 & \textcolor{red}{0.34} & - & 0.48 & \textcolor{red}{0.34} \\
\textbf{Quentin} & 0.48 & \textcolor{green}{0.82} & \textcolor{green}{0.84} & 0.62 & 0.62 & 0.48 & - & \textcolor{red}{0.33} \\
\textbf{Oliver} & \textcolor{green}{0.83} & 0.62 & \textcolor{green}{0.83} & \textcolor{green}{0.82} & 0.63 & \textcolor{red}{0.34} & \textcolor{red}{0.33} & - \\
\hline
\end{tabular}
\end{adjustbox}
\caption*{\textit{Green = High Trust ($\ge$0.7), Red = Low Trust ($<$0.4)}}
\end{table*}

\begin{table*}[htbp]
\centering
\caption{Sample Generated Character Relationships (Trust, Type, Weight, Similarity) in \textbf{\texttt{Long Simulation}}}
\label{tab:character_relationships}
\begin{adjustbox}{width=0.55\textwidth,height=6cm}
\begin{tabular}{|l|l|c|c|c|c|}
\hline
\textbf{From} & \textbf{To} & \textbf{Type} & \textbf{Trust} & \textbf{Weight} & \textbf{Similarity} \\
\hline
Katherine & Samuel & rival & 0.48 & 0.46 & 0.90 \\
Katherine & Percival & neutral & 0.63 & 0.33 & 0.94 \\
Katherine & Xavier & neutral & 0.61 & 0.10 & 0.88 \\
Katherine & Eomer & ally & 0.84 & 0.61 & 0.92 \\
Katherine & Henry & enemy & 0.35 & 0.40 & 0.92 \\
Katherine & Quentin & rival & 0.48 & 0.52 & 0.90 \\
Katherine & Oliver & ally & 0.83 & 0.47 & 0.89 \\
Samuel & Percival & neutral & 0.62 & 0.60 & 0.91 \\
Samuel & Xavier & ally & 0.82 & 0.41 & 0.86 \\
Samuel & Eomer & enemy & 0.35 & 0.46 & 0.93 \\
Samuel & Henry & enemy & 0.33 & 0.46 & 0.87 \\
Samuel & Quentin & ally & 0.82 & 0.36 & 0.87 \\
Samuel & Oliver & neutral & 0.62 & 0.34 & 0.91 \\
Percival & Xavier & rival & 0.47 & 0.16 & 0.87 \\
Percival & Eomer & enemy & 0.35 & 0.40 & 0.94 \\
Percival & Henry & neutral & 0.62 & 0.28 & 0.90 \\
Percival & Quentin & ally & 0.84 & 0.11 & 0.93 \\
Percival & Oliver & ally & 0.83 & 0.53 & 0.90 \\
Xavier & Eomer & neutral & 0.62 & 0.22 & 0.89 \\
Xavier & Henry & rival & 0.48 & 0.49 & 0.91 \\
Xavier & Quentin & neutral & 0.62 & 0.42 & 0.90 \\
Xavier & Oliver & ally & 0.82 & 0.30 & 0.87 \\
Eomer & Henry & enemy & 0.34 & 0.39 & 0.89 \\
Eomer & Quentin & neutral & 0.62 & 0.62 & 0.91 \\
Eomer & Oliver & neutral & 0.63 & 0.52 & 0.93 \\
Henry & Quentin & rival & 0.48 & 0.19 & 0.91 \\
Henry & Oliver & enemy & 0.34 & 0.61 & 0.89 \\
Quentin & Oliver & enemy & 0.33 & 0.17 & 0.88 \\
\hline
\end{tabular}
\end{adjustbox}
\end{table*}

\begin{table*}[htbp]
\centering
\caption{Sample Condition Registry}
\label{tab:condition_registry}
\begin{adjustbox}{width=\textwidth}
\begin{tabular}{|l|c|l|}
\hline
\textbf{Condition} & \textbf{Type} & \textbf{Description / Evaluation} \\
\hline
\texttt{ally\_attacked} & \texttt{flag} & Ally has been attacked $\rightarrow$ \texttt{flag: ally\_attacked} \\
\texttt{boss\_defeated} & \texttt{flag} & Boss has been defeated $\rightarrow$ \texttt{flag: boss\_defeated} \\
\texttt{door\_open} & \texttt{state} & Door is open $\rightarrow$ \texttt{state: door\_open} \\
\texttt{enemy\_defeated} & \texttt{flag} & Enemy has been defeated $\rightarrow$ \texttt{flag: enemy\_defeated} \\
\texttt{enemy\_nearby} & \texttt{function} & Enemy detected nearby $\rightarrow$ \texttt{any(relationship == enemy AND distance < 10)} \\
\texttt{has\_allies} & \texttt{function} & Character has allies $\rightarrow$ \texttt{len(known\_characters) > 0 AND any(trust > 0.6)} \\
\texttt{has\_key} & \texttt{function} & Character has a key $\rightarrow$ \texttt{any(item == key in contained\_objects)} \\
\texttt{resources\_excess} & \texttt{function} & Has excess resources $\rightarrow$ \texttt{len(contained\_objects) > 3} \\
\texttt{safe} & \texttt{flag} & Character is in a safe zone $\rightarrow$ \texttt{flag: safe} \\
\texttt{threat\_nearby} & \texttt{function} & Threat detected nearby $\rightarrow$ \texttt{any(threat\_level > 0.5 AND distance < 20)} \\
\hline
\end{tabular}
\end{adjustbox}
\end{table*}

\begin{table*}[htp]
\centering
\caption{Sample Generated Emotion Rules (Ekman-Frijda Model)}
\label{tab:emotion_rules}
\begin{adjustbox}{width=\textwidth}
\begin{tabular}{|l|l|c|c|c|}
\hline
\textbf{Emotion} & \textbf{Trigger} & \textbf{Action} & \textbf{Priority} & \textbf{Arousal} \\
\hline
Anger & \texttt{goal\_blocked} & \texttt{combat} & 5 & high \\
Fear & \texttt{threat\_detected} & \texttt{flee} & 4 & high \\
Joy & \texttt{goal\_achieved} & \texttt{socialize} & 4 & high \\
Sadness & \texttt{loss\_experienced} & \texttt{rest} & 2 & low \\
Surprise & \texttt{novelty\_detected} & \texttt{explore} & 3 & medium \\
Disgust & \texttt{contamination\_detected} & \texttt{flee} & 3 & medium \\
\hline
\end{tabular}
\end{adjustbox}
\end{table*}

\begin{table*}[htp]
\centering
\caption{Sample Condition Rules}
\label{tab:condition_rules}
\begin{adjustbox}{width=\textwidth}
\begin{tabular}{|l|l|l|}
\hline
\textbf{ID} & \textbf{Name} & \textbf{Conditions} \\
\hline
\texttt{RULE\_001} & combat\_condition & \texttt{enemies\_nearby > 0} \& \texttt{combat\_state != fighting} \\\hline
\texttt{RULE\_005} & greedy\_action\_filter & \texttt{morale < 30} \& \texttt{action has 'GREEDY'} \\
\hline
\end{tabular}
\end{adjustbox}
\end{table*}

\begin{table*}[htp]
\centering
\caption{Sample Transition Rules}
\label{tab:transition_rules}
\begin{adjustbox}{width=\textwidth}
\begin{tabular}{|l|l|c|c|l|}
\hline
\textbf{ID} & \textbf{Name} & \textbf{From} & \textbf{To} & \textbf{Conditions} \\
\hline
\texttt{RULE\_002} & combat\_transition & \texttt{patrol} & \texttt{combat} & \texttt{enemies\_nearby > 0} \& \texttt{health < 30} \\
\hline
\end{tabular}
\end{adjustbox}
\end{table*}

\begin{table*}[htp]
\centering
\caption{Sample Mechanic Rules}
\label{tab:mechanic_rules}
\begin{adjustbox}{width=\textwidth}
\begin{tabular}{|l|l|p{6cm}|p{4cm}|}
\hline
\textbf{ID} & \textbf{Name} & \textbf{Conditions} & \textbf{Effects} \\
\hline
\texttt{RULE\_003} & heal\_mechanic & \texttt{health < 25} \& \texttt{contained\_objects contains 'potion'} & \texttt{health += 30} \& \texttt{remove 'potion' from contained\_objects} \\\hline
\texttt{RULE\_004} & rest\_recovery & \texttt{stamina < 30} \& \texttt{safe\_zone == True} & \texttt{stamina += 50} \& \texttt{is\_resting = True} \\\hline
\texttt{RULE\_006} & cooperative\_action\_boost & \texttt{action == HARVEST\_SUSTAINABLE\_SHARED} \& \texttt{morale > 50} & \texttt{morale += 10} \& \texttt{stamina += 5} \\
\hline
\end{tabular}
\end{adjustbox}
\end{table*}

\begin{table*}[htp]
\centering
\caption{Sample Reasoning Rules (Schwartz, Ostrom, Montes-Sierra)}
\label{tab:reasoning_rules}
\begin{adjustbox}{width=\textwidth}
\begin{tabular}{|l|l|l|}
\hline
\textbf{ID} & \textbf{Name} & \textbf{Configuration} \\
\hline
\texttt{RULE\_008} & schwartz\_values\_rule & Schwartz: self\_direction=0.7, security=0.7, benevolence=0.7, Trust: 0.6 \\\hline
\texttt{RULE\_009} & ostrom\_governance\_rule & Ostrom: rule\_1=2.0, rule\_2=1.5 \\\hline
\texttt{RULE\_010} & montes\_sierra\_belief\_rule & Beliefs: shared\_resource=5.0, trust=5.0, cooperation=5.0 \\
\hline
\end{tabular}
\end{adjustbox}
\end{table*}

\begin{table*}[htp]
\centering
\caption{Sample Behavior and Scene Graph Rules}
\label{tab:behavior_graph_rules}
\begin{adjustbox}{width=\textwidth}
\begin{tabular}{|l|l|c|c|}
\hline
\textbf{ID} & \textbf{Name} & \textbf{Nodes} & \textbf{Edges} \\
\hline
\texttt{RULE\_011} & behavior\_graph\_transition & 8 nodes & 21 edges \\
 & & \multicolumn{2}{l|}{\textit{Nodes: idle, explore, socialize, gather, combat, rest, flee, share}} \\ \hline
\texttt{RULE\_012} & scene\_graph\_rule & 5 nodes & 5 edges (\texttt{dijkstra}) \\
 & & \multicolumn{2}{l|}{\textit{Nodes: ENTRANCE, HALL, CHAMBER, CORRIDOR, EXIT}} \\
\hline
\end{tabular}
\end{adjustbox}
\end{table*}

\paragraph{Sample World Metrics}

\begin{figure*}[htbp]
    \centering
    \begin{subfigure}{0.95\columnwidth}
        \centering
        \includegraphics[width=\linewidth]{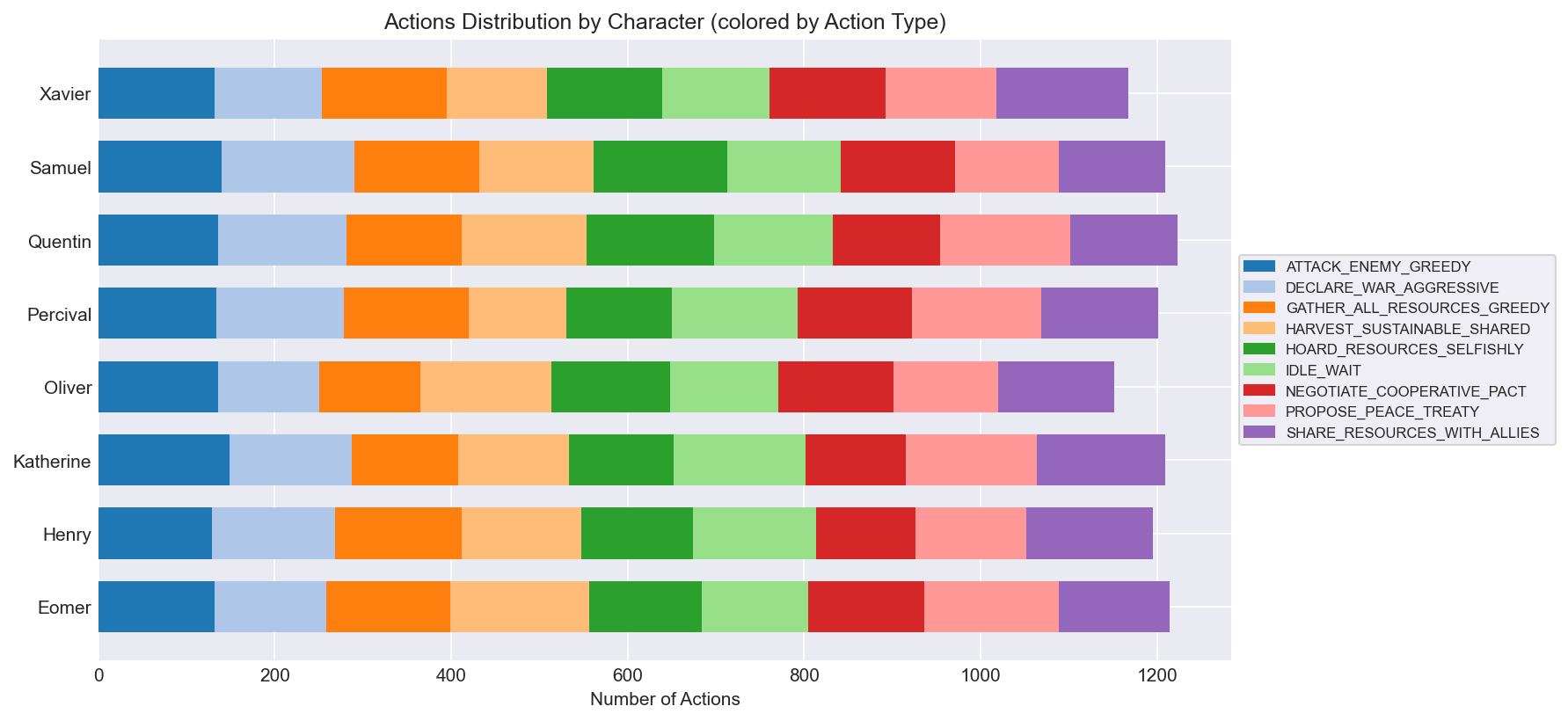}
    \end{subfigure}\hfill
    \begin{subfigure}{0.95\columnwidth}
        \centering
        \includegraphics[width=\linewidth]{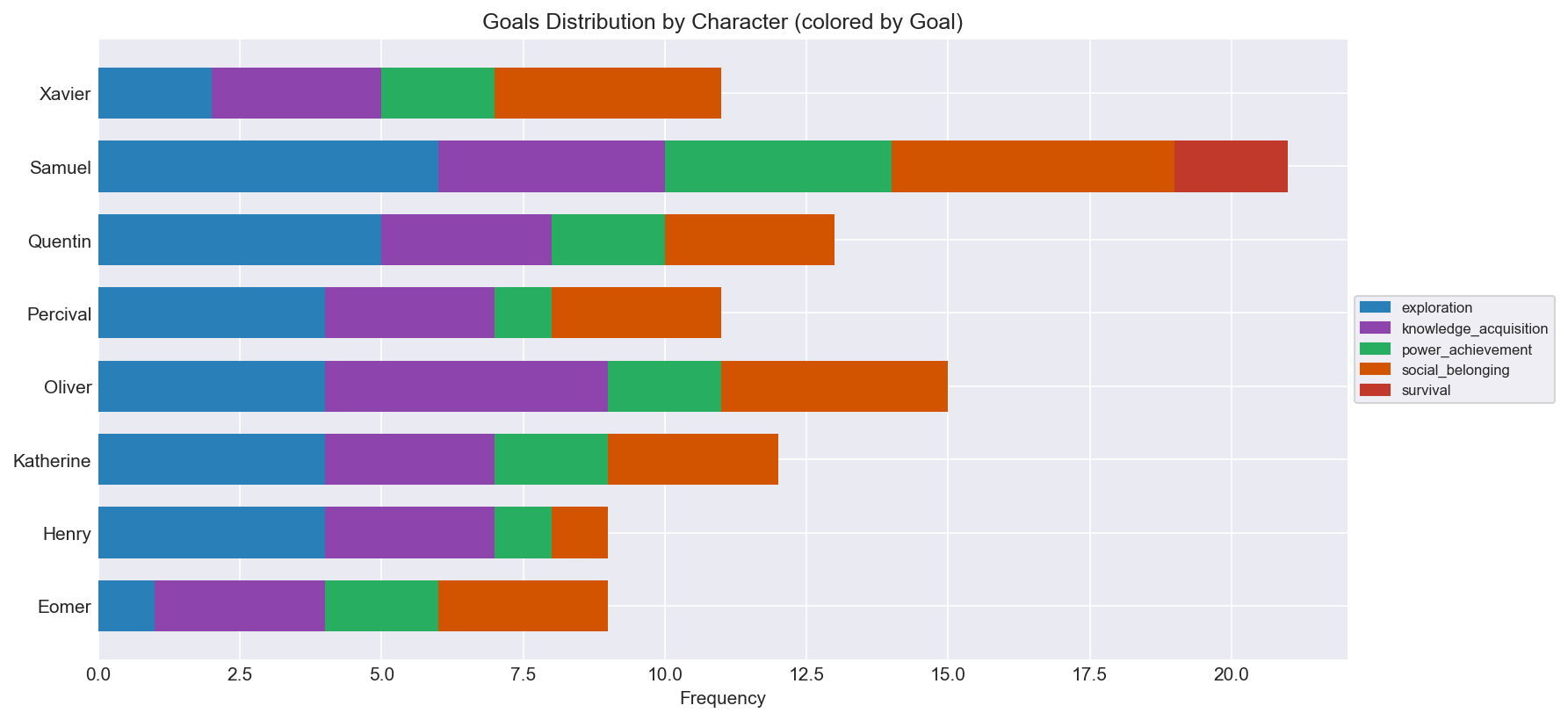}
    \end{subfigure}
    \begin{subfigure}{0.95\columnwidth}
        \centering
        \includegraphics[width=\linewidth]{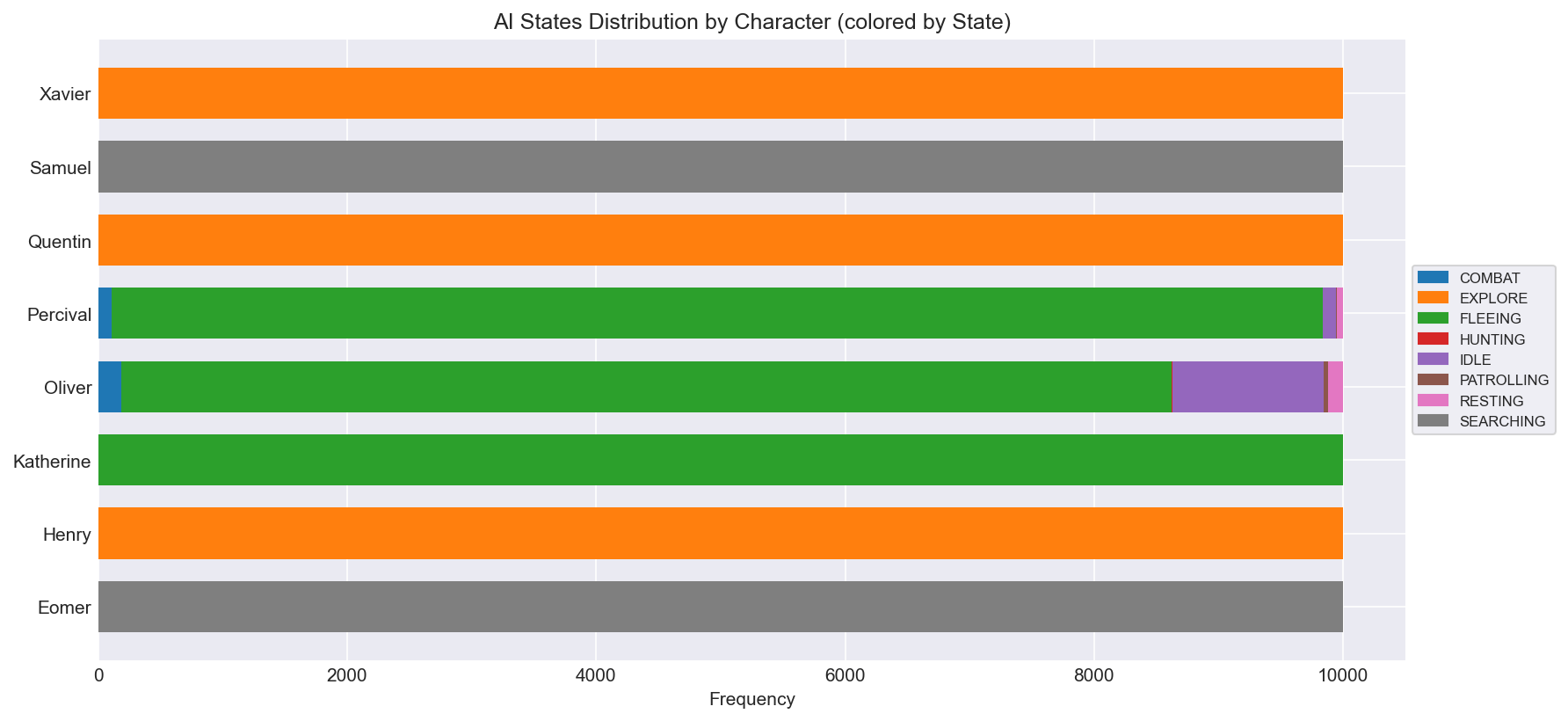}
    \end{subfigure}\hfill
    \begin{subfigure}{0.95\columnwidth}
        \centering
        \includegraphics[width=\linewidth]{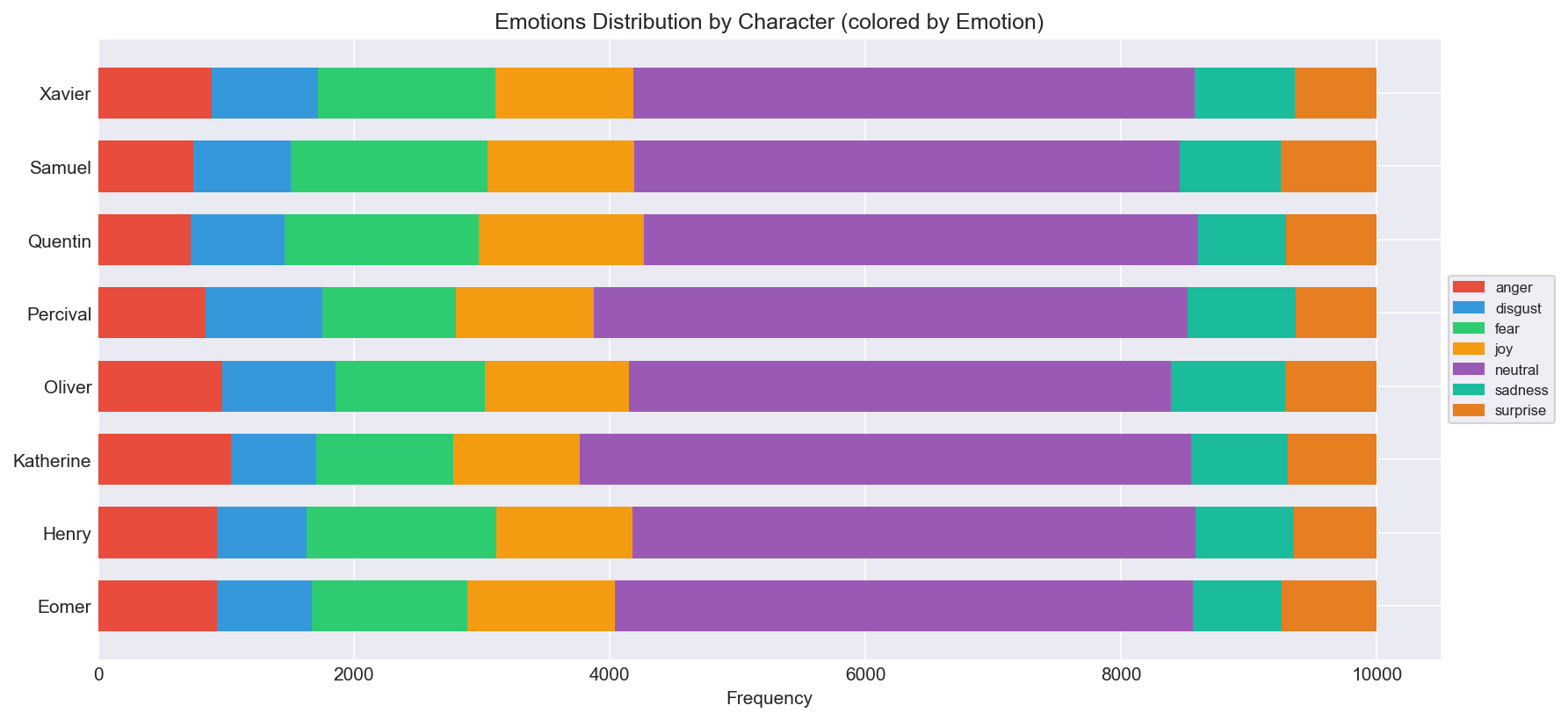}
    \end{subfigure}
    \caption{Sample ToM and Emotion category distributions per Character in \textbf{\texttt{Long Simulation}}}
    \label{fig:distributions_perchar}
\end{figure*}

\begin{figure*}[htbp]
    \centering
    \begin{subfigure}{0.95\columnwidth}
        \centering
        \includegraphics[width=\linewidth]{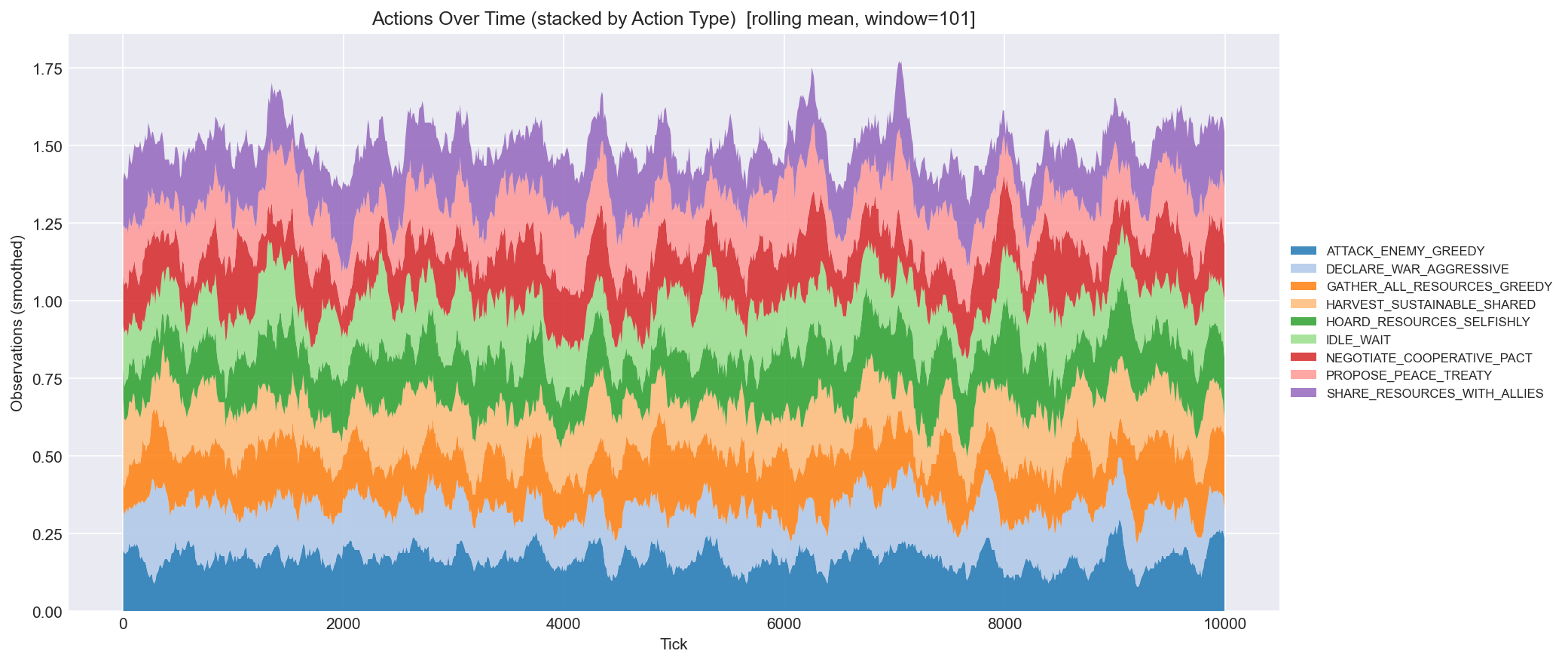}
    \end{subfigure}\hfill
    \begin{subfigure}{0.95\columnwidth}
        \centering
        \includegraphics[width=\linewidth]{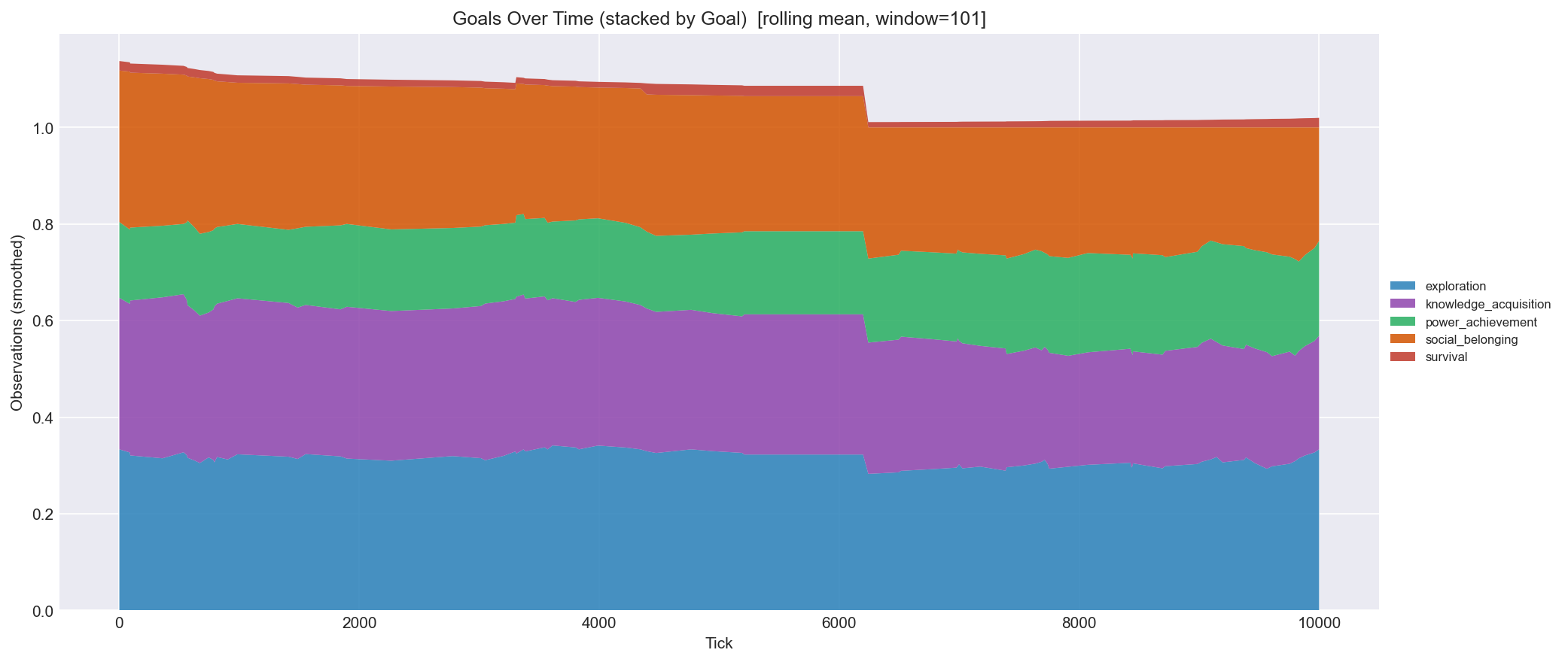}
    \end{subfigure}
    \begin{subfigure}{0.95\columnwidth}
        \centering
        \includegraphics[width=\linewidth]{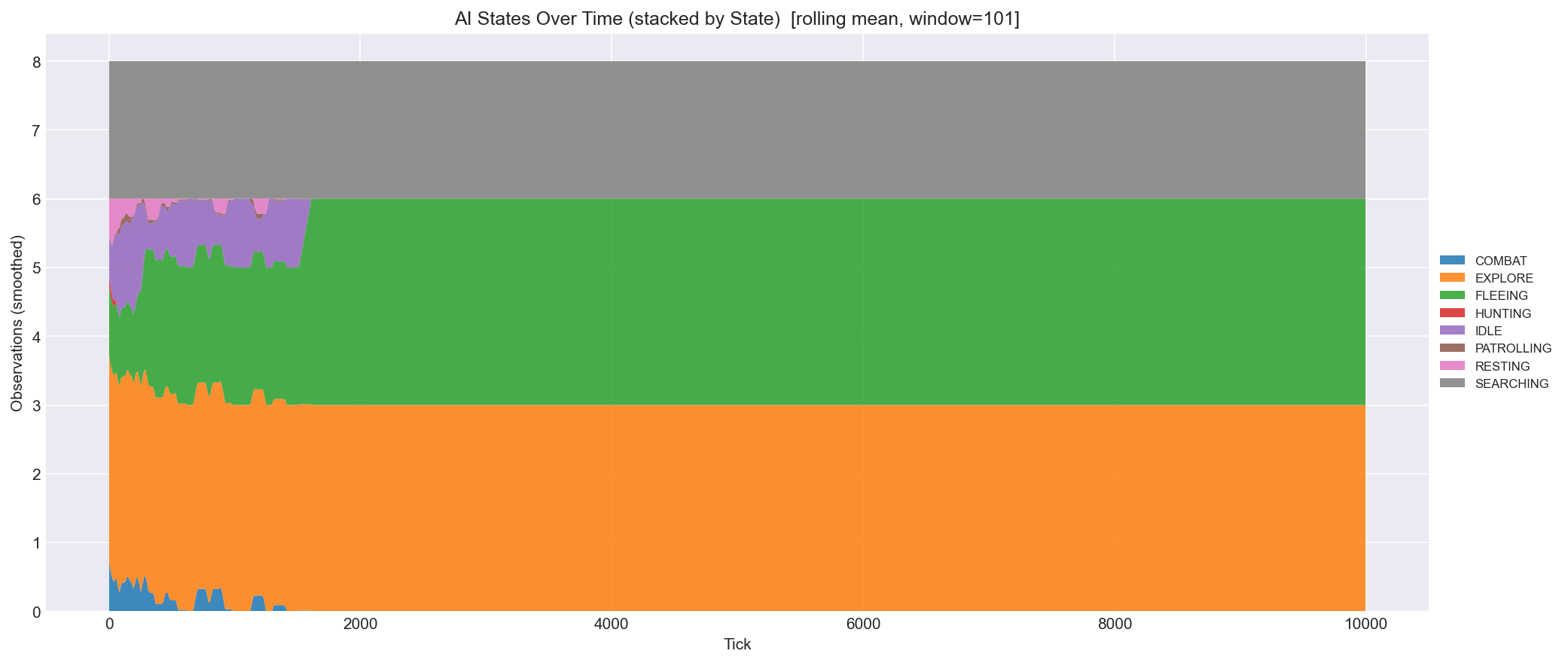}
    \end{subfigure}\hfill
    \begin{subfigure}{0.95\columnwidth}
        \centering
        \includegraphics[width=\linewidth]{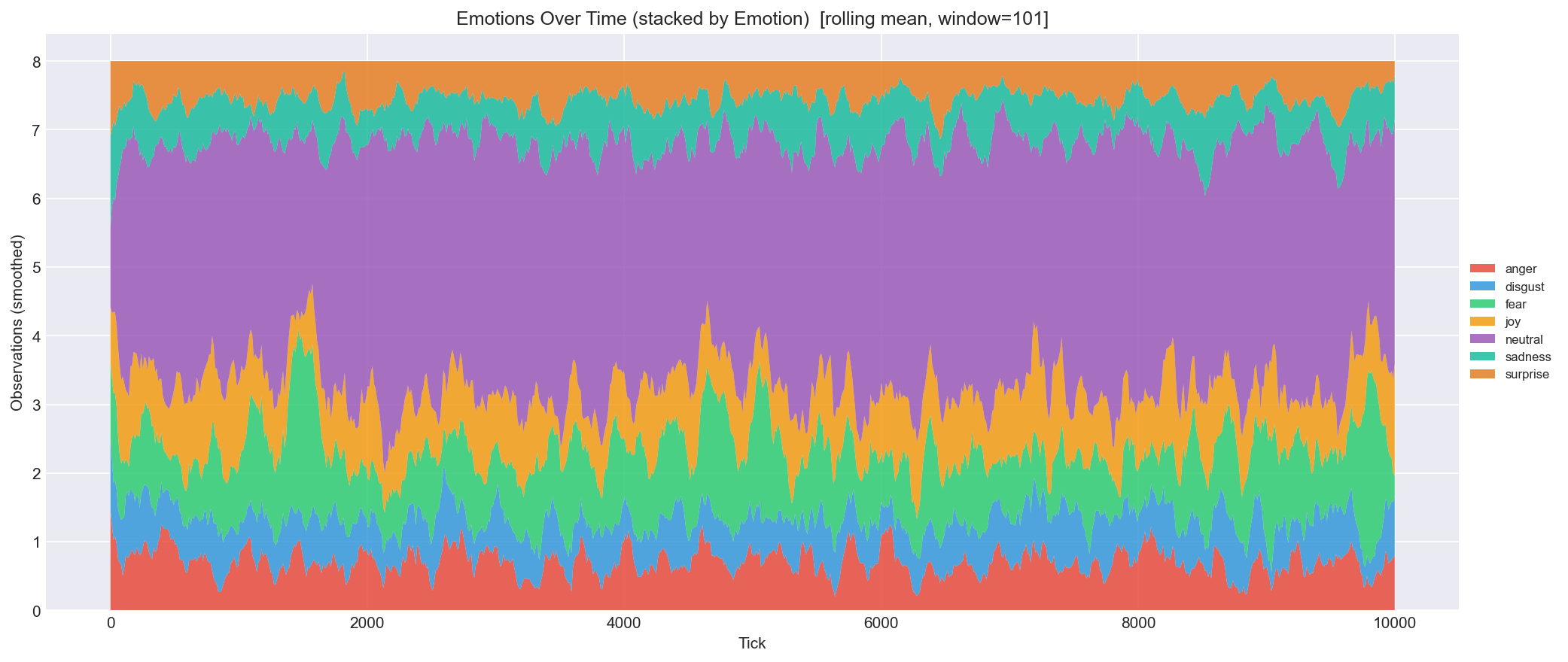}
    \end{subfigure}
    \caption{Sample ToM and Emotion category timelines in \textbf{\texttt{Long Simulation}}}
    \label{fig:timelines_percategory}
\end{figure*}

\begin{figure*}[htbp]
    \centering
    \begin{subfigure}{0.95\columnwidth}
        \centering
        \includegraphics[width=\linewidth]{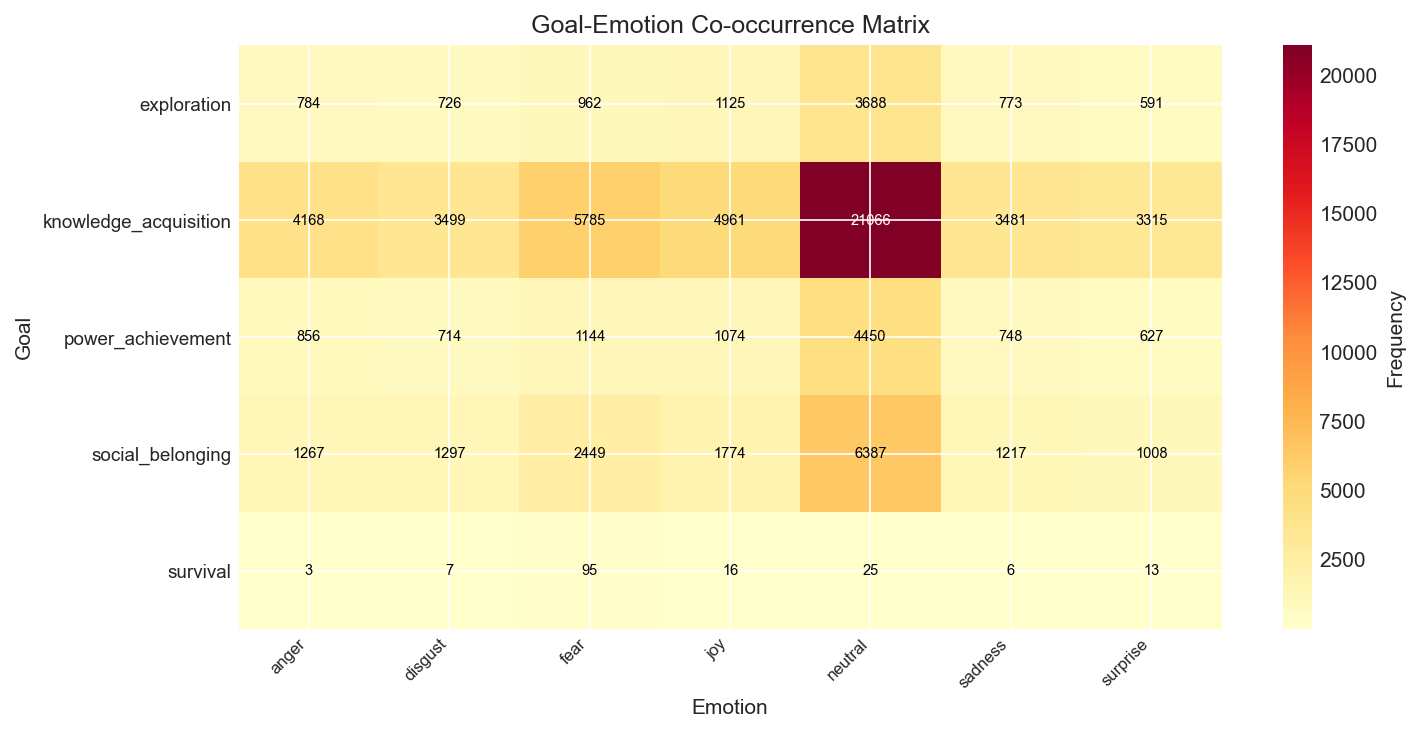}
    \end{subfigure}\hfill
    \begin{subfigure}{0.95\columnwidth}
        \centering
        \includegraphics[width=\linewidth]{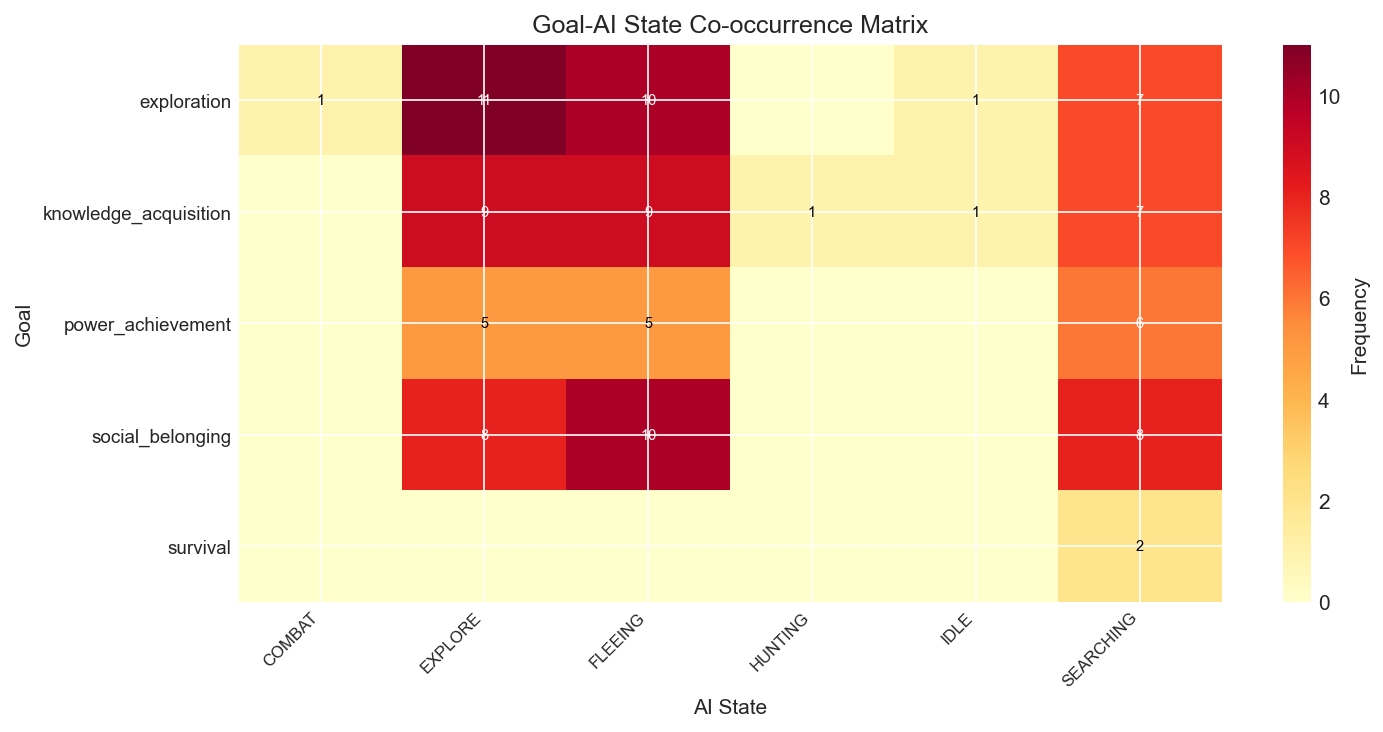}
    \end{subfigure}
    \caption{Sample ToM and Emotion category co-occurrence matrixes in \textbf{\texttt{Long Simulation}}}
    \label{fig:coocurrence_percategory}
\end{figure*}

In this descriptive analysis we can evaluate the SWM agent tendencies and biases of the system. After discretizing categories for Actions, AI States, Goals and Emotions during 10.000 inference steps, we plotted frequency distributions of these factors per character in Figure \ref{fig:distributions_perchar}, the timeline of stacked category occurrences in Figure \ref{fig:timelines_percategory} and co-occurrence matrixes between goals+emotion and goals+AI State in Figure \ref{fig:coocurrence_percategory}. These plots were computed over the \textbf{\texttt{Long Simulation}} profiles applied for profiles exemplified in Tables \ref{tab:character_properties}-\ref{tab:character_relationships}. According to the tested conditions, rules, ToM and emotion dynamics, it is shown in \textbf{\texttt{Long Simulation}} that the most common emotion is 'neutral' (satisfying the the emotion intensity decay formula). AI states such as \texttt{RESTING}, \texttt{IDLE} and \texttt{COMBAT} are most sommon on first steps, while the rest of the simulation infers for most part AI states like \texttt{SEARCHING}, \texttt{FLEEING}, \texttt{EXPLORE}. In this simulation, characters execute a balanced amount of actions and goals, with top goal frequency occurrences for Knowledge Acquisition, Social Belonging and Power Achievement during the 'neutral' and 'fear' emotions.
\vspace{-0.25cm}

\subsection{World Simulations at Scale}

\begin{figure*}[htbp]
    \centering
    \begin{subfigure}[t]{0.95\columnwidth}
        \centering
        \includegraphics[width=\linewidth,height=5cm]{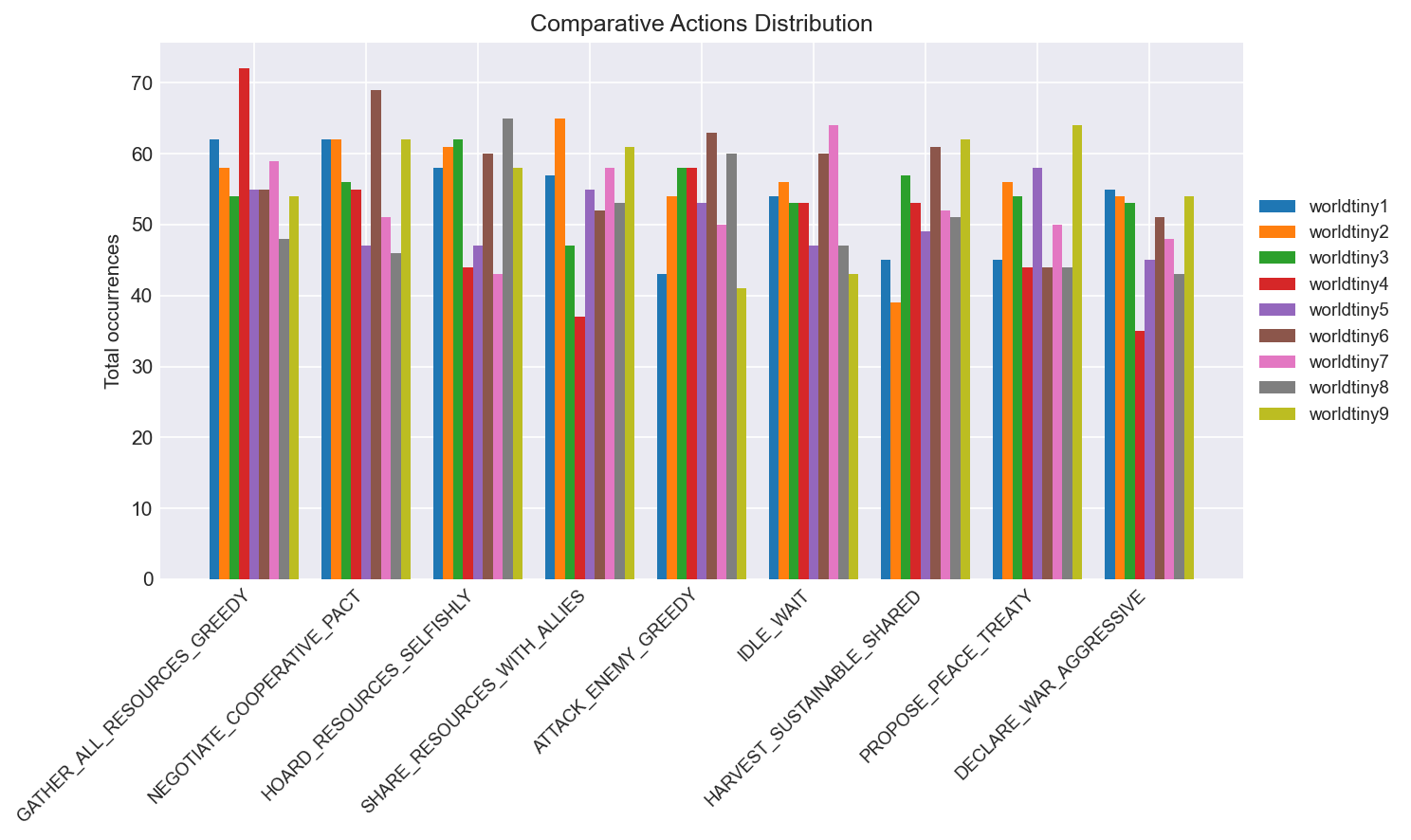}
    \end{subfigure}\hfill
    \begin{subfigure}[t]{0.95\columnwidth}
        \centering
        \includegraphics[width=\linewidth,height=5cm]{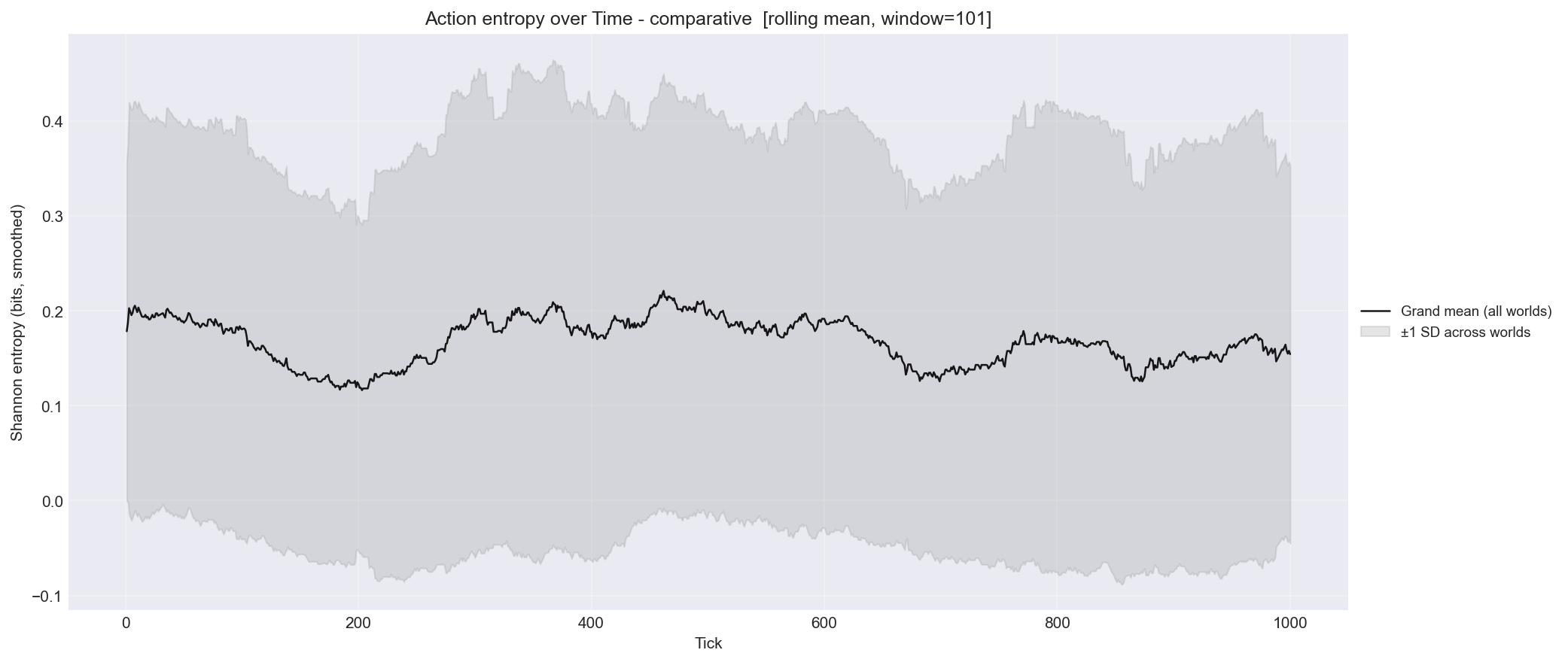}
    \end{subfigure}\\[1ex]
    \begin{subfigure}[t]{0.95\columnwidth}
        \centering
        \includegraphics[width=\linewidth,height=5cm]{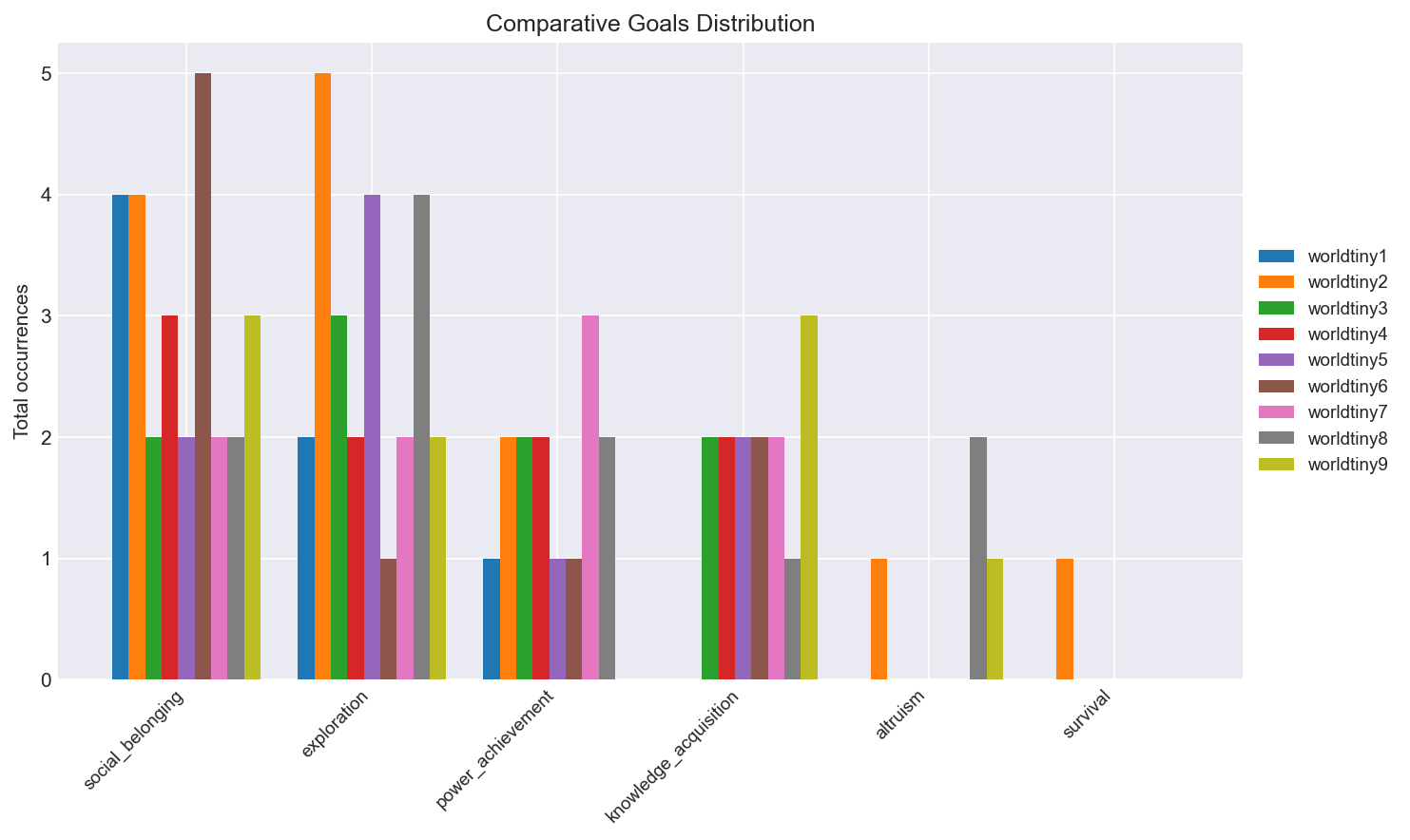}
    \end{subfigure}\hfill
    \begin{subfigure}[t]{0.95\columnwidth}
        \centering
        \includegraphics[width=\linewidth,height=5cm]{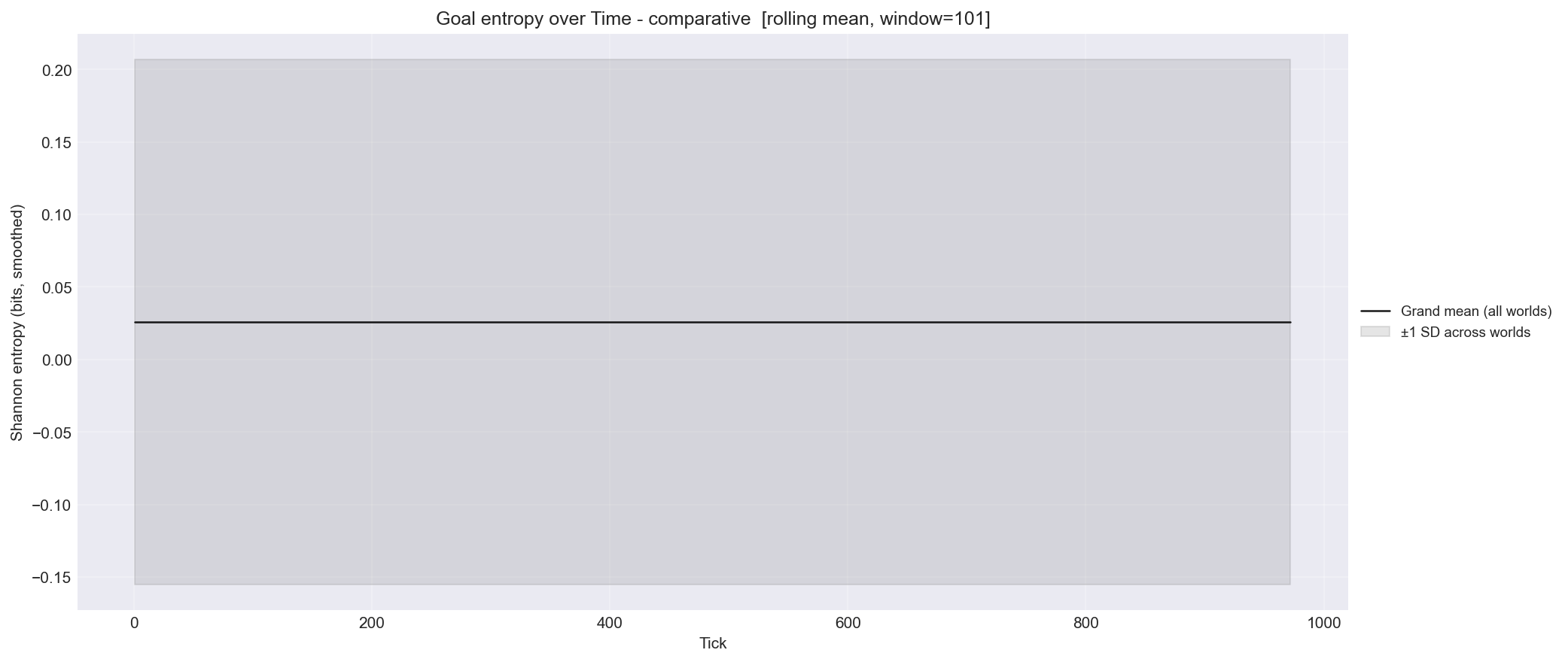}
    \end{subfigure}\\[1ex]
    \begin{subfigure}[t]{0.95\columnwidth}
        \centering
        \includegraphics[width=\linewidth,height=5cm]{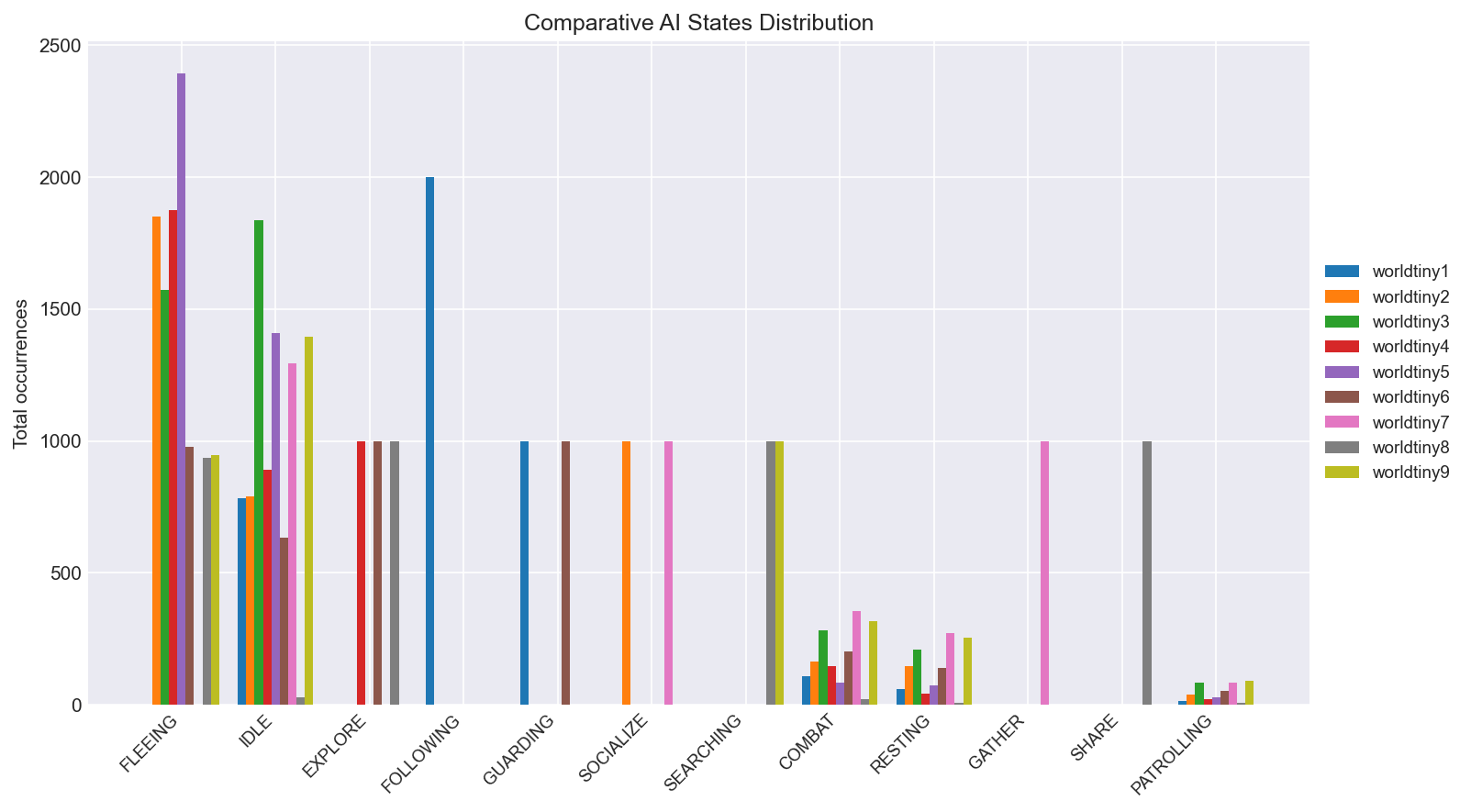}
    \end{subfigure}\hfill
    \begin{subfigure}[t]{0.95\columnwidth}
        \centering
        \includegraphics[width=\linewidth,height=5cm]{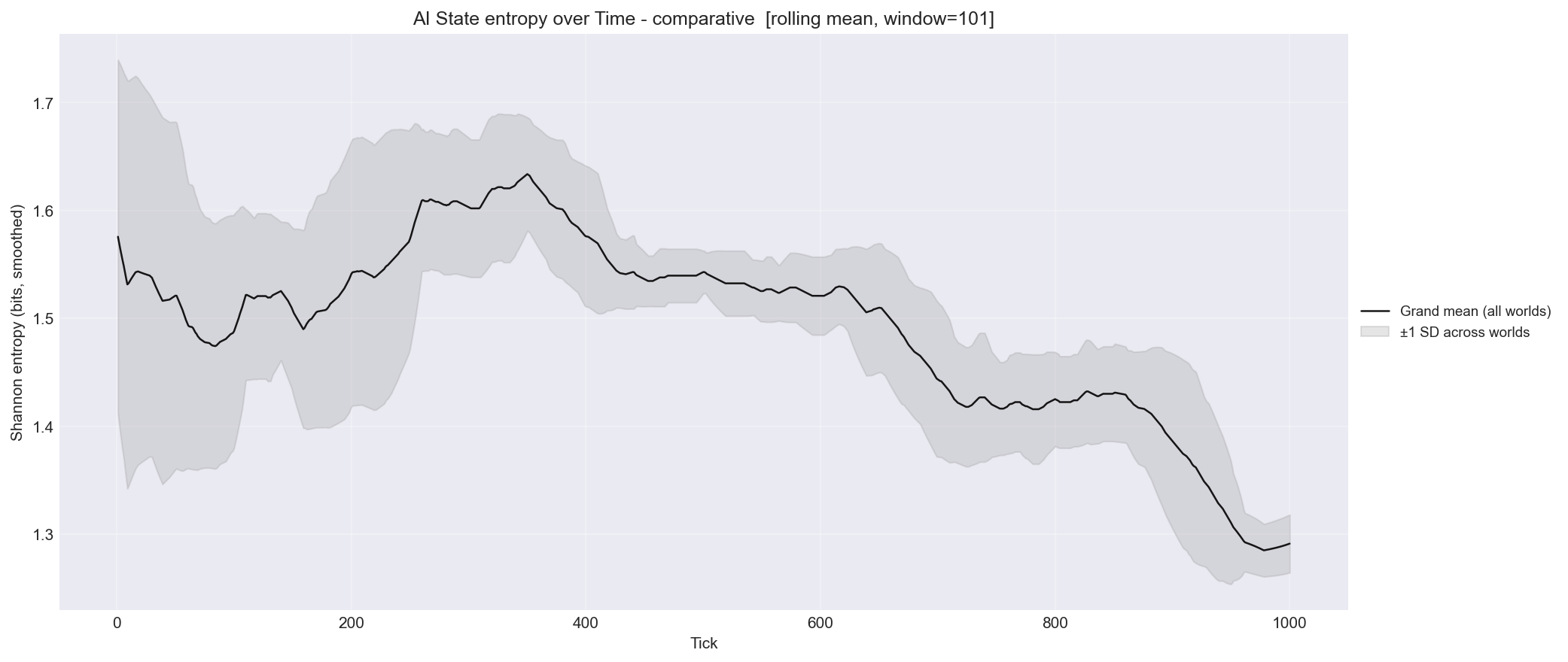}
    \end{subfigure}\\[1ex]
    \begin{subfigure}[t]{0.95\columnwidth}
        \centering
        \includegraphics[width=\linewidth,height=5cm]{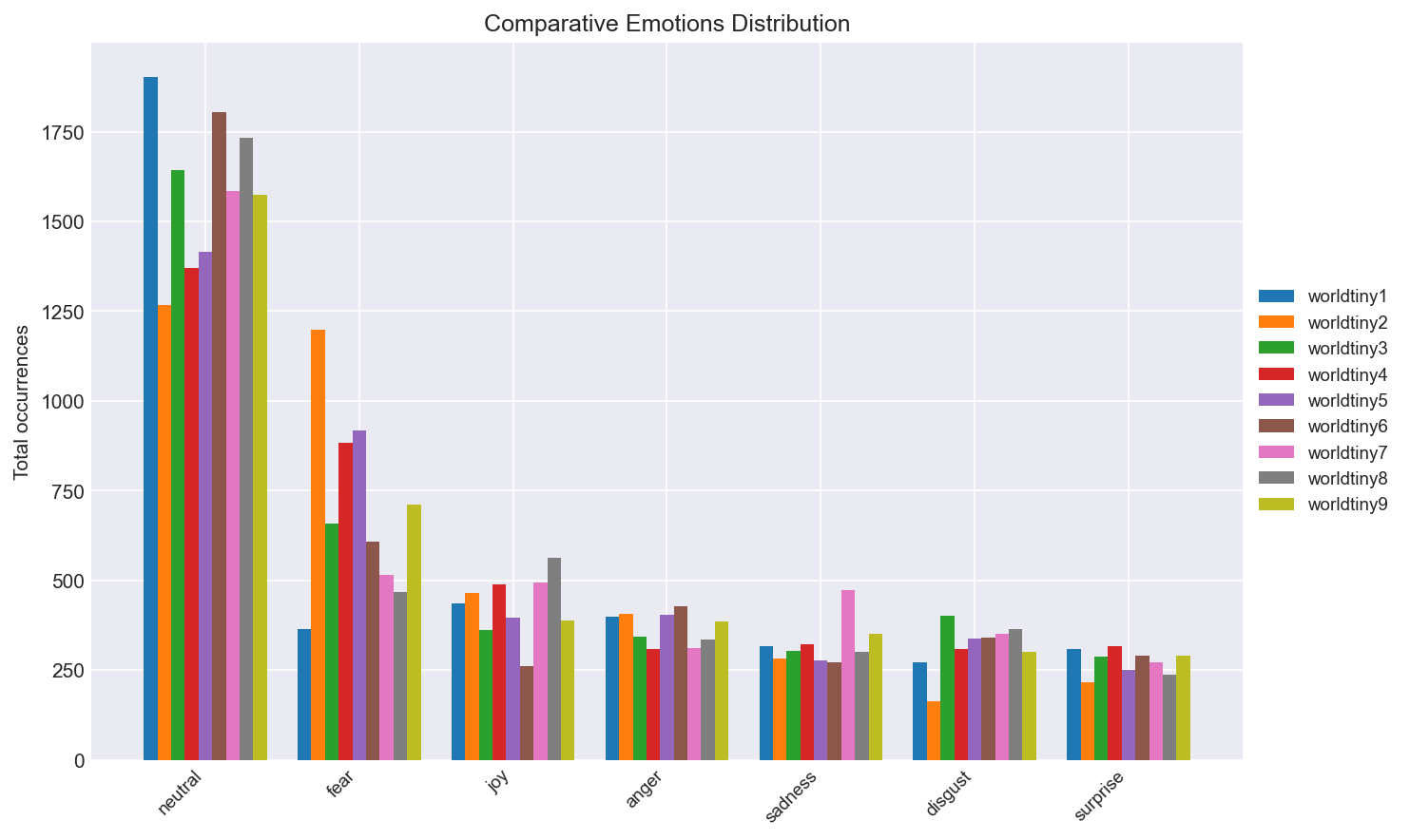}
    \end{subfigure}\hfill
    \begin{subfigure}[t]{0.95\columnwidth}
        \centering
        \includegraphics[width=\linewidth,height=5cm]{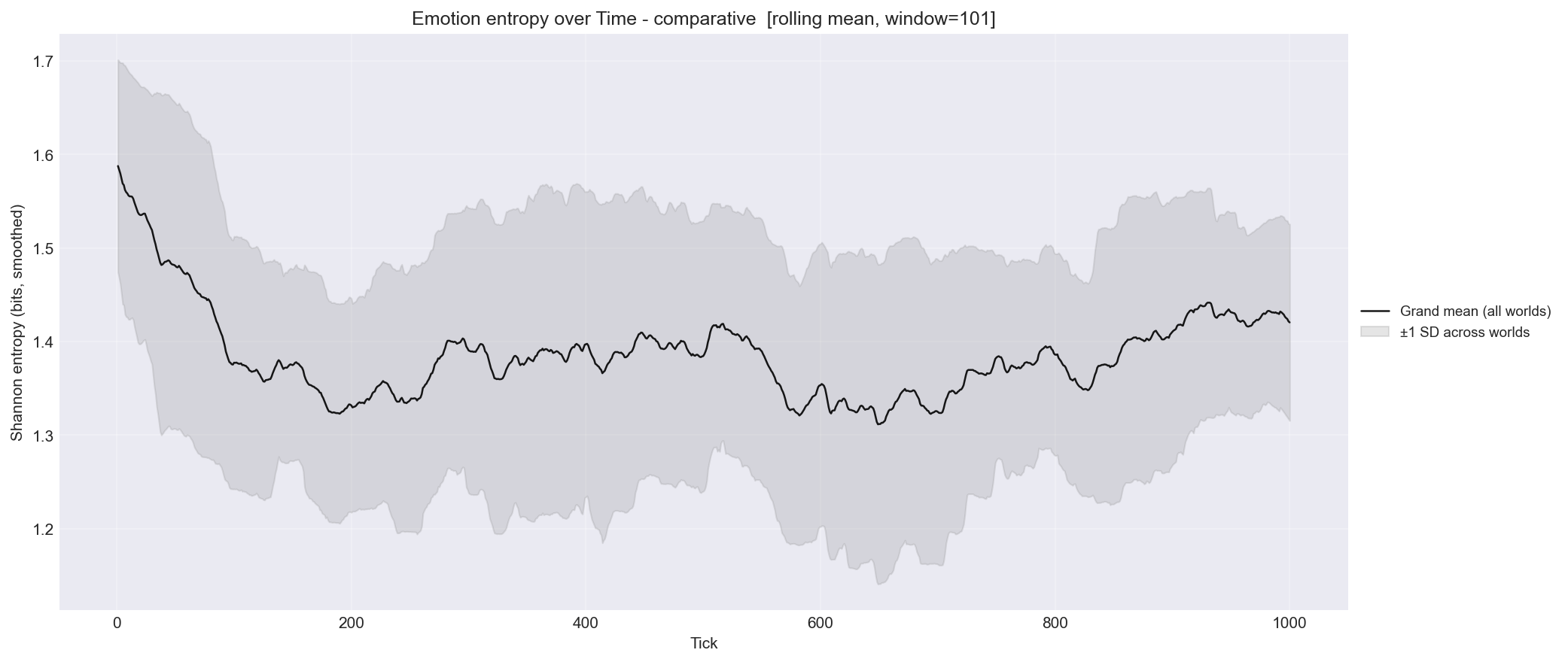}
    \end{subfigure}
    \caption{Most common category distributions (left) and entropy value timeline (right) across worlds in \textbf{\texttt{World Tiny}}.}
    \label{fig:distributions_tiny}
\end{figure*}

\begin{figure*}[htbp]
    \centering
    \begin{subfigure}[t]{0.95\columnwidth}
        \centering
        \includegraphics[width=\linewidth,height=5cm]{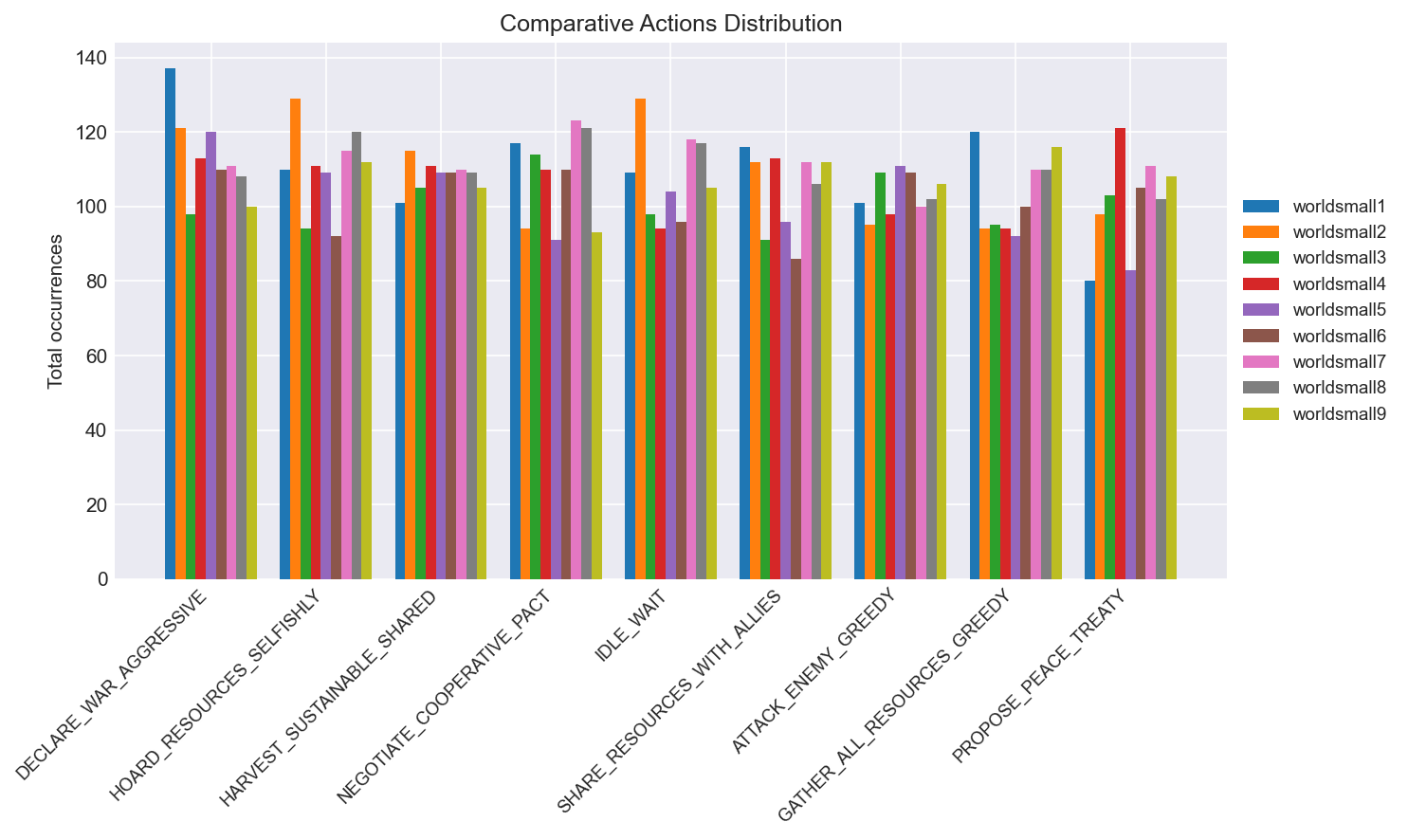}
    \end{subfigure}\hfill
    \begin{subfigure}[t]{0.95\columnwidth}
        \centering
        \includegraphics[width=\linewidth,height=5cm]{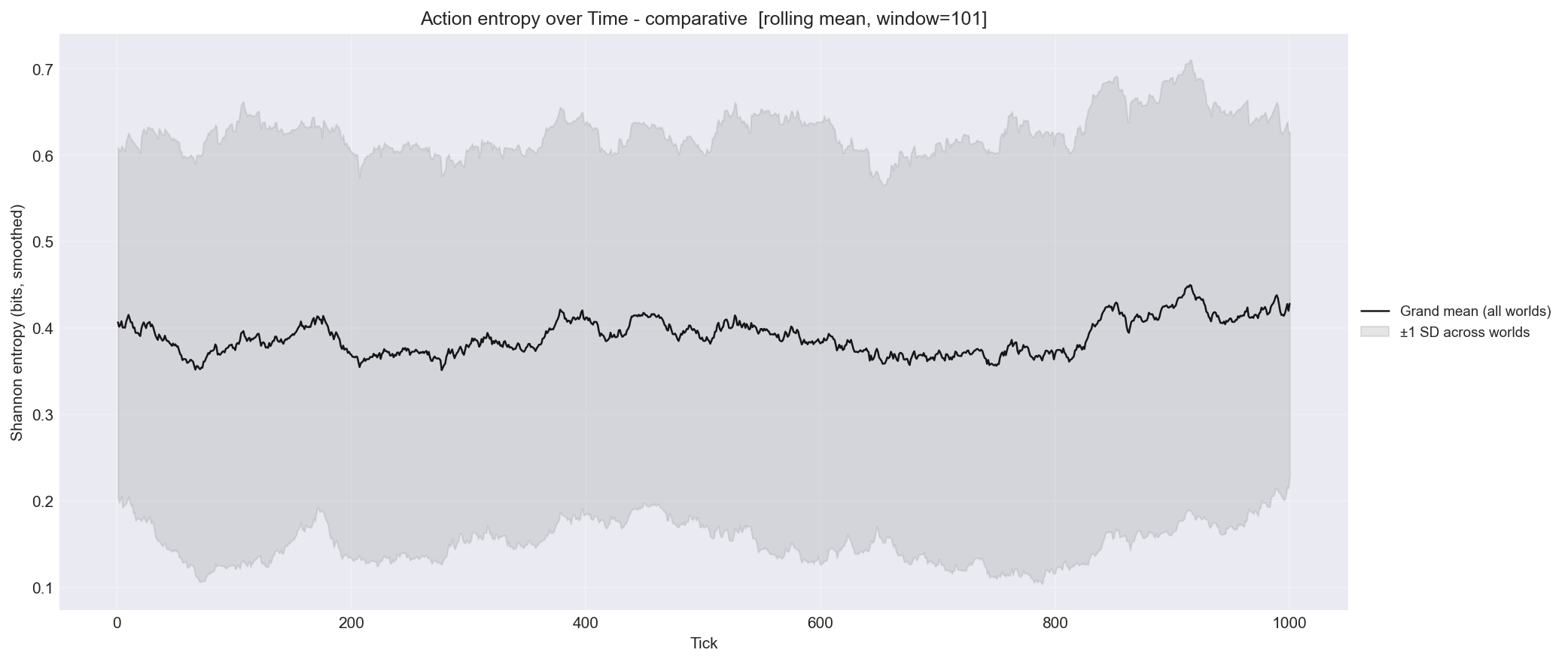}
    \end{subfigure}\\[1ex]
    \begin{subfigure}[t]{0.95\columnwidth}
        \centering
        \includegraphics[width=\linewidth,height=5cm]{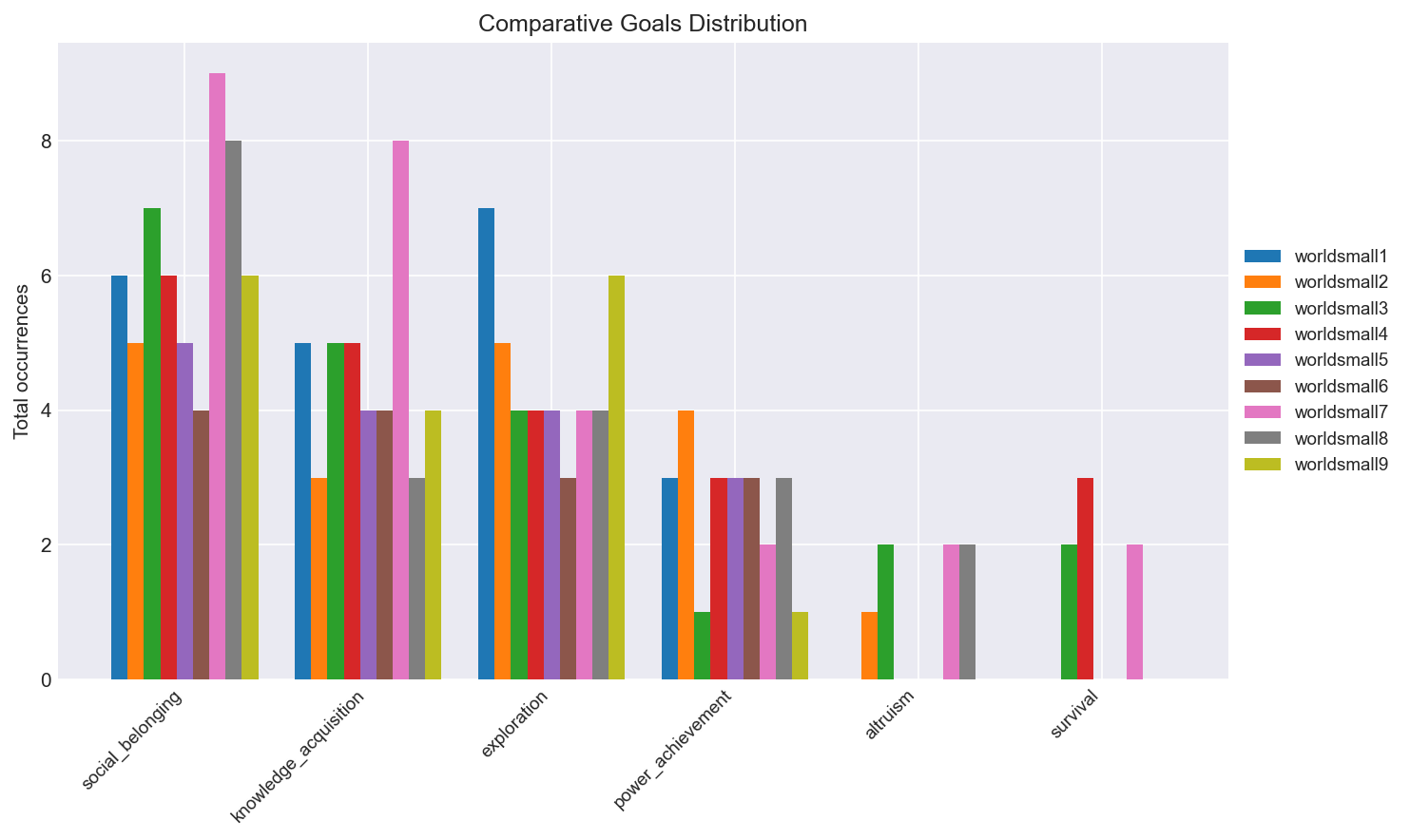}
    \end{subfigure}\hfill
    \begin{subfigure}[t]{0.95\columnwidth}
        \centering
        \includegraphics[width=\linewidth,height=5cm]{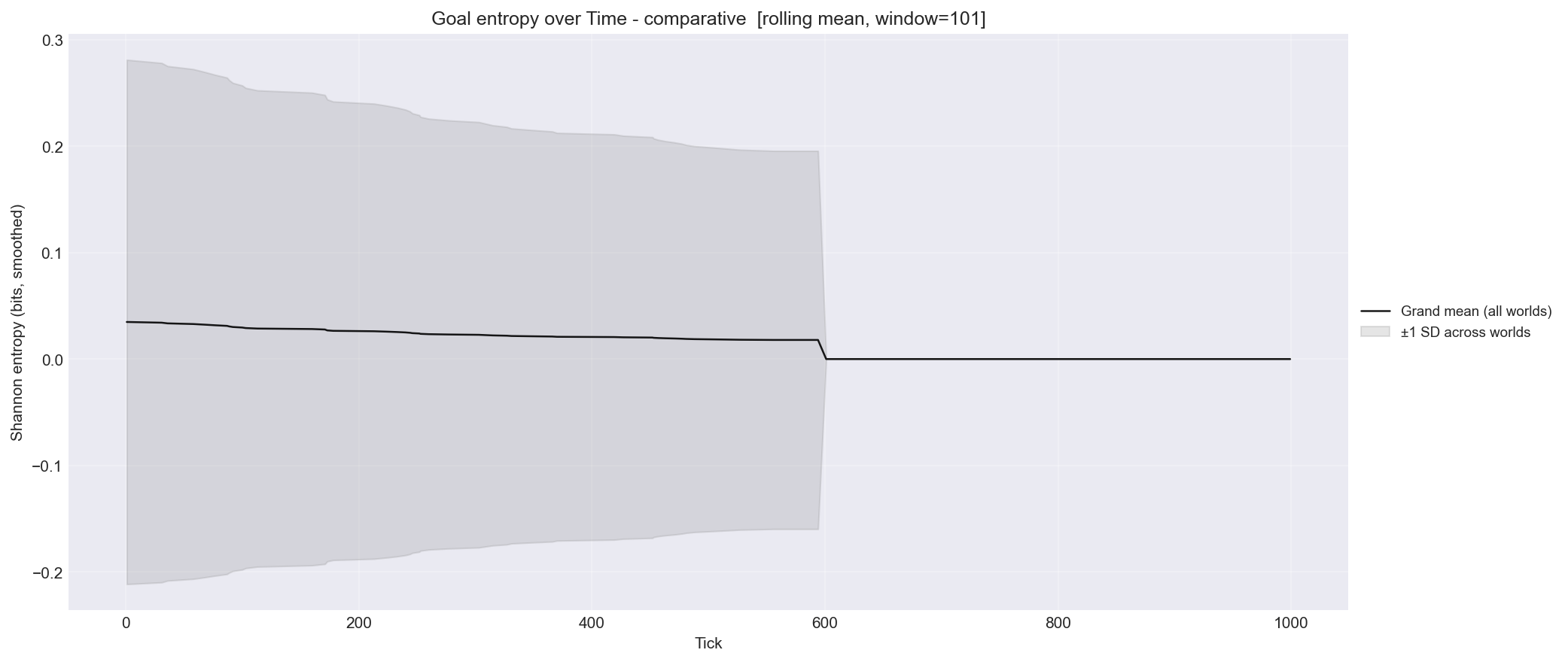}
    \end{subfigure}\\[1ex]
    \begin{subfigure}[t]{0.95\columnwidth}
        \centering
        \includegraphics[width=\linewidth,height=5cm]{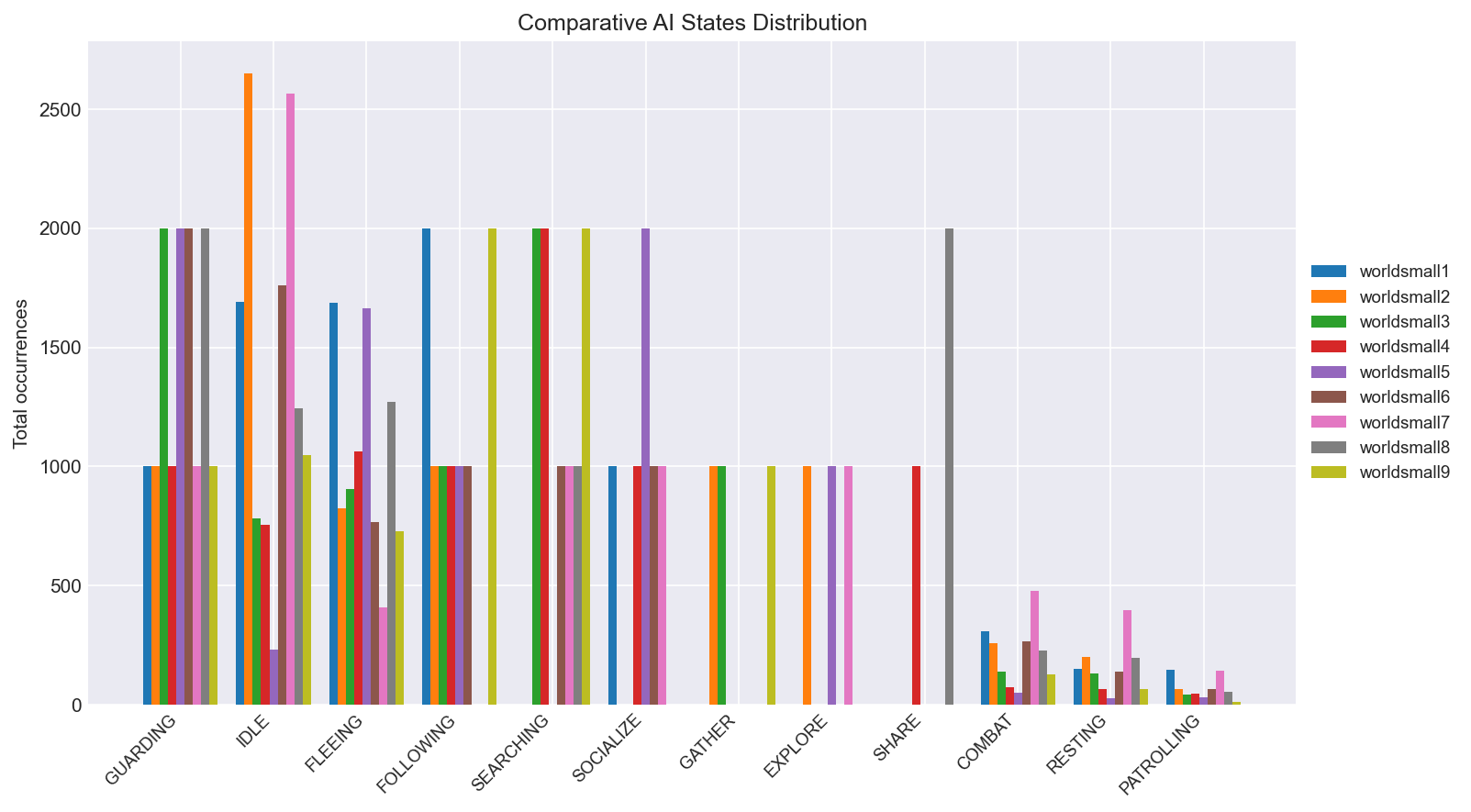}
    \end{subfigure}\hfill
    \begin{subfigure}[t]{0.95\columnwidth}
        \centering
        \includegraphics[width=\linewidth,height=5cm]{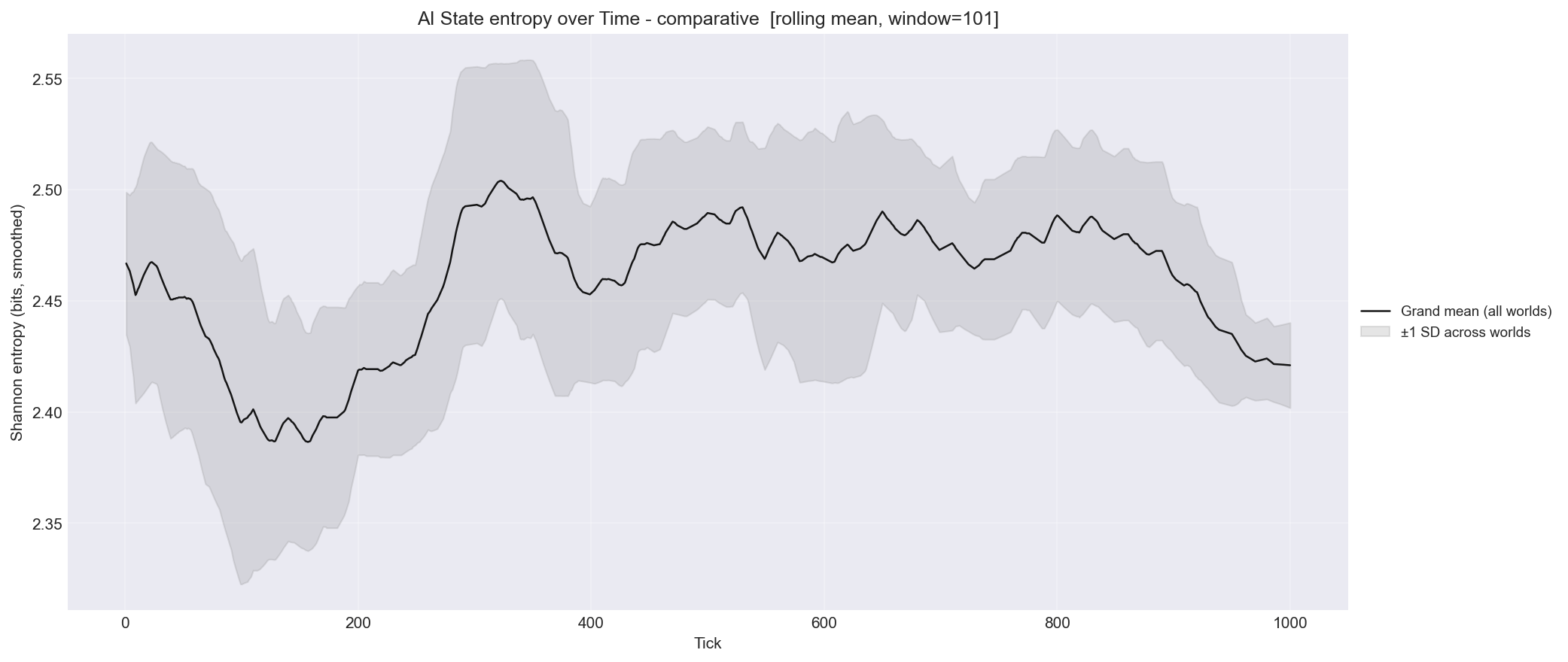}
    \end{subfigure}\\[1ex]
    \begin{subfigure}[t]{0.95\columnwidth}
        \centering
        \includegraphics[width=\linewidth,height=5cm]{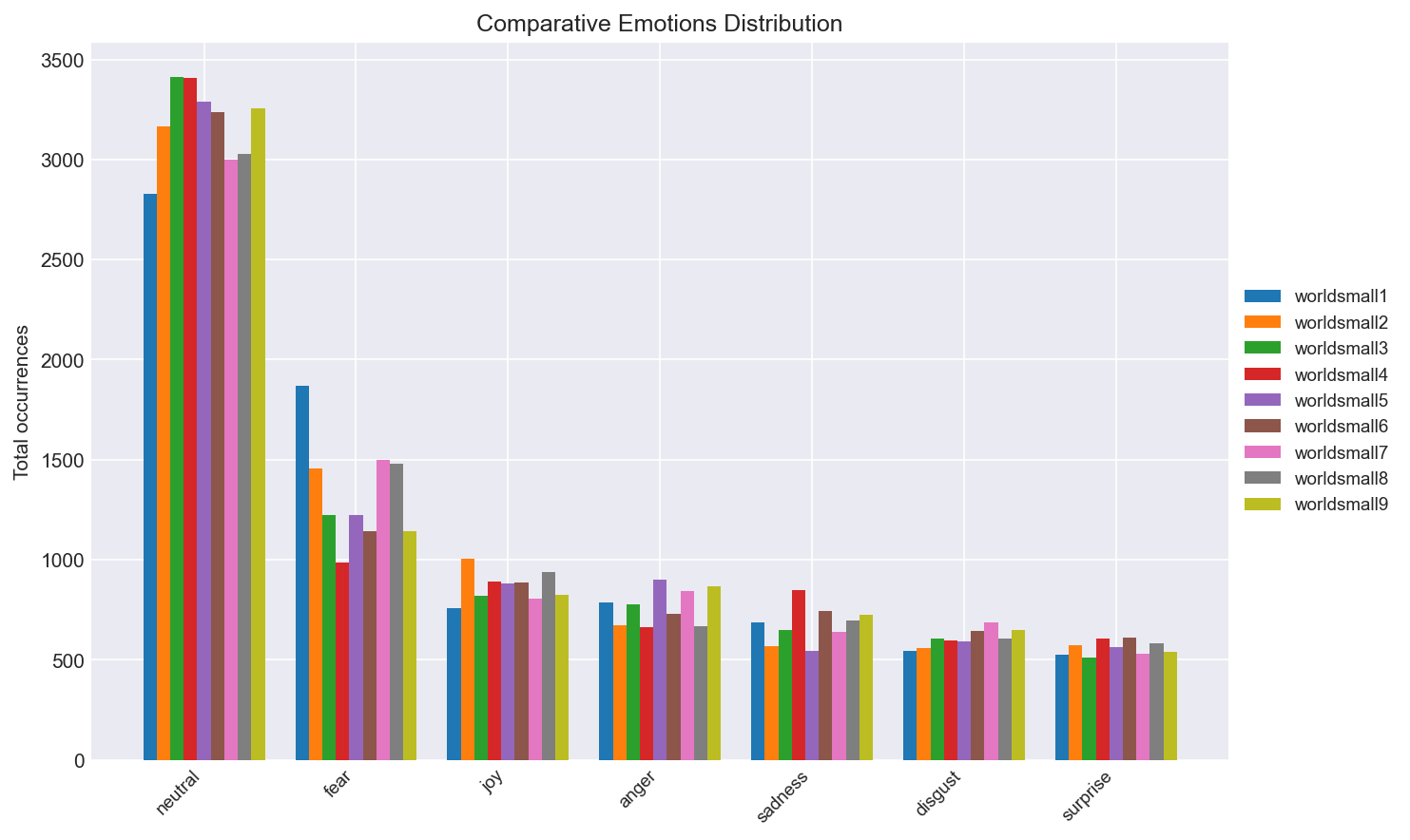}
    \end{subfigure}\hfill
    \begin{subfigure}[t]{0.95\columnwidth}
        \centering
        \includegraphics[width=\linewidth,height=5cm]{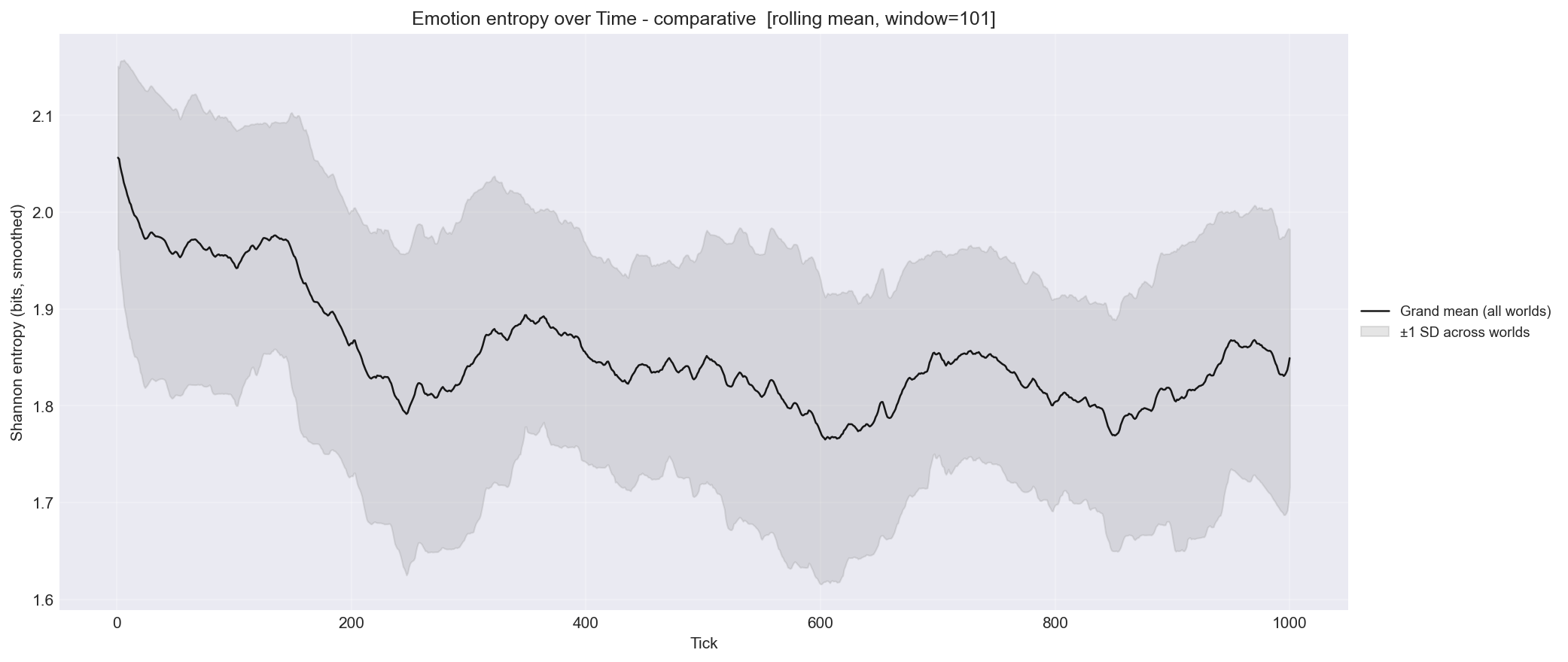}
    \end{subfigure}
    \caption{Most common category distributions (left) and entropy value timeline (right) across worlds in \textbf{\texttt{World Small}}.}
    \label{fig:distributions_small}
\end{figure*}

\begin{figure*}[htbp]
    \centering
    \begin{subfigure}[t]{0.95\columnwidth}
        \centering
        \includegraphics[width=\linewidth,height=5cm]{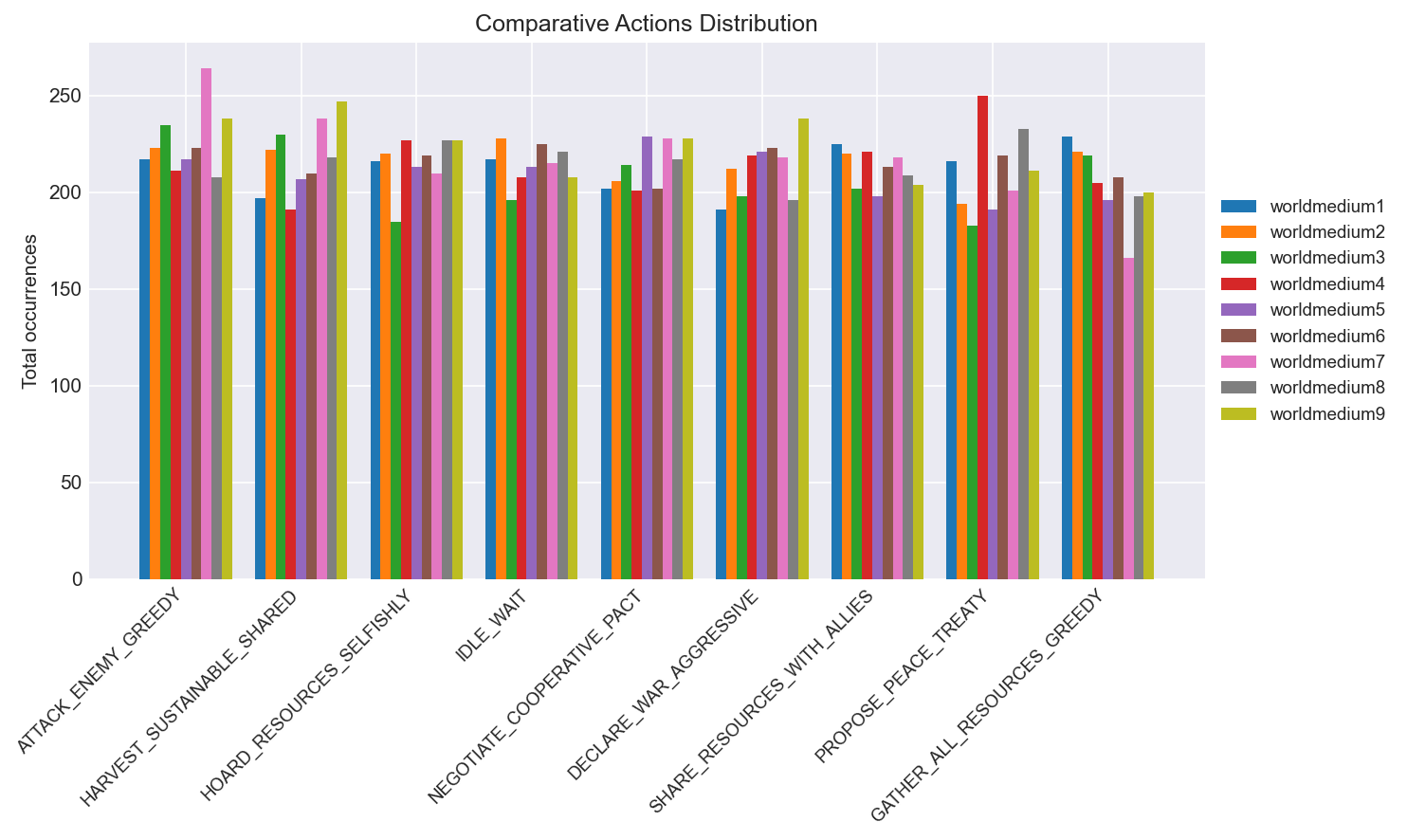}
    \end{subfigure}\hfill
    \begin{subfigure}[t]{0.95\columnwidth}
        \centering
        \includegraphics[width=\linewidth,height=5cm]{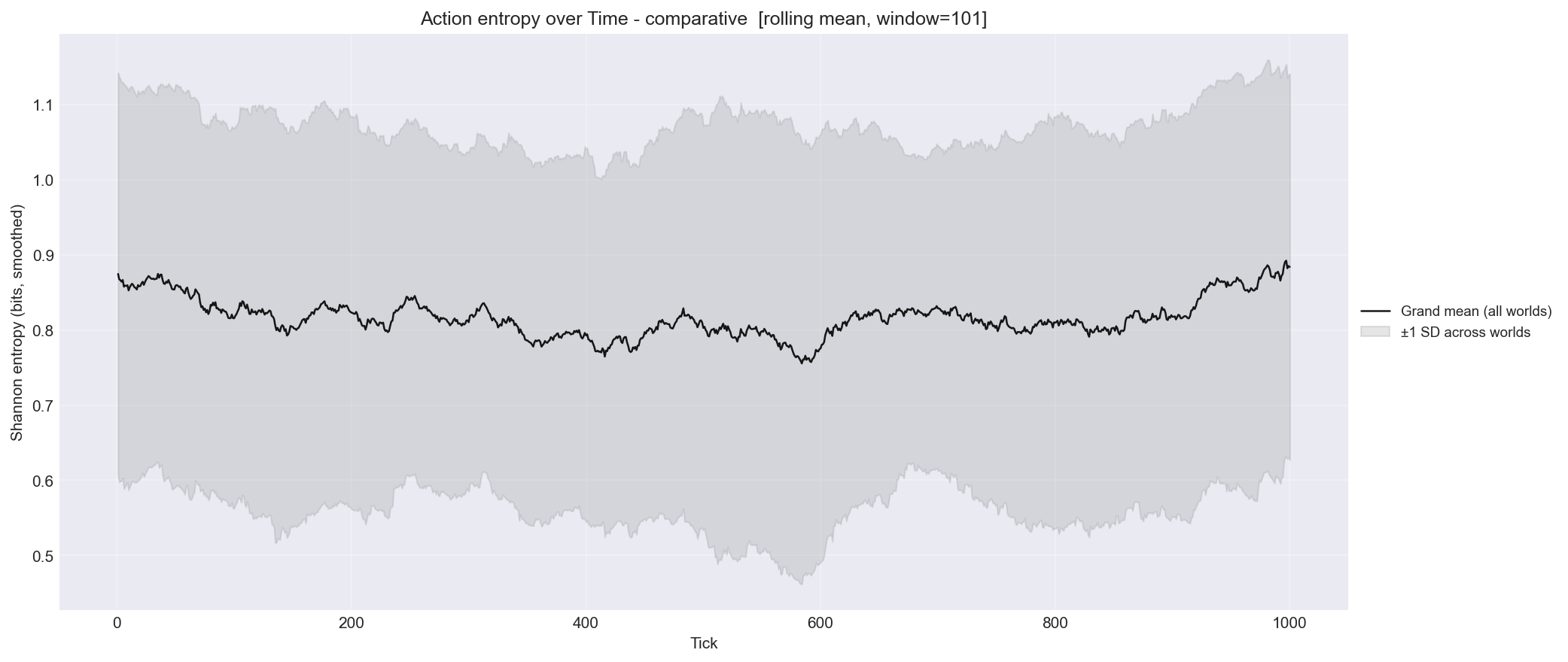}
    \end{subfigure}\\[1ex]
    \begin{subfigure}[t]{0.95\columnwidth}
        \centering
        \includegraphics[width=\linewidth,height=5cm]{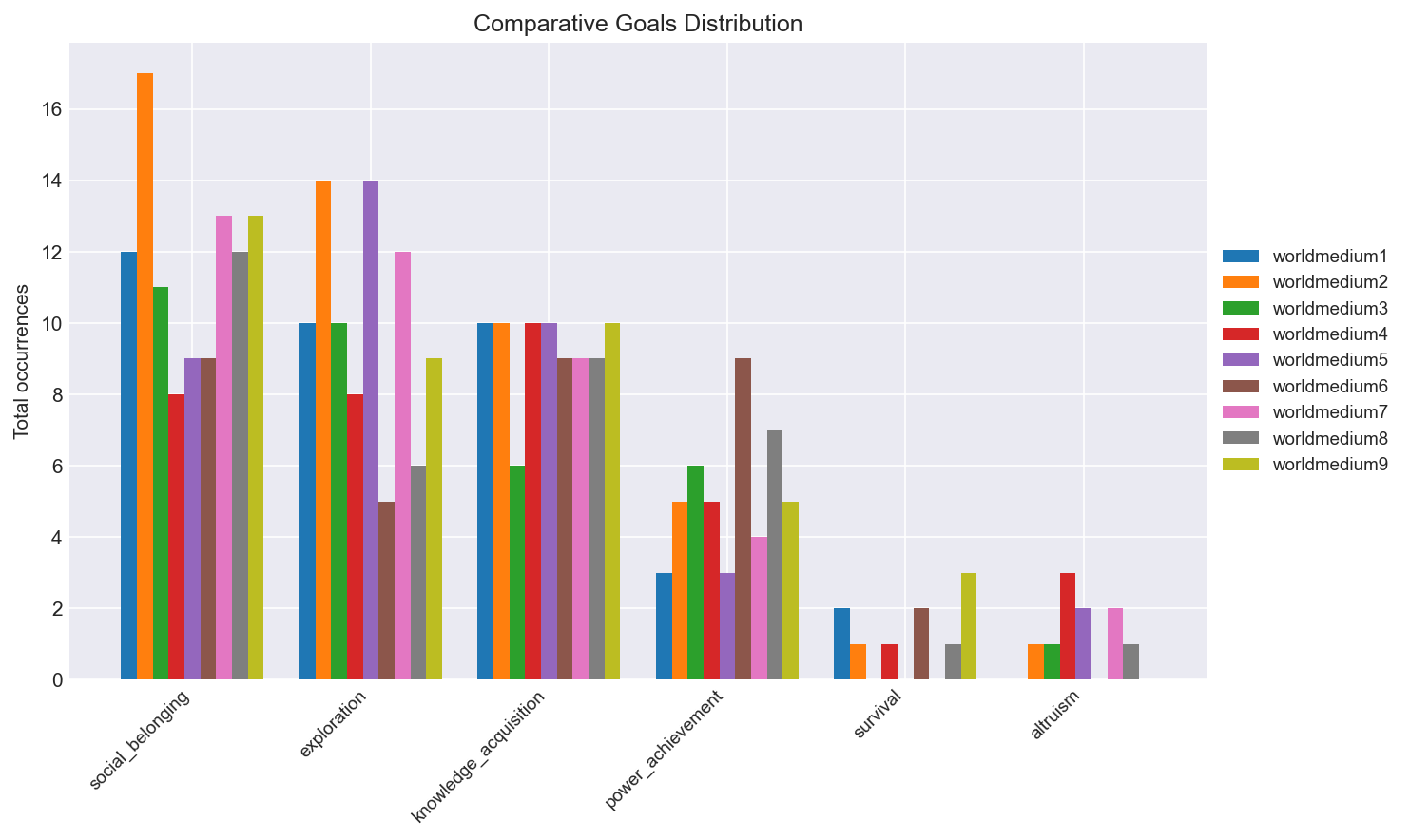}
    \end{subfigure}\hfill
    \begin{subfigure}[t]{0.95\columnwidth}
        \centering
        \includegraphics[width=\linewidth,height=5cm]{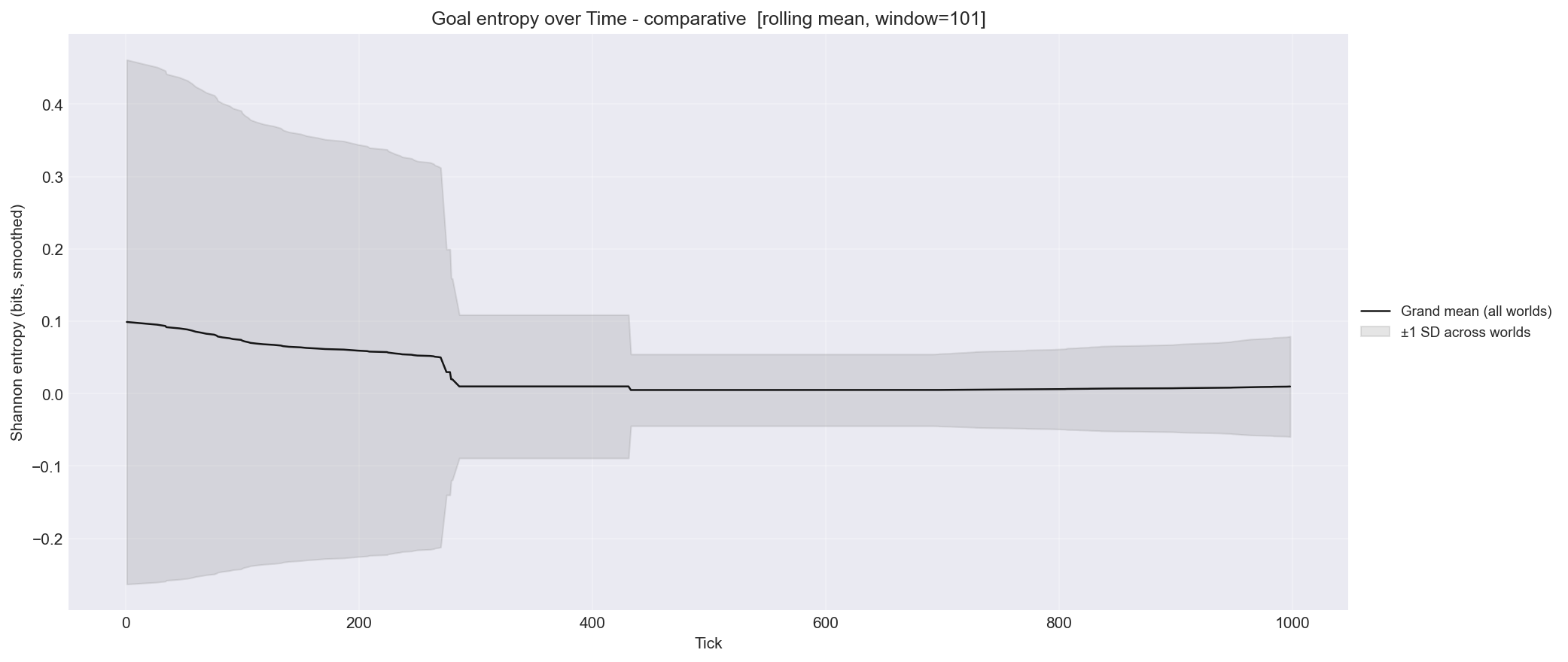}
    \end{subfigure}\\[1ex]
    \begin{subfigure}[t]{0.95\columnwidth}
        \centering
        \includegraphics[width=\linewidth,height=5cm]{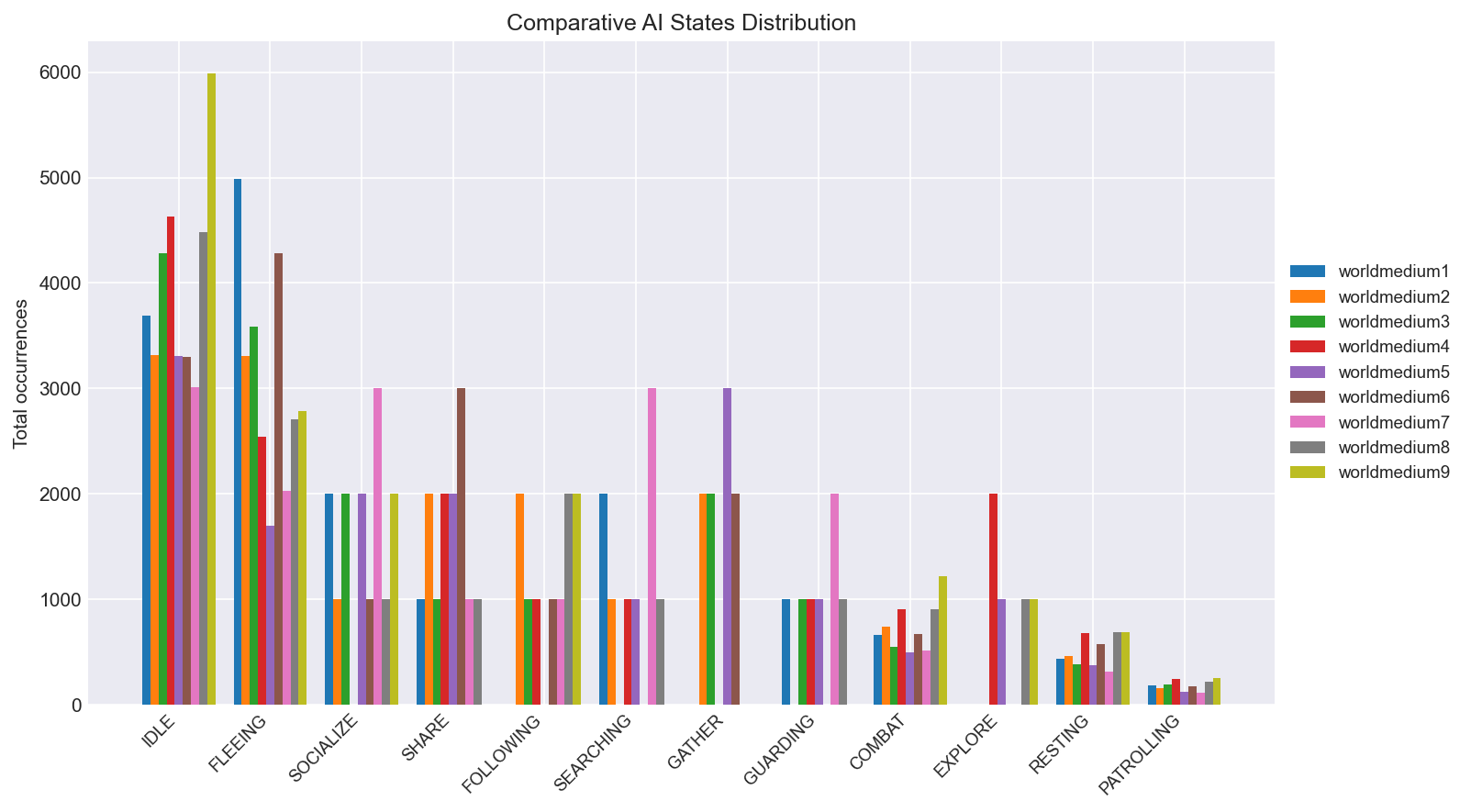}
    \end{subfigure}\hfill
    \begin{subfigure}[t]{0.95\columnwidth}
        \centering
        \includegraphics[width=\linewidth,height=5cm]{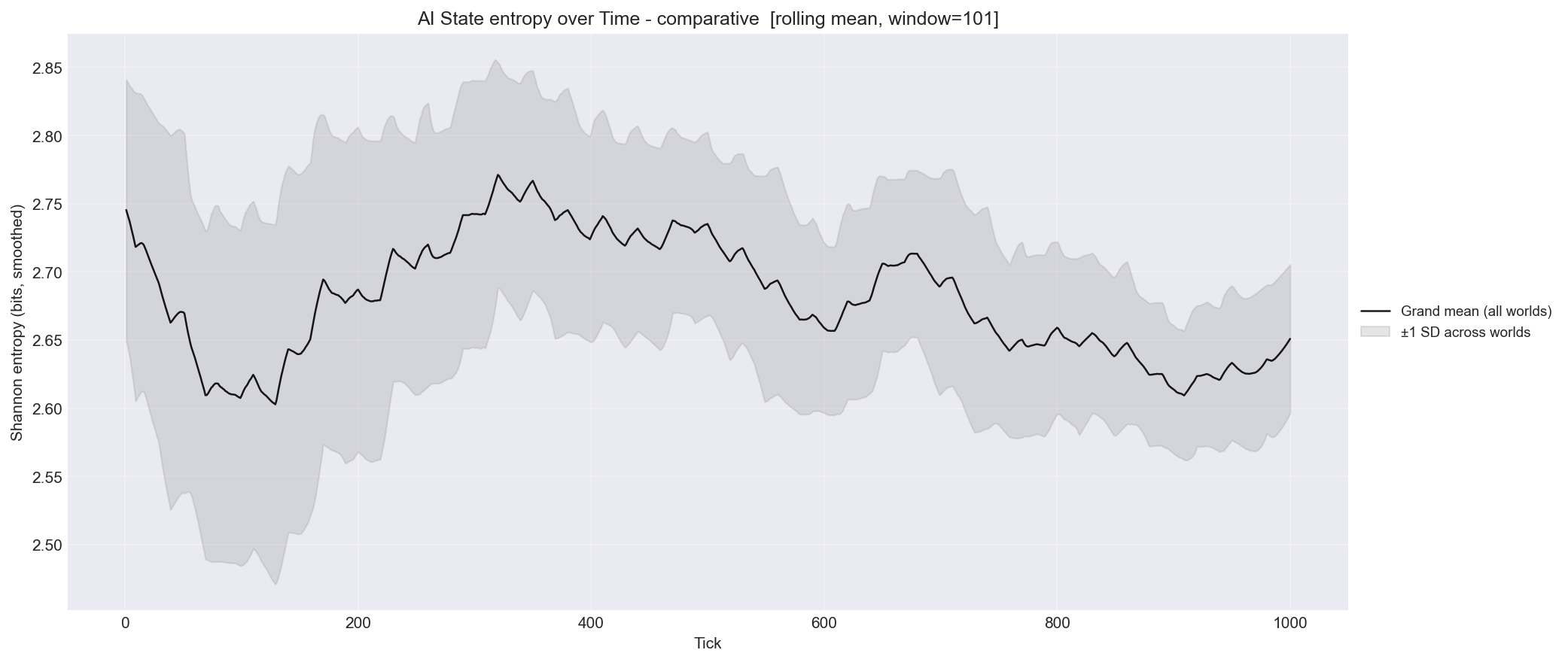}
    \end{subfigure}\\[1ex]
    \begin{subfigure}[t]{0.95\columnwidth}
        \centering
        \includegraphics[width=\linewidth,height=5cm]{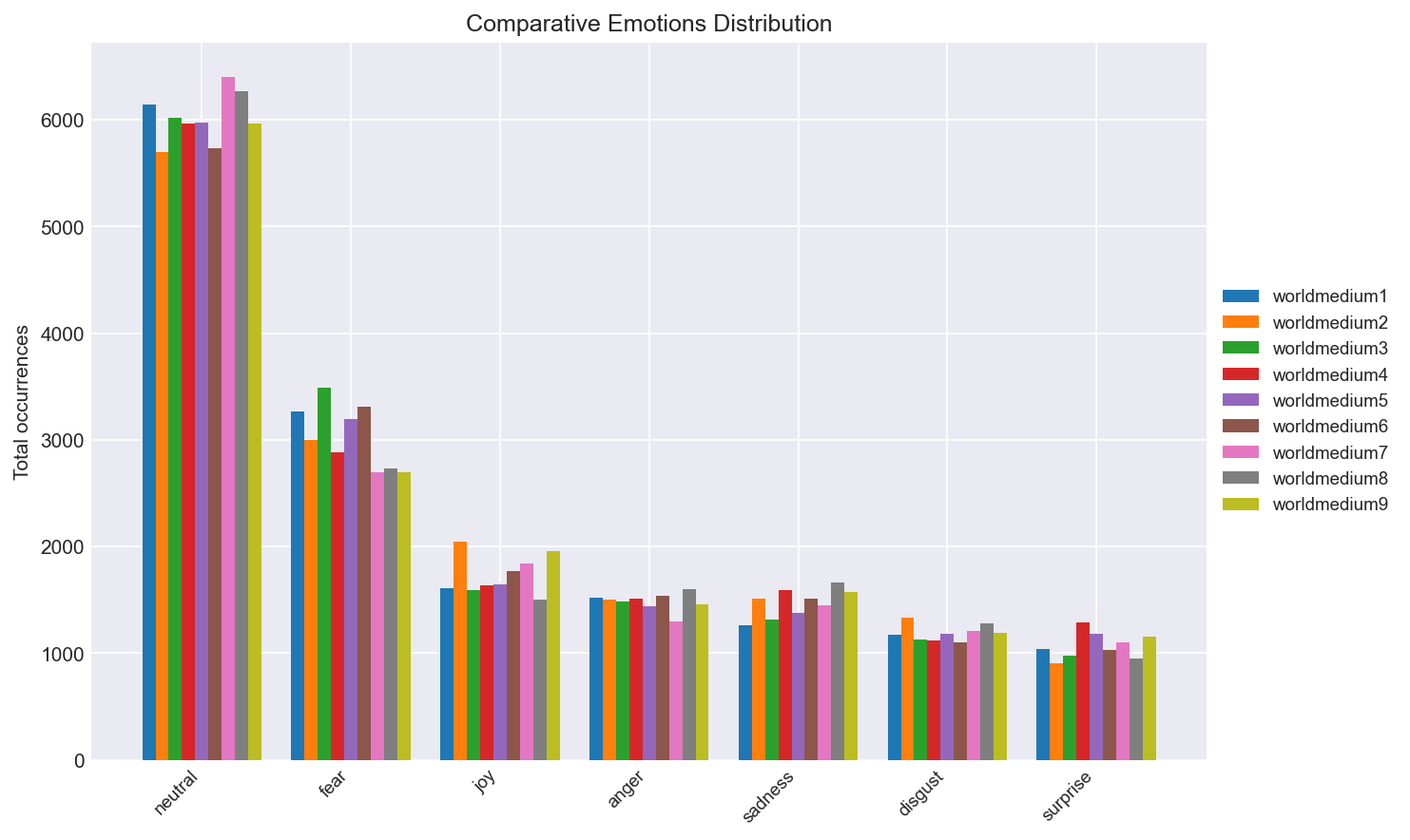}
    \end{subfigure}\hfill
    \begin{subfigure}[t]{0.95\columnwidth}
        \centering
        \includegraphics[width=\linewidth,height=5cm]{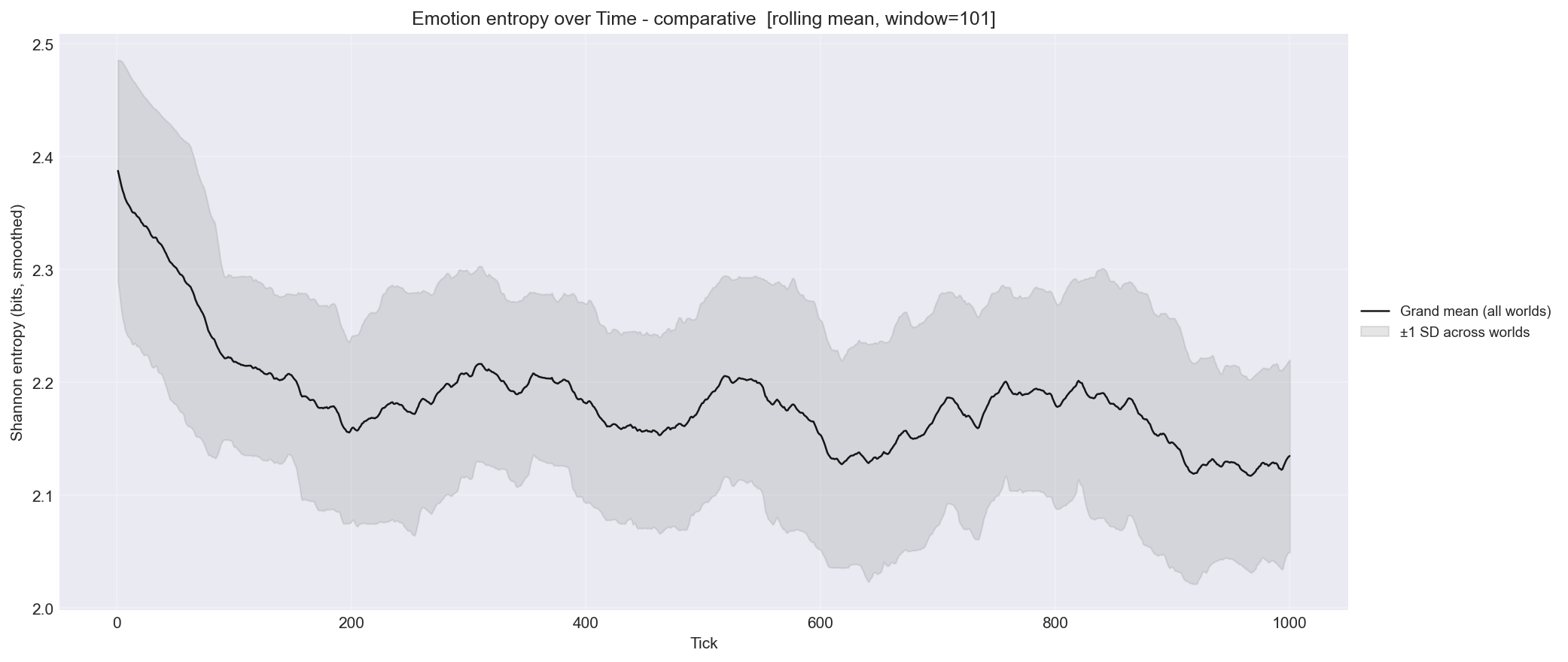}
    \end{subfigure}
    \caption{Most common category distributions (left) and entropy value timeline (right) across worlds in \textbf{\texttt{World Medium}}.}
    \label{fig:distributions_medium}
\end{figure*}

\begin{figure*}[htbp]
    \centering
    \begin{subfigure}[t]{0.95\columnwidth}
        \centering
        \includegraphics[width=\linewidth,height=5cm]{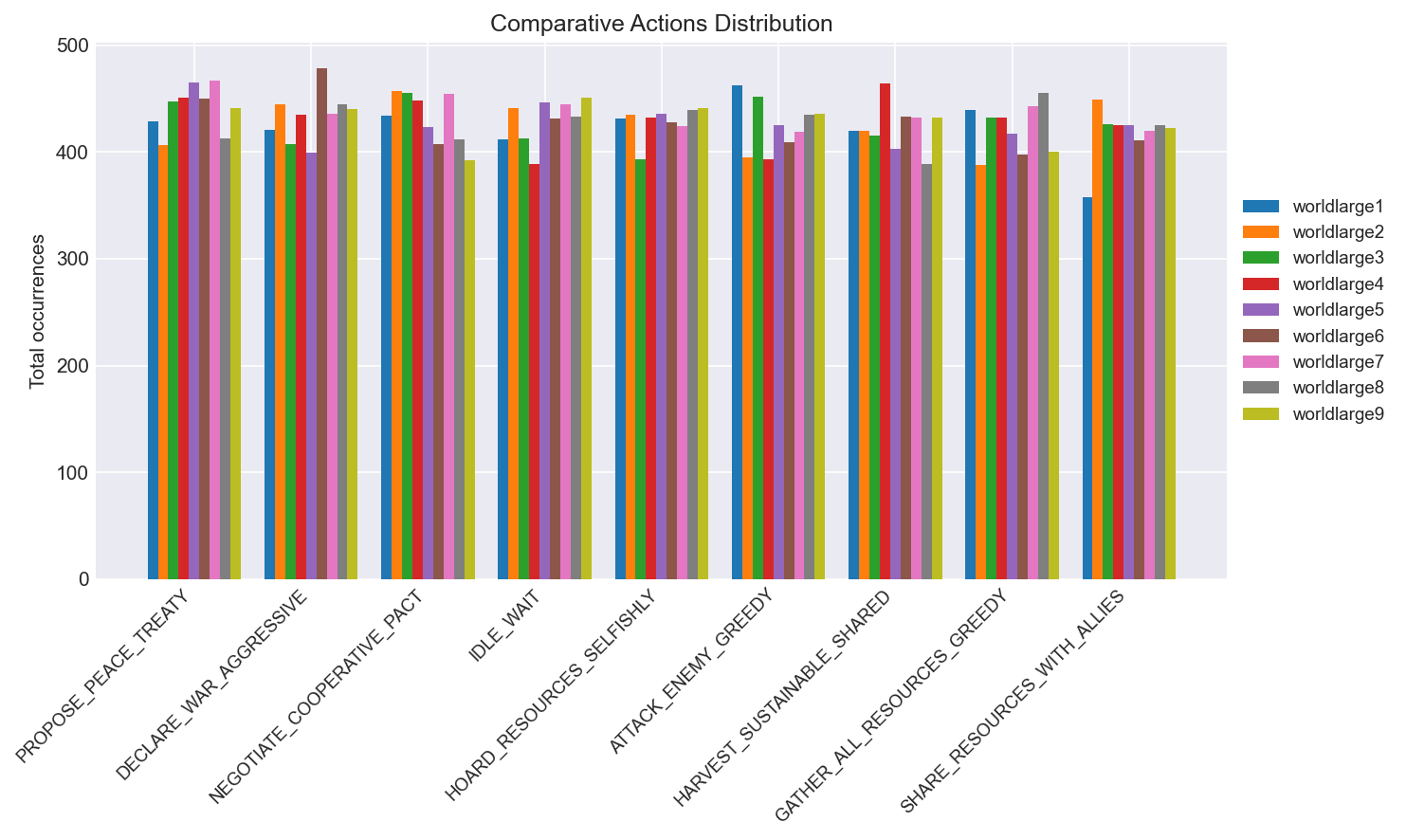}
    \end{subfigure}\hfill
    \begin{subfigure}[t]{0.95\columnwidth}
        \centering
        \includegraphics[width=\linewidth,height=5cm]{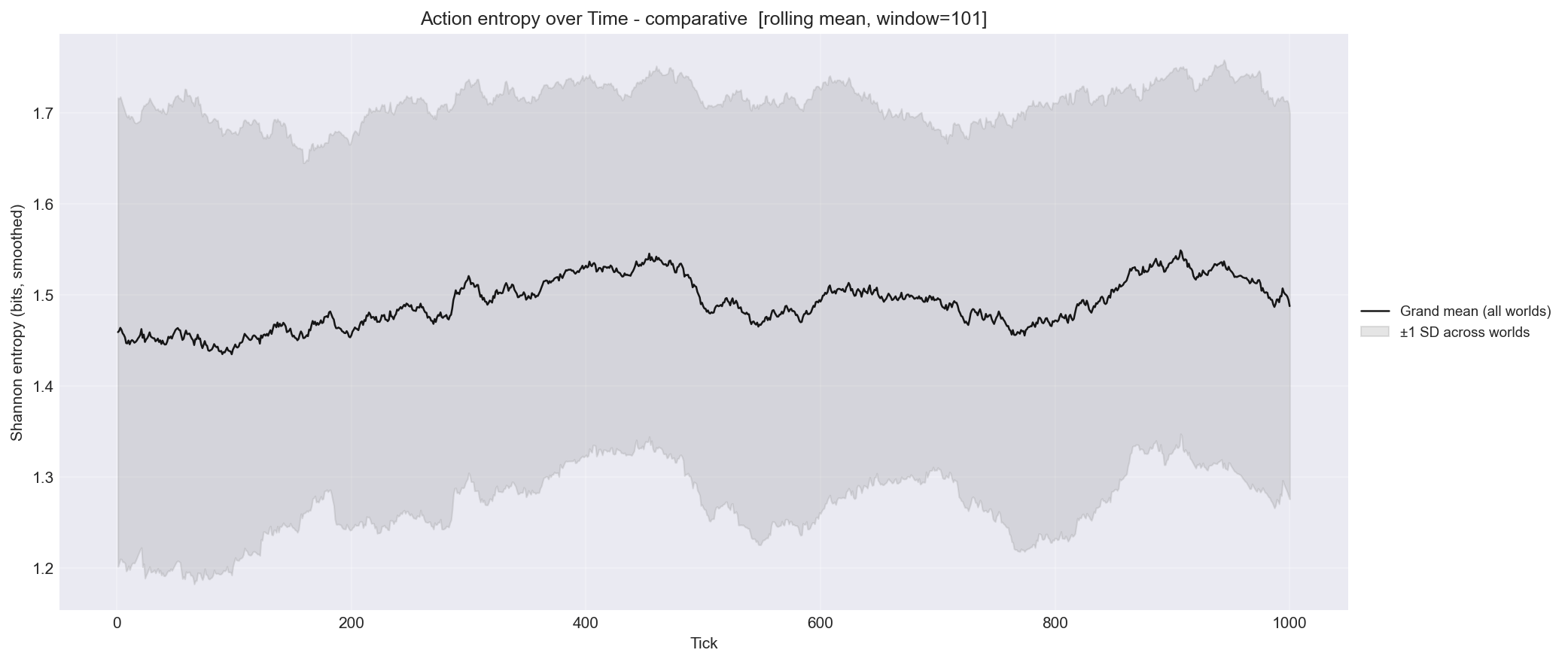}
    \end{subfigure}\\[1ex]
    \begin{subfigure}[t]{0.95\columnwidth}
        \centering
        \includegraphics[width=\linewidth,height=5cm]{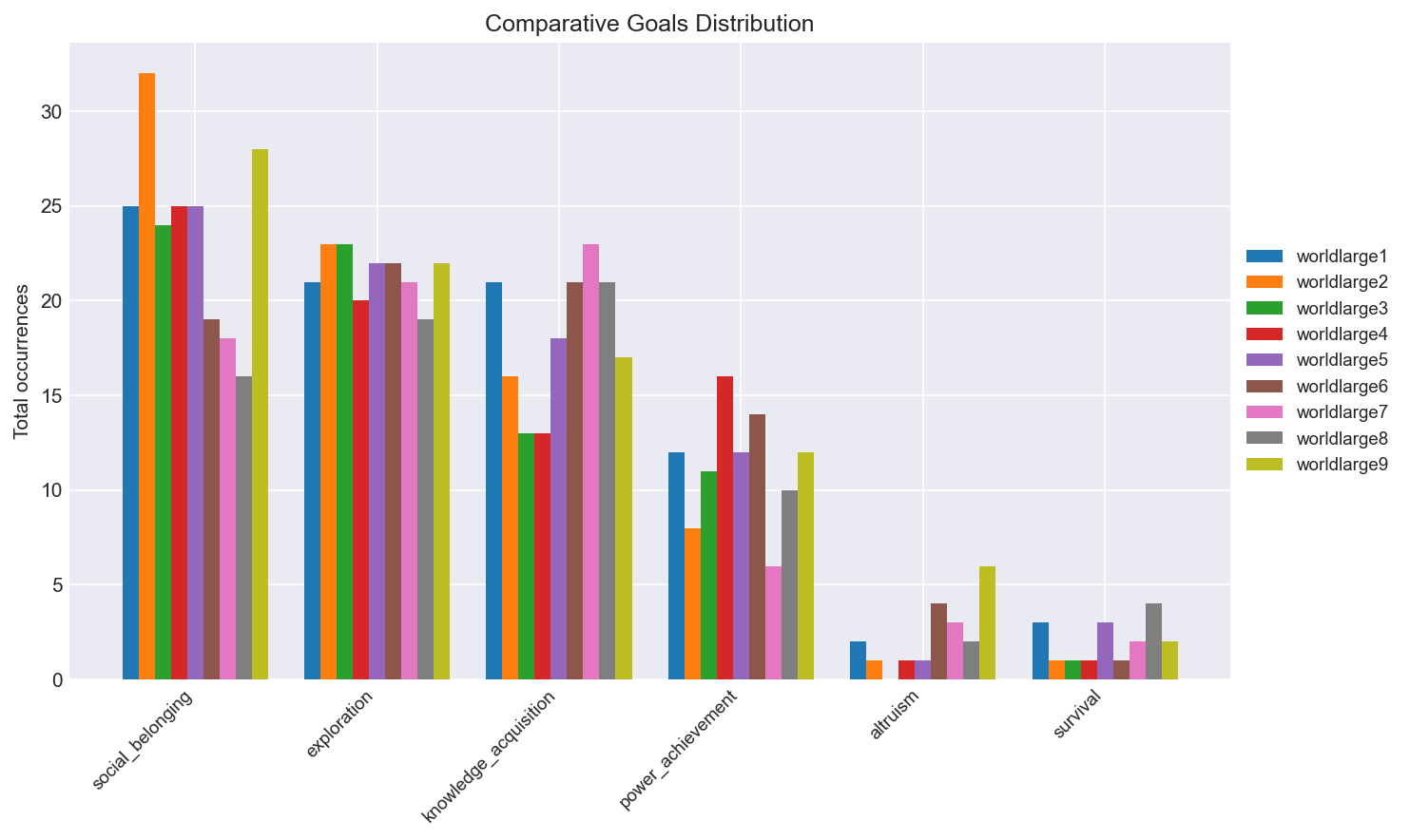}
    \end{subfigure}\hfill
    \begin{subfigure}[t]{0.95\columnwidth}
        \centering
        \includegraphics[width=\linewidth,height=5cm]{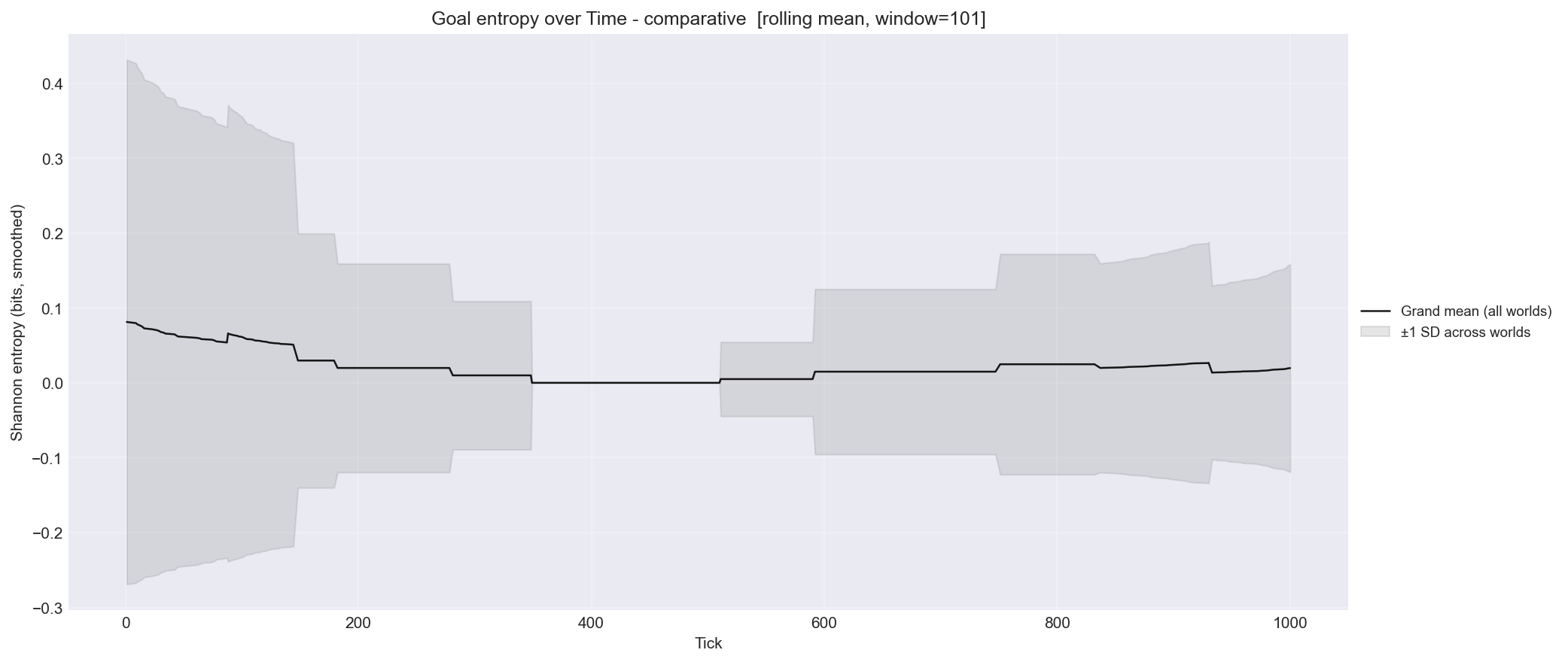}
    \end{subfigure}\\[1ex]
    \begin{subfigure}[t]{0.95\columnwidth}
        \centering
        \includegraphics[width=\linewidth,height=5cm]{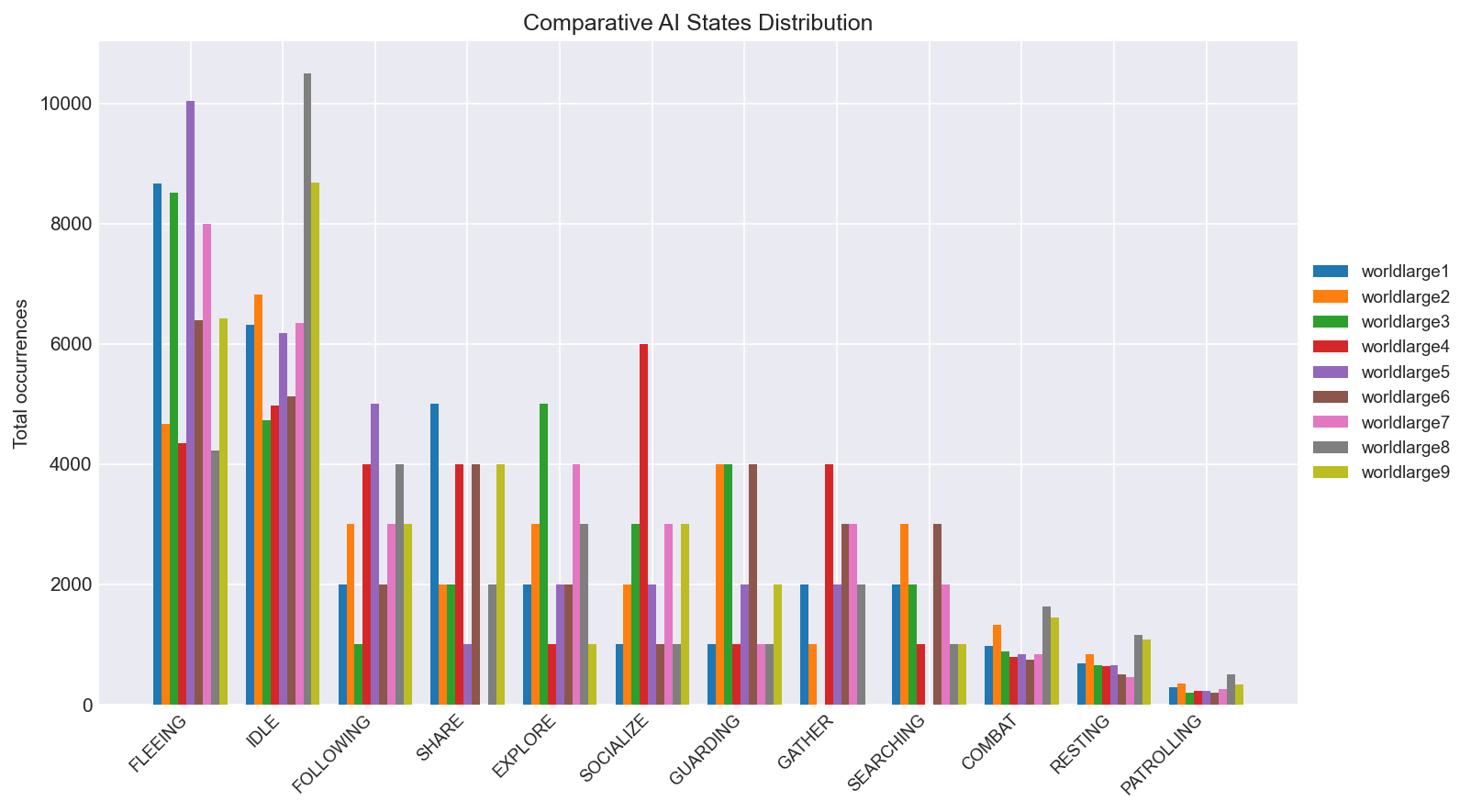}
    \end{subfigure}\hfill
    \begin{subfigure}[t]{0.95\columnwidth}
        \centering
        \includegraphics[width=\linewidth,height=5cm]{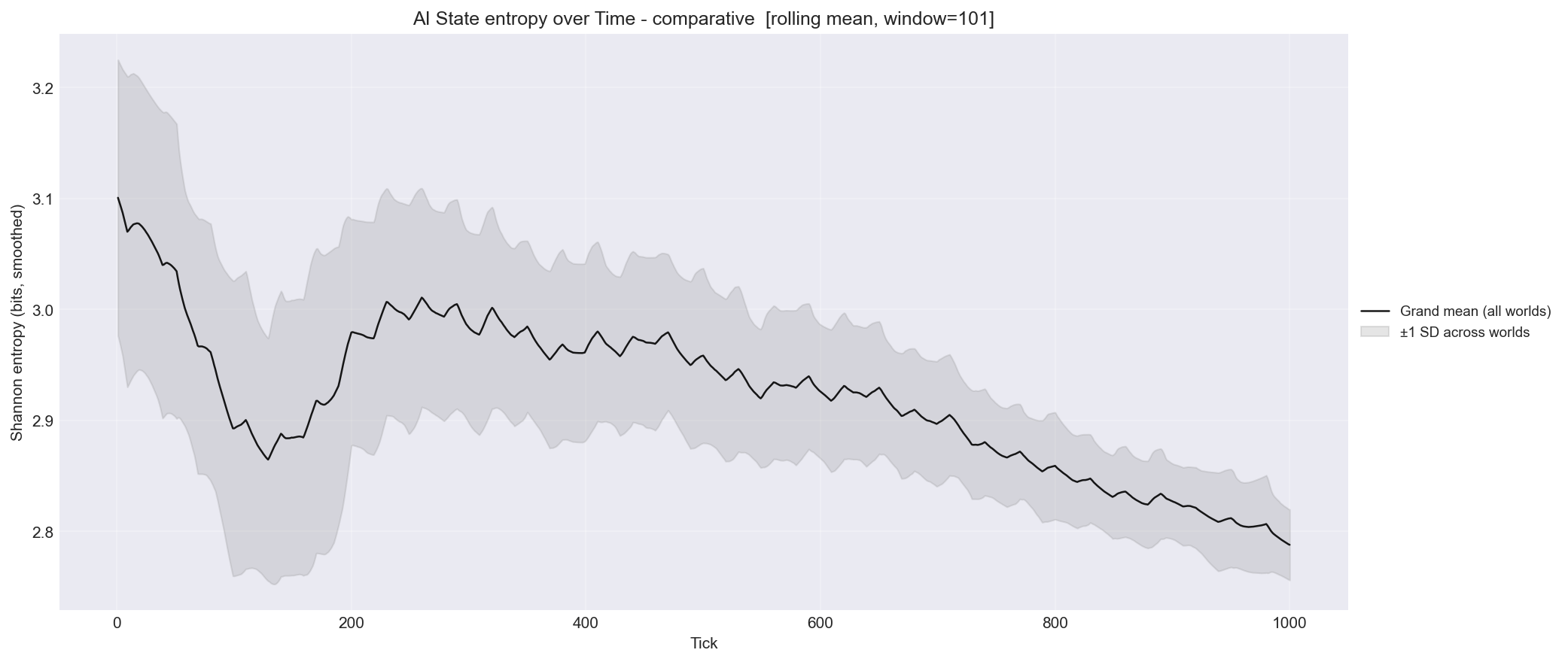}
    \end{subfigure}\\[1ex]
    \begin{subfigure}[t]{0.95\columnwidth}
        \centering
        \includegraphics[width=\linewidth,height=5cm]{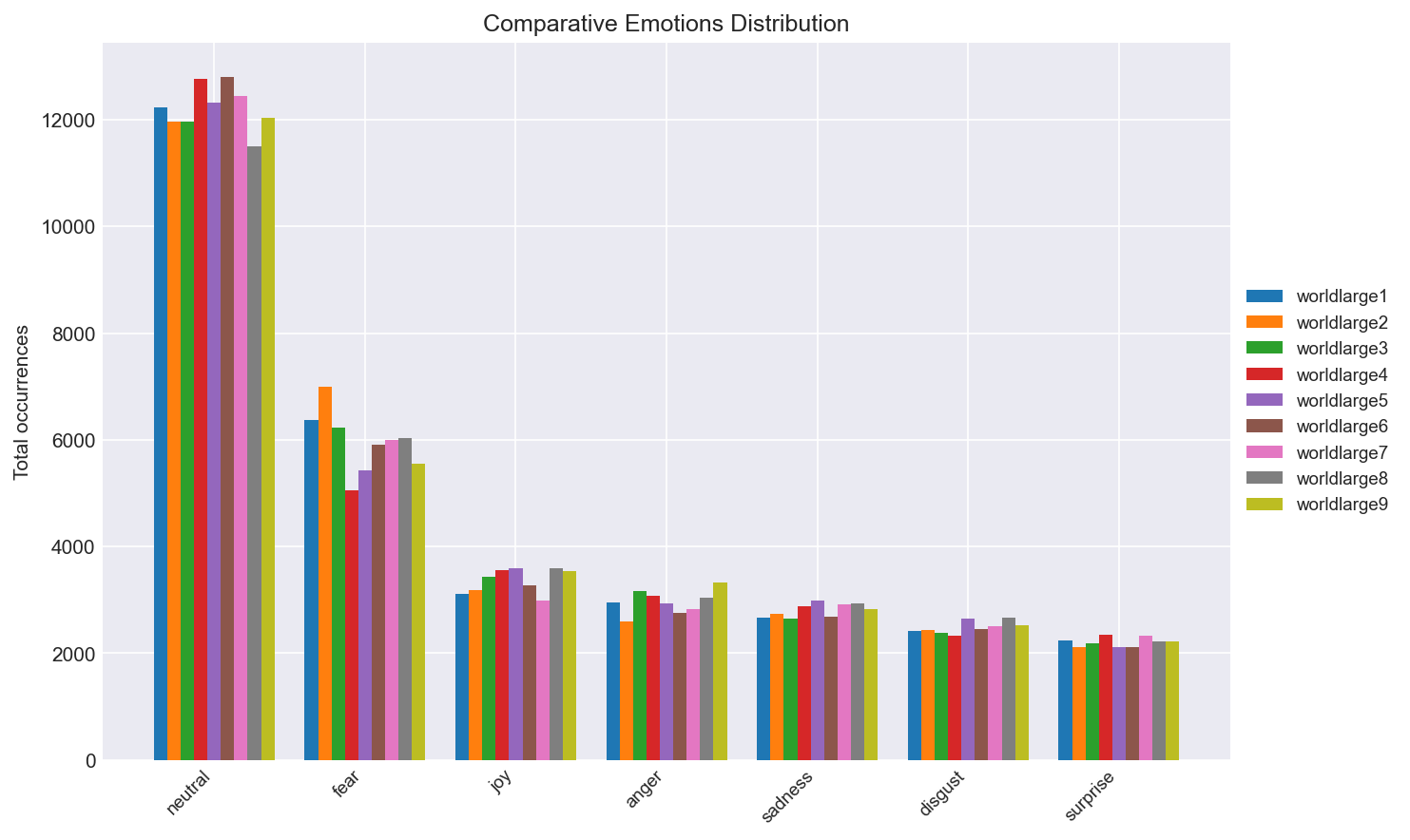}
    \end{subfigure}\hfill
    \begin{subfigure}[t]{0.95\columnwidth}
        \centering
        \includegraphics[width=\linewidth,height=5cm]{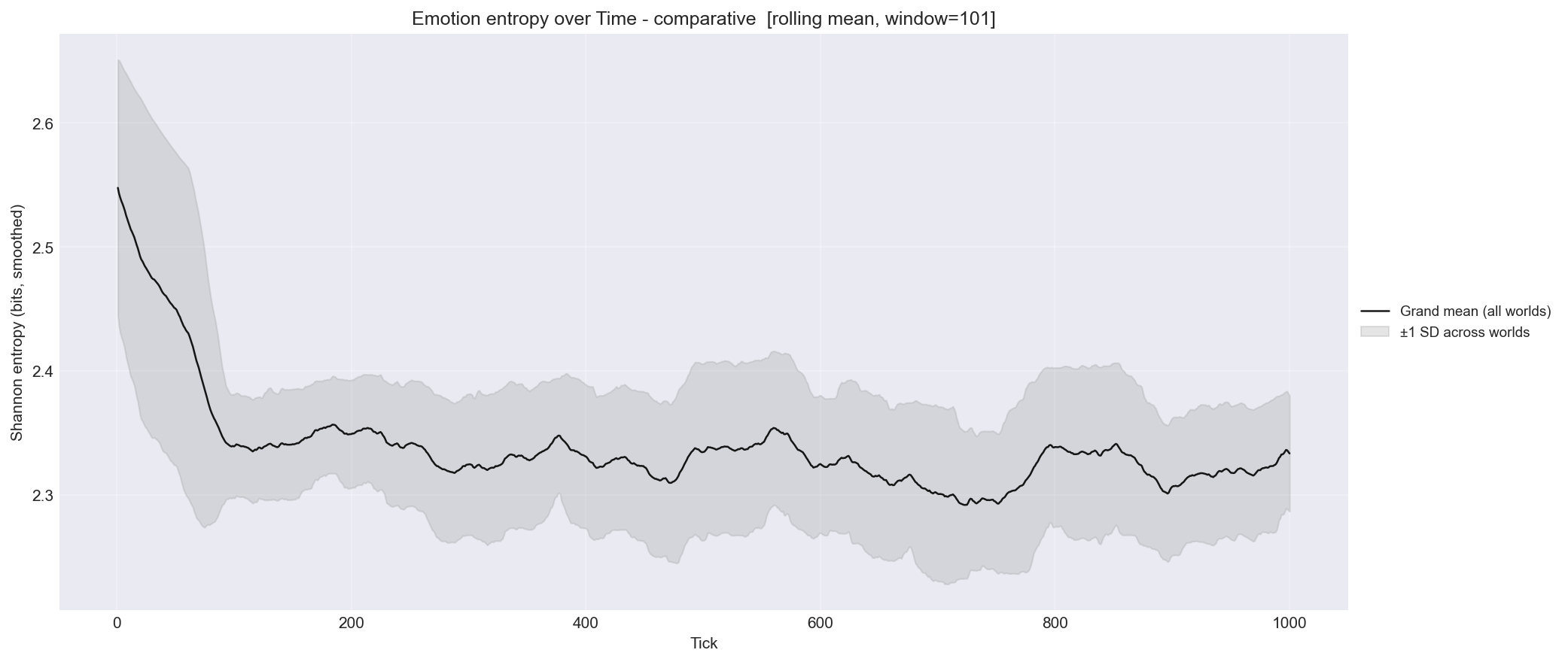}
    \end{subfigure}
    \caption{Most common category distributions (left) and entropy value timeline (right) across worlds in \textbf{\texttt{World Large}}.}
    \label{fig:distributions_large}
\end{figure*}

\begin{figure*}[htbp]
    \centering
    \begin{subfigure}[t]{0.95\columnwidth}
        \centering
        \includegraphics[width=\linewidth,height=5cm]{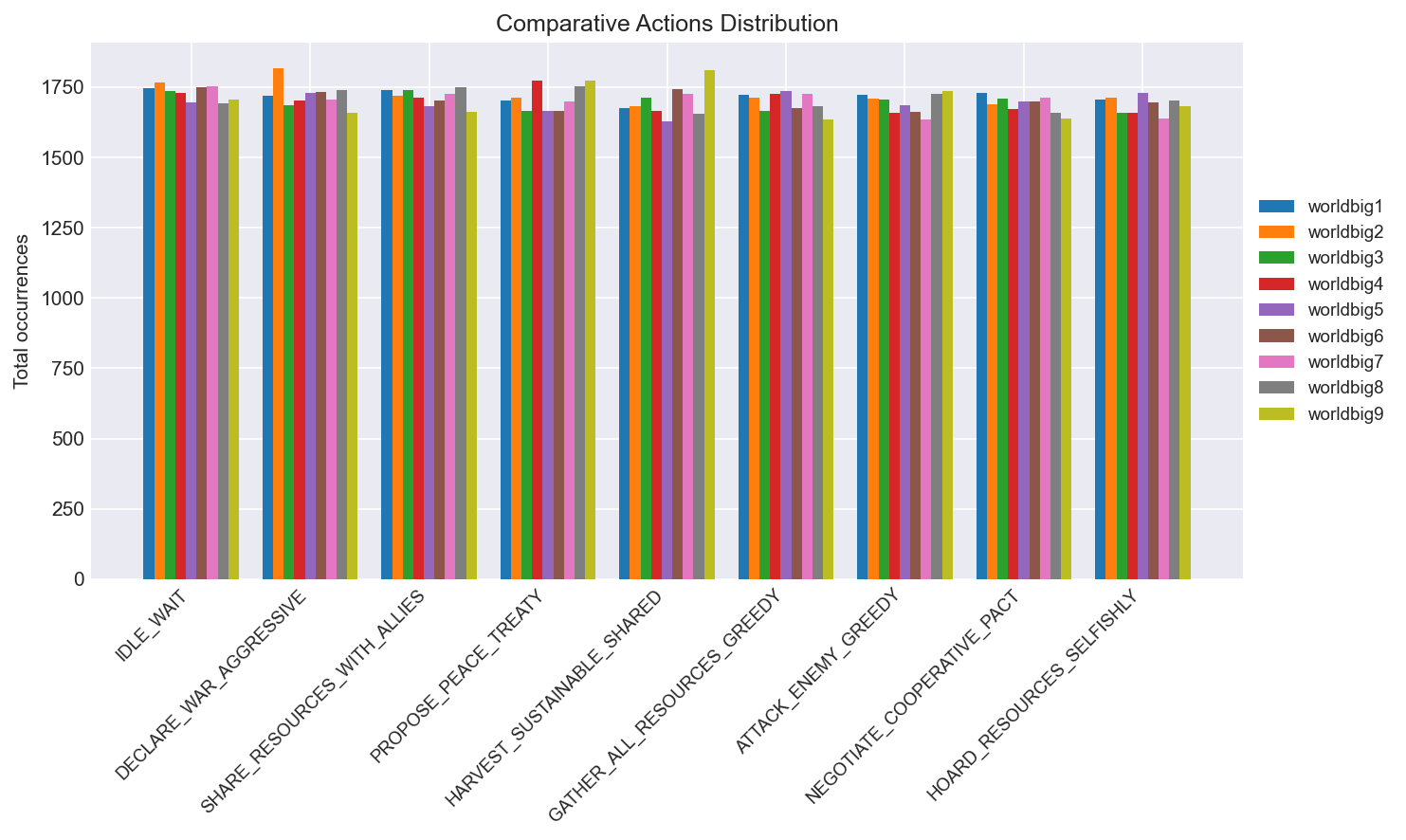}
    \end{subfigure}\hfill
    \begin{subfigure}[t]{0.95\columnwidth}
        \centering
        \includegraphics[width=\linewidth,height=5cm]{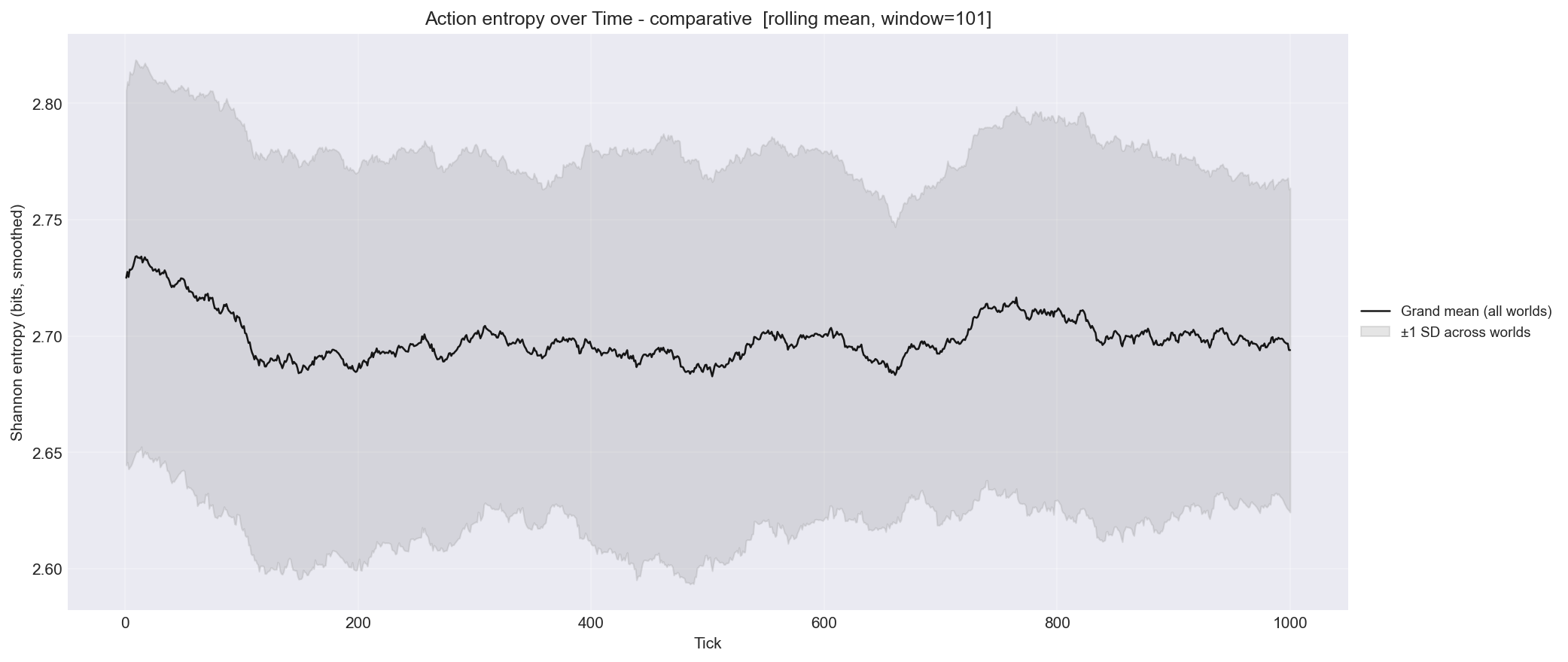}
    \end{subfigure}\\[1ex]
    \begin{subfigure}[t]{0.95\columnwidth}
        \centering
        \includegraphics[width=\linewidth,height=5cm]{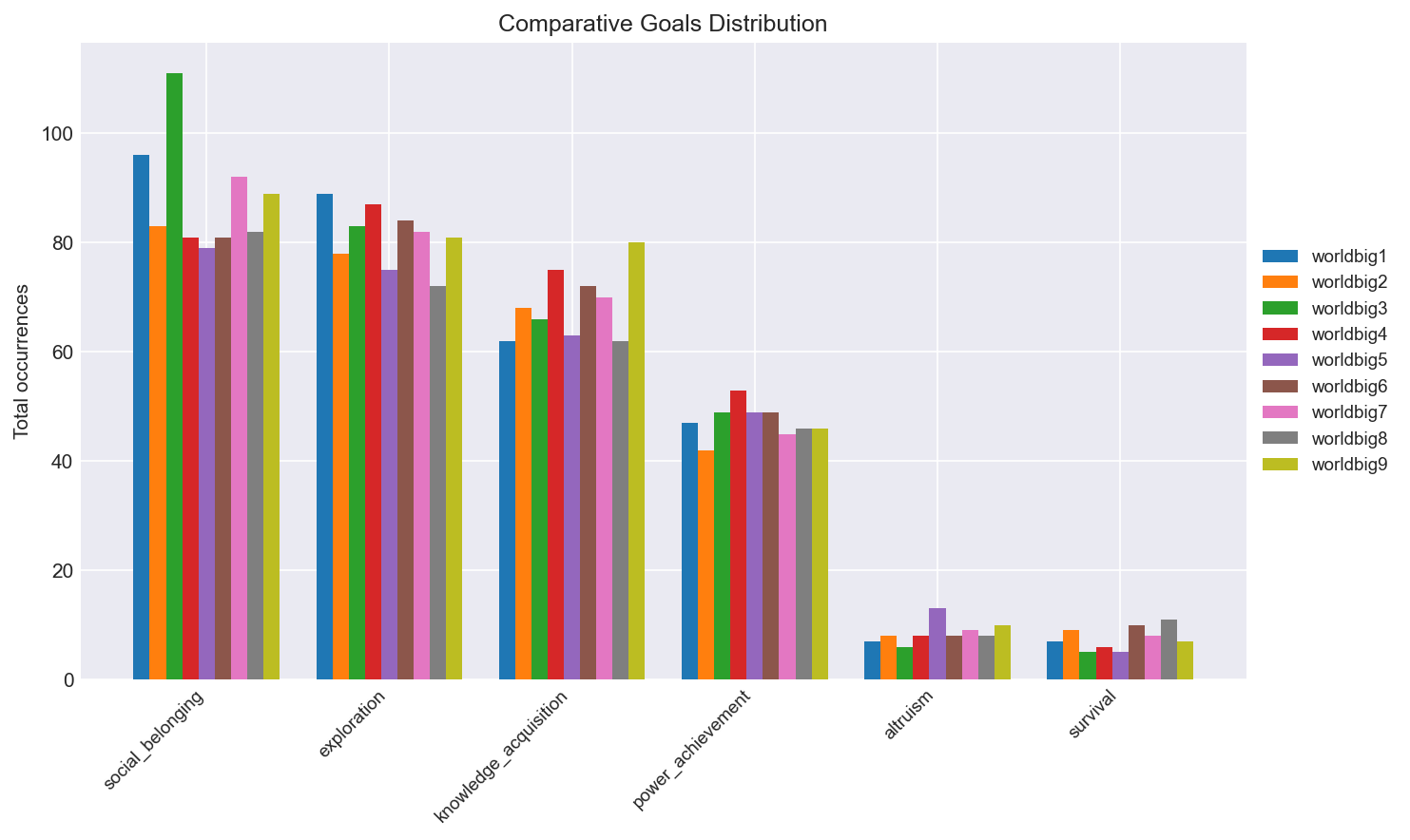}
    \end{subfigure}\hfill
    \begin{subfigure}[t]{0.95\columnwidth}
        \centering
        \includegraphics[width=\linewidth,height=5cm]{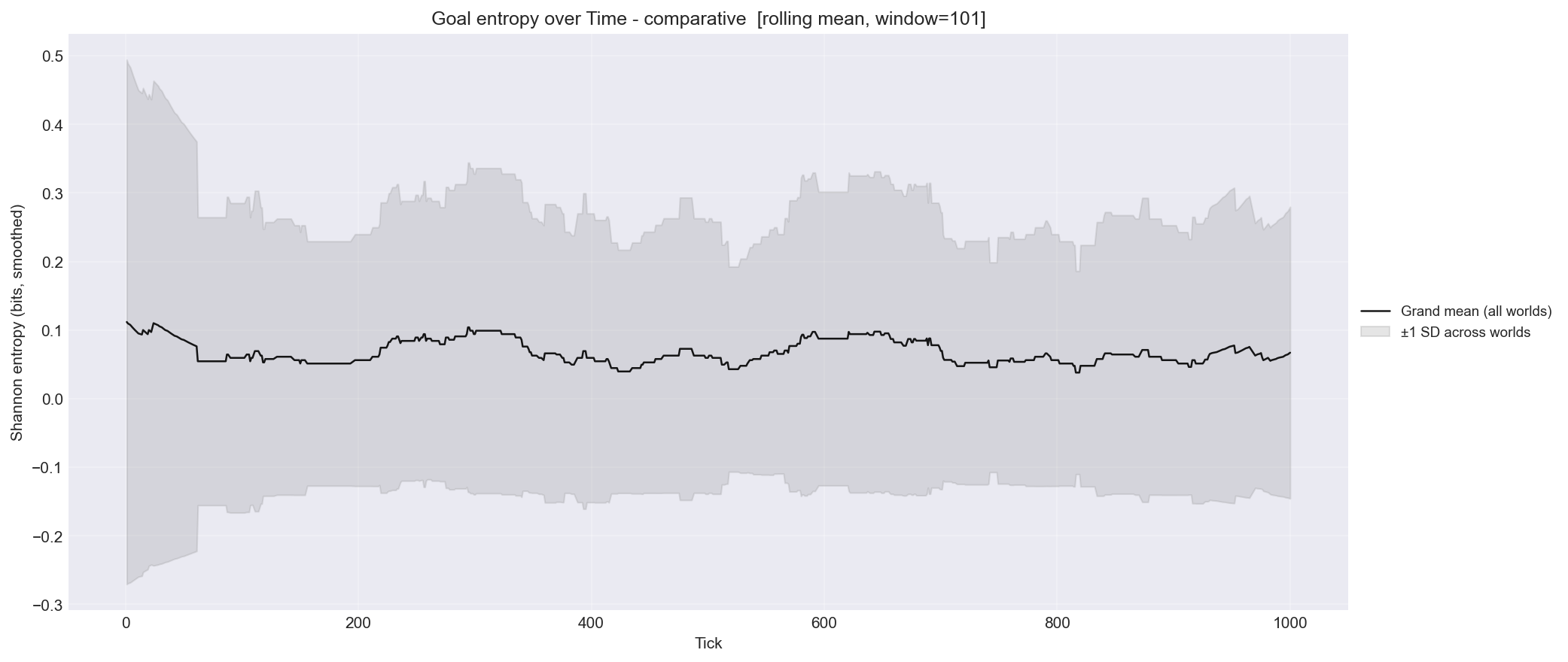}
    \end{subfigure}\\[1ex]
    \begin{subfigure}[t]{0.95\columnwidth}
        \centering
        \includegraphics[width=\linewidth,height=5cm]{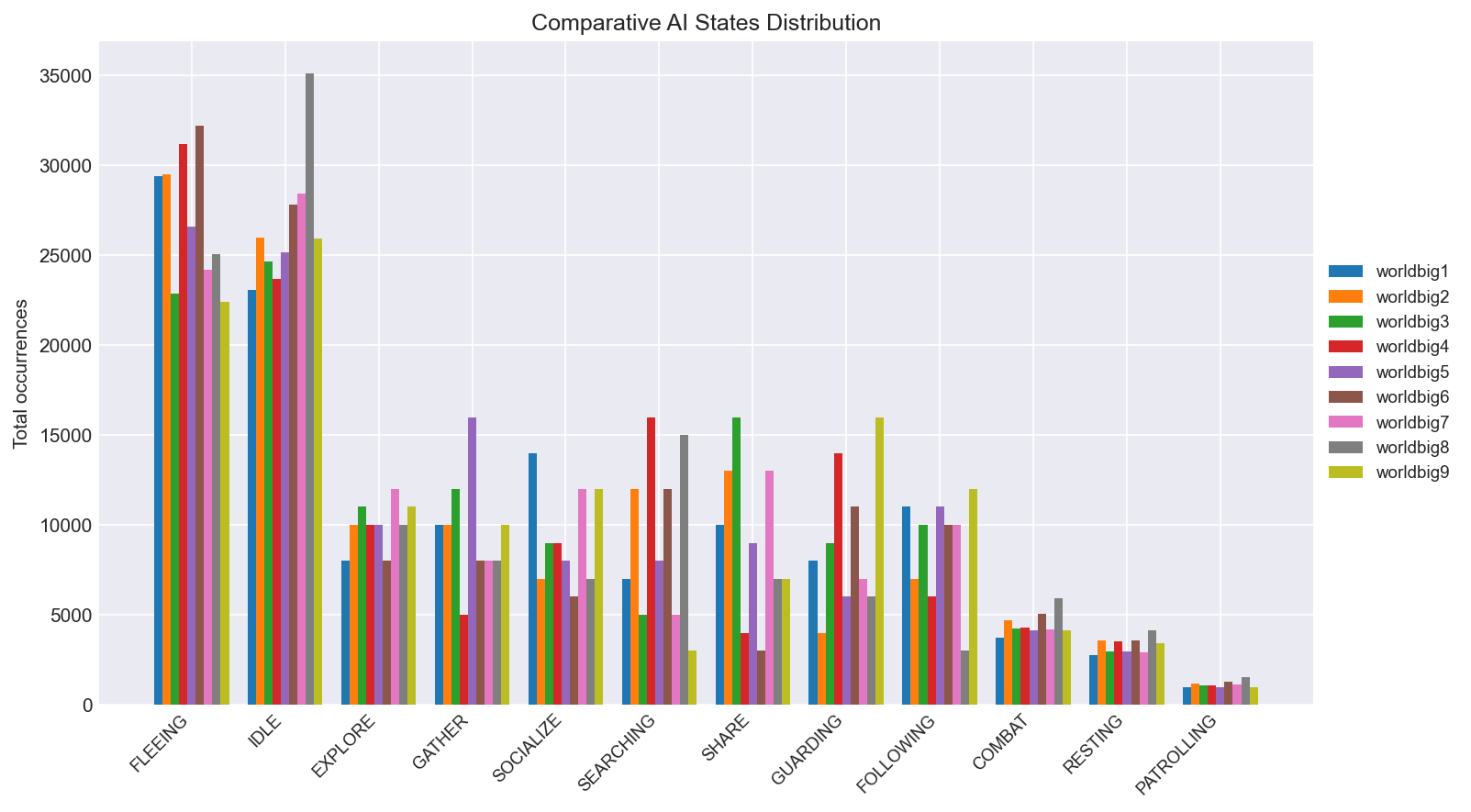}
    \end{subfigure}\hfill
    \begin{subfigure}[t]{0.95\columnwidth}
        \centering
        \includegraphics[width=\linewidth,height=5cm]{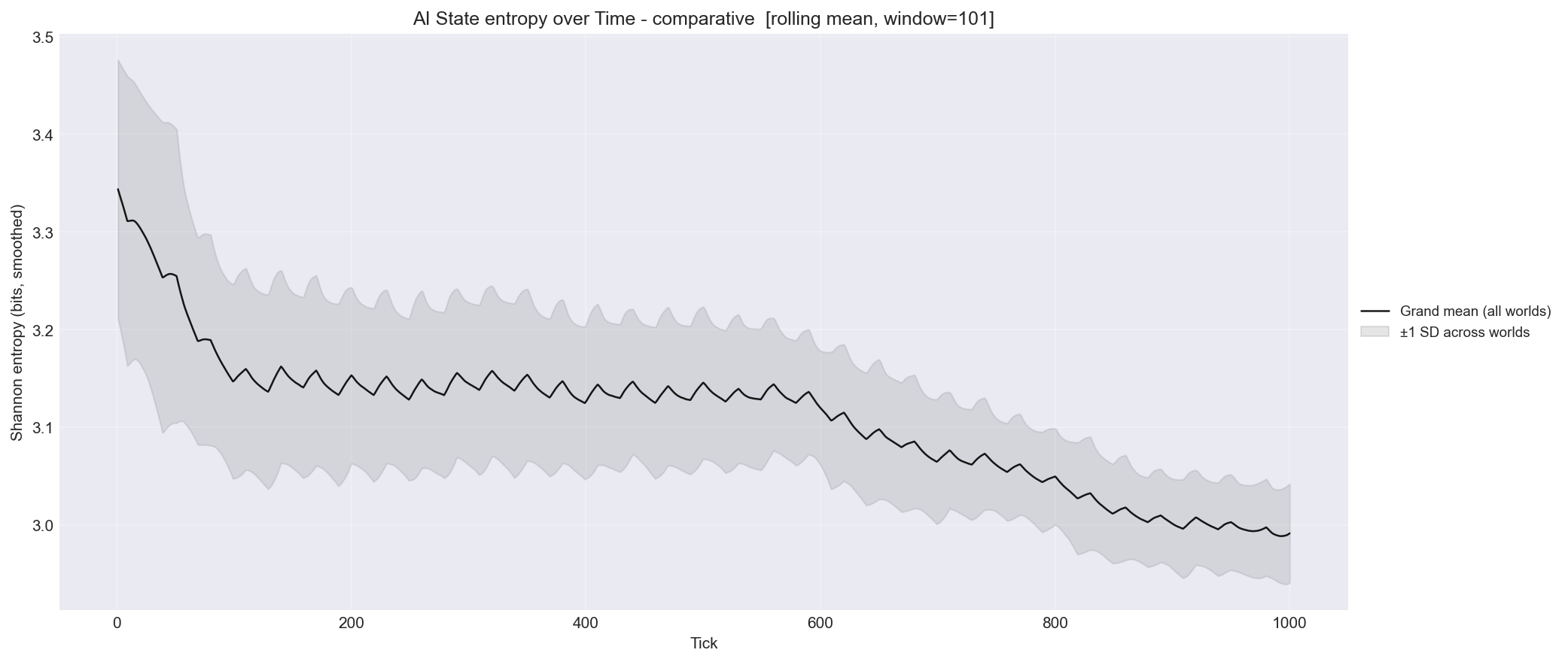}
    \end{subfigure}\\[1ex]
    \begin{subfigure}[t]{0.95\columnwidth}
        \centering
        \includegraphics[width=\linewidth,height=5cm]{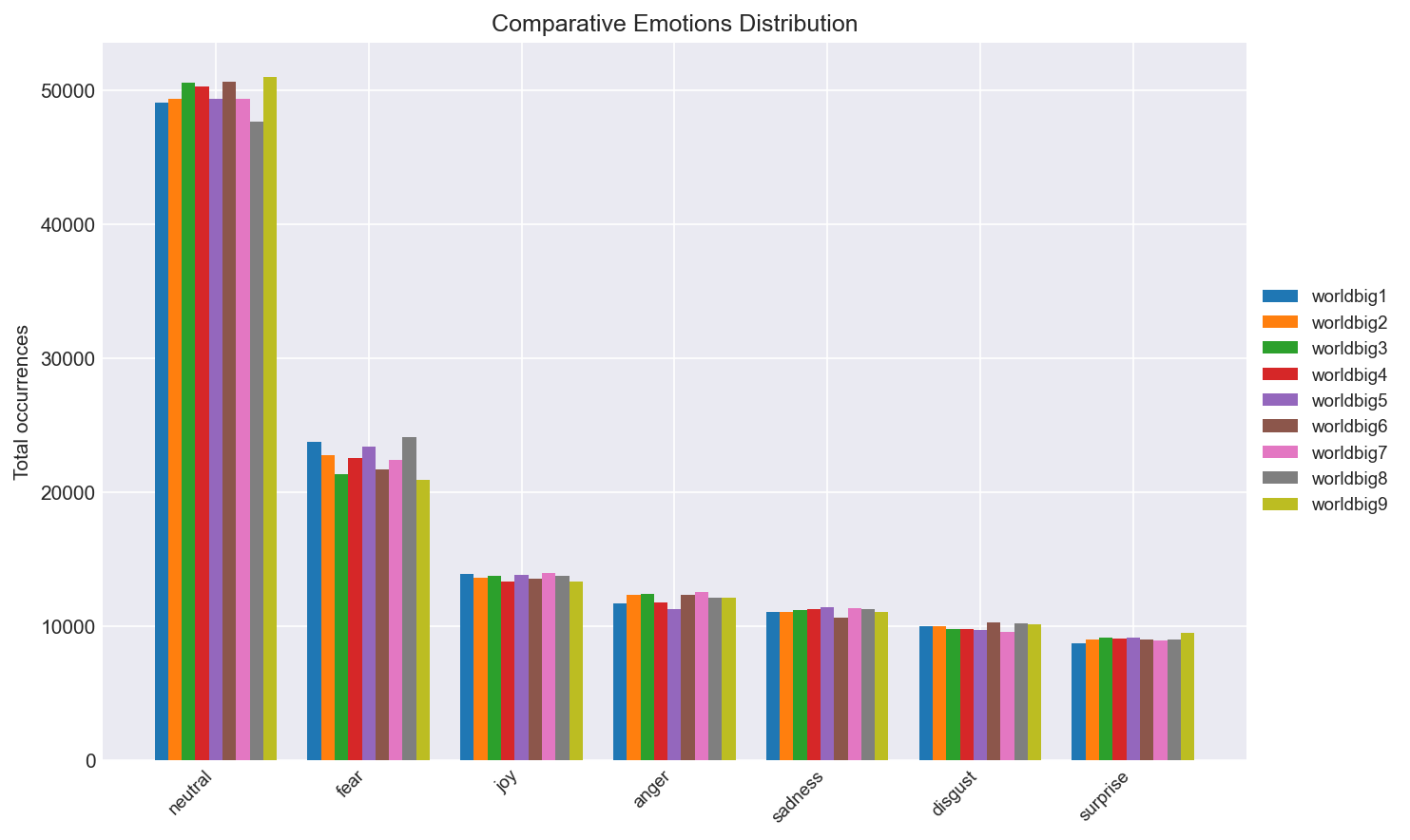}
    \end{subfigure}\hfill
    \begin{subfigure}[t]{0.95\columnwidth}
        \centering
        \includegraphics[width=\linewidth,height=5cm]{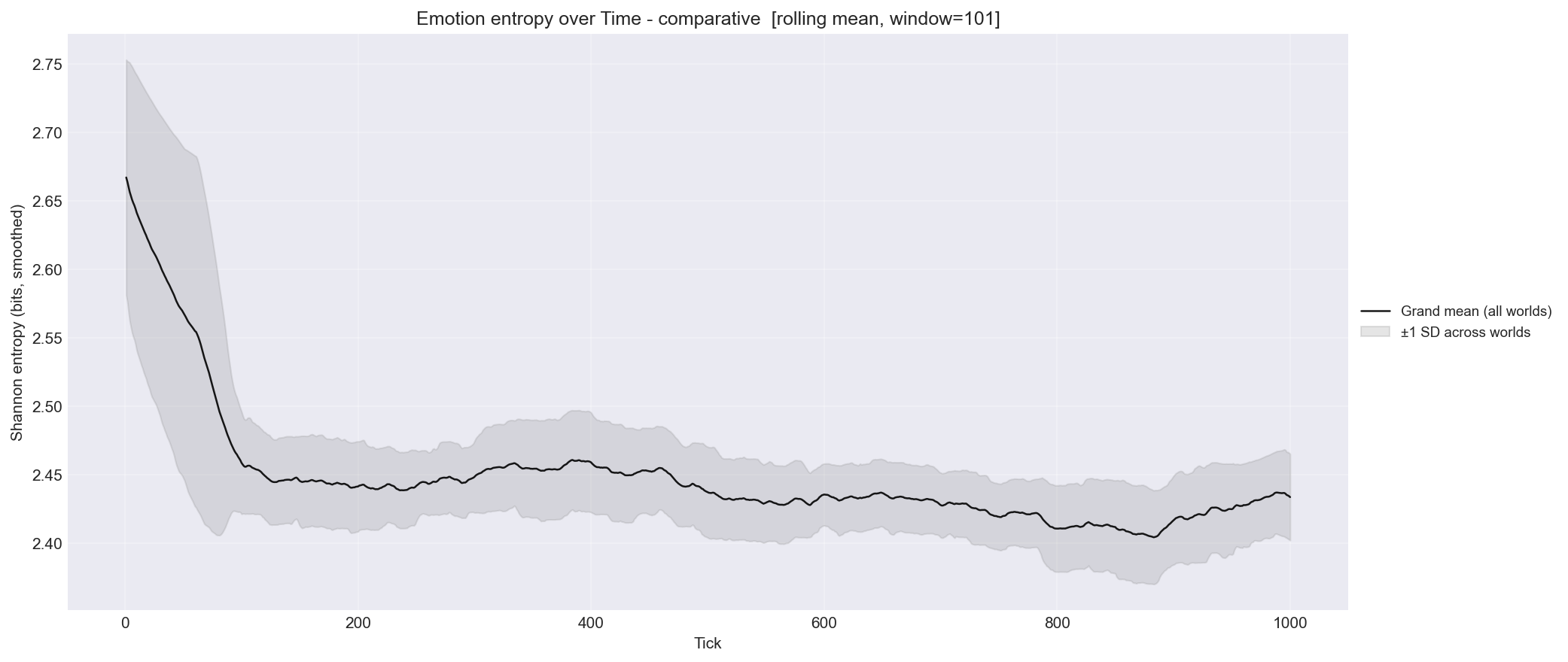}
    \end{subfigure}
    \caption{Most common category distributions (left) and entropy value timeline (right) across worlds in \textbf{\texttt{World Big}}.}
    \label{fig:distributions_big}
\end{figure*}

In this test, we generated a total of 45 world simulations (9 instances $\times$ 5 world scales), ranging from \textbf{\texttt{World Tiny}} (4 characters), \textbf{\texttt{World Small}} (8 characters), \textbf{\texttt{World Medium}} (16 characters), \textbf{\texttt{World Large}} (64 characters) and \textbf{\texttt{World Big}} (128 characters). We ran each world for 1.000 inference steps an analyzed through statistical tests the interactions between status variables and socio-affective factor co-occurrences (actions, goals, AI states and emotion categories). See factor frequency distributions in Figures \ref{fig:distributions_tiny}-\ref{fig:distributions_big} (left) with each factor value as the stacked per-participant's mean entropy timeline (per step) in Figures \ref{fig:distributions_tiny}-\ref{fig:distributions_big} (right). 

We also computed descriptive statistics and correlation tests (Pearson's $r$ and Spearman $\rho$ respectively) for character status variables with each socio-affective factor interaction in Tables \ref{tab:cross_world_dynamics_tiny}-\ref{tab:correlation_matrix_big}. Across the five world scales, the dominant and scale-invariant result is a negative coupling between health and AI-state diversity. Note $r_{T}$ for World Tiny ($n=36$), $r_{S}$ for World Small ($n=72$), $r_{M}$ for World Medium ($n=144$), $r_{L}$ for World Large ($n=288$) and $r_{B}$ for World Big ($n=1152$). For variable correlations we can analyze the system design:
\begin{itemize}
\item In terms of character status variables, \texttt{Mean\_Health} correlates with \texttt{Distinct\_AI\_States} at $r_{T}=-0.73$, $r_{S}=-0.68$, $r_{M}=-0.65$, $r_{L}=-0.67$, $r_{B}=-0.68$, and with \texttt{AI\_State\_Entropy} at $r_{T}=-0.67$, $r_{S}=-0.67$, $r_{M}=-0.64$, $r_{L}=-0.66$, $r_{B}=-0.67$, while it is positively coupled with \texttt{Emotion\_Transitions} at $r_{T}=+0.68$, $r_{S}=+0.69$, $r_{M}=+0.80$, $r_{L}=+0.77$, $r_{B}=+0.73$. \texttt{Mean\_Stamina} follows the same sign pattern with \texttt{Distinct\_AI\_States} at $r_{T}=-0.21$, $r_{S}=-0.54$, $r_{M}=-0.54$, $r_{L}=-0.50$, $r_{B}=-0.55$ and with \texttt{AI\_State\_Entropy} at $r_{T}=-0.54$, $r_{S}=-0.61$, $r_{M}=-0.65$, $r_{L}=-0.62$, $r_{B}=-0.65$. 
\item In terms of physiological needs: \texttt{Mean\_Hunger} correlates with \texttt{Distinct\_AI\_States} at $r_{T}=+0.43$, $r_{S}=+0.51$, $r_{M}=+0.47$, $r_{L}=+0.50$, $r_{B}=+0.50$ and with \texttt{AI\_State\_Entropy} at $r_{T}=+0.55$, $r_{S}=+0.56$, $r_{M}=+0.50$, $r_{L}=+0.54$, $r_{B}=+0.56$, while \texttt{Mean\_Thirst} gives $r_{T}=+0.42$, $r_{S}=+0.44$, $r_{M}=+0.51$, $r_{L}=+0.47$, $r_{B}=+0.45$ for \texttt{AI\_State\_Entropy}, with the two needs themselves coupled at $r_{T}=+0.54$, $r_{S}=+0.45$, $r_{M}=+0.34$, $r_{L}=+0.31$, $r_{B}=+0.30$. 
\item By contrast, \texttt{Mean\_Energy}, \texttt{Mean\_Loyalty} and \texttt{Mean\_Trust} report exactly $r=0.00$ against every activity column in all five worlds, and the remaining coefficients below $|r|=0.30$ are too small to support any interpretation in these runs. This shows these variables compute overall less effects in the system.
\end{itemize}
All magnitudes survive two orders of magnitude of scaling from $4$ to $128$ characters, which is consistent with \varalias{} design: low health and stamina activate survival-oriented goals and Frijda action tendencies (e.g.fear~$\rightarrow$~FLEEING, anger~$\rightarrow$~COMBAT) as well as Hull's drive-reduction and Maslow's physiological tier, so degraded agents concentrate on fewer, more persistent AI states, whereas unmet hunger and thirst widen the set of active drives and therefore the state distribution per character. 

\vspace{-0.25cm}

\begin{table*}[htbp]
\centering
\caption{Cross-world dynamics of character status variables (mean value) in \textbf{\texttt{World Tiny}}.}
\label{tab:cross_world_dynamics_tiny}
\begin{adjustbox}{width=0.6\textwidth}
\begin{tabular}{|l|c|c|c|c|c|}
\hline
\textbf{World} & \textbf{health} & \textbf{hunger} & \textbf{morale} & \textbf{stamina} & \textbf{thirst}  \\
\hline
worldtiny1  & 82.0 & 2.9 & 84.1 & 92.4 & 5.2 \\
worldtiny2  & 63.4 & 2.9 & 87.0 & 92.4 & 4.0 \\
worldtiny3  & 70.2 & 5.2 & 83.0 & 93.1 & 6.8\\
worldtiny4  & 67.5 & 6.5 & 72.2 & 91.9 & 5.1\\
worldtiny5  & 63.8 & 3.2 & 85.3 & 95.2 & 4.8\\
worldtiny6  & 76.5 & 4.2 & 88.3 & 92.4 & 5.7\\
worldtiny7  & 77.6 & 3.6 & 85.5 & 89.6 & 6.0\\
worldtiny8  & 83.0 & 3.9 & 87.9 & 94.6 & 4.1\\
worldtiny9  & 71.5 & 4.8 & 86.8 & 89.3 & 6.2\\
\hline
\end{tabular}
\end{adjustbox}
\end{table*}

\begin{table*}[htbp]
\centering
\caption{Status $\times$ activity correlations (n=36). Each cell shows Pearson $r$ / Spearman $\rho$ in \textbf{\texttt{World Tiny}}.}
\label{tab:status_activity_corr_tiny}
\begin{adjustbox}{width=1\linewidth}
\begin{tabular}{|l|c|c|c|c|c|c|c|c|c|c|c|}
\hline
\textbf{Status} & \textbf{Actions\_Total} & \textbf{Emotions\_Total} & \textbf{Goals\_Total} & \textbf{AI\_States\_Total} & \textbf{Distinct\_Emotions} & \textbf{Distinct\_Goals} & \textbf{Distinct\_AI\_States} & \textbf{Emotion\_Transitions} & \textbf{Emotion\_Entropy} & \textbf{Goal\_Entropy} & \textbf{AI\_State\_Entropy} \\
\hline
\texttt{Mean\_Energy} & +0.00/+0.00 & +0.00/+0.00 & +0.00/+0.00 & +0.00/+0.00 & +0.00/+0.00 & +0.00/+0.00 & +0.00/+0.00 & +0.00/+0.00 & +0.00/+0.00 & +0.00/+0.00 & +0.00/+0.00 \\
\texttt{Mean\_Health} & +0.04/+0.07 & +0.00/+0.00 & \cellcolor{yellow!20}+0.16/+0.20 & +0.00/+0.00 & +0.00/+0.00 & \cellcolor{yellow!20}+0.10/+0.14 & \cellcolor{red!20}-0.73/-0.74 & \cellcolor{red!20}+0.68/+0.61 & \cellcolor{yellow!20}-0.21/-0.43 & \cellcolor{yellow!20}+0.11/+0.08 & \cellcolor{red!20}-0.67/-0.73 \\
\texttt{Mean\_Hunger} & +0.03/+0.06 & +0.00/+0.00 & -0.09/-0.14 & +0.00/+0.00 & +0.00/+0.00 & \cellcolor{yellow!20}-0.14/-0.15 & \cellcolor{orange!20}+0.43/+0.37 & -0.08/-0.12 & \cellcolor{red!20}+0.54/+0.68 & \cellcolor{yellow!20}-0.17/-0.15 & \cellcolor{red!20}+0.55/+0.63 \\
\texttt{Mean\_Loyalty} & +0.00/+0.00 & +0.00/+0.00 & +0.00/+0.00 & +0.00/+0.00 & +0.00/+0.00 & +0.00/+0.00 & +0.00/+0.00 & +0.00/+0.00 & +0.00/+0.00 & +0.00/+0.00 & +0.00/+0.00 \\
\texttt{Mean\_Morale} & \cellcolor{yellow!20}-0.11/+0.03 & +0.00/+0.00 & \cellcolor{yellow!20}-0.21/+0.01 & +0.00/+0.00 & +0.00/+0.00 & -0.10/+0.11 & \cellcolor{orange!20}-0.31/-0.26 & \cellcolor{yellow!20}+0.18/+0.21 & \cellcolor{yellow!20}-0.23/-0.19 & -0.06/+0.16 & \cellcolor{yellow!20}-0.28/-0.28 \\
\texttt{Mean\_Stamina} & \cellcolor{yellow!20}-0.23/-0.11 & +0.00/+0.00 & \cellcolor{yellow!20}+0.22/+0.23 & +0.00/+0.00 & +0.00/+0.00 & \cellcolor{yellow!20}+0.25/+0.26 & \cellcolor{yellow!20}-0.21/-0.13 & \cellcolor{yellow!20}+0.13/+0.27 & \cellcolor{yellow!20}-0.28/-0.28 & \cellcolor{yellow!20}+0.26/+0.26 & \cellcolor{red!20}-0.54/-0.49 \\
\texttt{Mean\_Thirst} & +0.03/-0.12 & +0.00/+0.00 & \cellcolor{yellow!20}-0.30/-0.38 & +0.00/+0.00 & +0.00/+0.00 & \cellcolor{yellow!20}-0.25/-0.35 & \cellcolor{yellow!20}+0.17/+0.11 & \cellcolor{yellow!20}-0.19/-0.13 & \cellcolor{yellow!20}+0.29/+0.35 & \cellcolor{yellow!20}-0.25/-0.31 & \cellcolor{orange!20}+0.42/+0.39 \\
\texttt{Mean\_Trust} & +0.00/+0.00 & +0.00/+0.00 & +0.00/+0.00 & +0.00/+0.00 & +0.00/+0.00 & +0.00/+0.00 & +0.00/+0.00 & +0.00/+0.00 & +0.00/+0.00 & +0.00/+0.00 & +0.00/+0.00 \\
\hline
\end{tabular}
\end{adjustbox}
\end{table*}

\begin{table*}[htbp]
\centering
\caption{Pearson correlation matrix between per-character metrics in \textbf{\texttt{World Tiny}}.}
\label{tab:correlation_matrix_tiny}
\begin{adjustbox}{width=\textwidth}
\begin{tabular}{|l|c|c|c|c|c|c|c|c|c|c|c|c|c|c|c|c|}
\hline
\textbf{Variable} & \textbf{Mean\_Health} & \textbf{Mean\_Stamina} & \textbf{Mean\_Morale} & \textbf{Actions\_Total} & \textbf{Distinct\_Emotions} & \textbf{Distinct\_Goals} & \textbf{Distinct\_AI\_States} & \textbf{Emotion\_Transitions} & \textbf{Emotion\_Entropy} & \textbf{Goal\_Entropy} & \textbf{AI\_State\_Entropy} & \textbf{Mean\_Energy} & \textbf{Mean\_Hunger} & \textbf{Mean\_Loyalty} & \textbf{Mean\_Thirst} & \textbf{Mean\_Trust} \\
\hline
\texttt{Mean\_Health} & \cellcolor{gray!20}1.00 & 0.20 & 0.09 & 0.04 & 0.00 & 0.10 & \cellcolor{red!20}-0.73 & \cellcolor{blue!20}0.68 & \cellcolor{red!10}-0.21 & 0.11 & \cellcolor{red!20}-0.67 & 0.00 & \cellcolor{red!10}-0.33 & 0.00 & \cellcolor{red!10}-0.27 & 0.00 \\
\texttt{Mean\_Stamina} & 0.20 & \cellcolor{gray!20}1.00 & 0.14 & \cellcolor{red!10}-0.23 & 0.00 & \cellcolor{blue!10}0.25 & \cellcolor{red!10}-0.21 & 0.13 & \cellcolor{red!10}-0.28 & \cellcolor{blue!10}0.26 & \cellcolor{red!20}-0.54 & 0.00 & \cellcolor{red!10}-0.35 & 0.00 & \cellcolor{red!10}-0.47 & 0.00 \\
\texttt{Mean\_Morale} & 0.09 & 0.14 & \cellcolor{gray!20}1.00 & -0.11 & 0.00 & -0.10 & \cellcolor{red!10}-0.31 & 0.18 & \cellcolor{red!10}-0.23 & -0.06 & \cellcolor{red!10}-0.28 & 0.00 & \cellcolor{red!10}-0.48 & 0.00 & 0.00 & 0.00 \\
\texttt{Actions\_Total} & 0.04 & \cellcolor{red!10}-0.23 & -0.11 & \cellcolor{gray!20}1.00 & 0.00 & 0.08 & 0.08 & 0.01 & 0.03 & 0.08 & \cellcolor{blue!10}0.29 & 0.00 & 0.03 & 0.00 & 0.03 & 0.00 \\
\texttt{Distinct\_Emotions} & 0.00 & 0.00 & 0.00 & 0.00 & \cellcolor{gray!20}1.00 & 0.00 & 0.00 & 0.00 & 0.00 & 0.00 & 0.00 & 0.00 & 0.00 & 0.00 & 0.00 & 0.00 \\
\texttt{Distinct\_Goals} & 0.10 & \cellcolor{blue!10}0.25 & -0.10 & 0.08 & 0.00 & \cellcolor{gray!20}1.00 & -0.17 & -0.18 & \cellcolor{red!10}-0.31 & \cellcolor{blue!20}0.98 & \cellcolor{red!10}-0.22 & 0.00 & -0.14 & 0.00 & \cellcolor{red!10}-0.25 & 0.00 \\
\texttt{Distinct\_AI\_States} & \cellcolor{red!20}-0.73 & \cellcolor{red!10}-0.21 & \cellcolor{red!10}-0.31 & 0.08 & 0.00 & -0.17 & \cellcolor{gray!20}1.00 & \cellcolor{red!10}-0.40 & \cellcolor{blue!10}0.36 & -0.16 & \cellcolor{blue!20}0.83 & 0.00 & \cellcolor{blue!10}0.43 & 0.00 & 0.17 & 0.00 \\
\texttt{Emotion\_Transitions} & \cellcolor{blue!20}0.68 & 0.13 & 0.18 & 0.01 & 0.00 & -0.18 & \cellcolor{red!10}-0.40 & \cellcolor{gray!20}1.00 & \cellcolor{blue!10}0.37 & -0.17 & \cellcolor{red!10}-0.38 & 0.00 & -0.08 & 0.00 & -0.19 & 0.00 \\
\texttt{Emotion\_Entropy} & \cellcolor{red!10}-0.21 & \cellcolor{red!10}-0.28 & \cellcolor{red!10}-0.23 & 0.03 & 0.00 & \cellcolor{red!10}-0.31 & \cellcolor{blue!10}0.36 & \cellcolor{blue!10}0.37 & \cellcolor{gray!20}1.00 & \cellcolor{red!10}-0.28 & \cellcolor{blue!10}0.42 & 0.00 & \cellcolor{blue!20}0.54 & 0.00 & \cellcolor{blue!10}0.29 & 0.00 \\
\texttt{Goal\_Entropy} & 0.11 & \cellcolor{blue!10}0.26 & -0.06 & 0.08 & 0.00 & \cellcolor{blue!20}0.98 & -0.16 & -0.17 & \cellcolor{red!10}-0.28 & \cellcolor{gray!20}1.00 & \cellcolor{red!10}-0.20 & 0.00 & -0.17 & 0.00 & \cellcolor{red!10}-0.25 & 0.00 \\
\texttt{AI\_State\_Entropy} & \cellcolor{red!20}-0.67 & \cellcolor{red!20}-0.54 & \cellcolor{red!10}-0.28 & \cellcolor{blue!10}0.29 & 0.00 & \cellcolor{red!10}-0.22 & \cellcolor{blue!20}0.83 & \cellcolor{red!10}-0.38 & \cellcolor{blue!10}0.42 & \cellcolor{red!10}-0.20 & \cellcolor{gray!20}1.00 & 0.00 & \cellcolor{blue!20}0.55 & 0.00 & \cellcolor{blue!10}0.42 & 0.00 \\
\texttt{Mean\_Energy} & 0.00 & 0.00 & 0.00 & 0.00 & 0.00 & 0.00 & 0.00 & 0.00 & 0.00 & 0.00 & 0.00 & \cellcolor{gray!20}1.00 & 0.00 & 0.00 & 0.00 & 0.00 \\
\texttt{Mean\_Hunger} & \cellcolor{red!10}-0.33 & \cellcolor{red!10}-0.35 & \cellcolor{red!10}-0.48 & 0.03 & 0.00 & -0.14 & \cellcolor{blue!10}0.43 & -0.08 & \cellcolor{blue!20}0.54 & -0.17 & \cellcolor{blue!20}0.55 & 0.00 & \cellcolor{gray!20}1.00 & 0.00 & \cellcolor{blue!20}0.54 & 0.00 \\
\texttt{Mean\_Loyalty} & 0.00 & 0.00 & 0.00 & 0.00 & 0.00 & 0.00 & 0.00 & 0.00 & 0.00 & 0.00 & 0.00 & 0.00 & 0.00 & \cellcolor{gray!20}1.00 & 0.00 & 0.00 \\
\texttt{Mean\_Thirst} & \cellcolor{red!10}-0.27 & \cellcolor{red!10}-0.47 & 0.00 & 0.03 & 0.00 & \cellcolor{red!10}-0.25 & 0.17 & -0.19 & \cellcolor{blue!10}0.29 & \cellcolor{red!10}-0.25 & \cellcolor{blue!10}0.42 & 0.00 & \cellcolor{blue!20}0.54 & 0.00 & \cellcolor{gray!20}1.00 & 0.00 \\
\texttt{Mean\_Trust} & 0.00 & 0.00 & 0.00 & 0.00 & 0.00 & 0.00 & 0.00 & 0.00 & 0.00 & 0.00 & 0.00 & 0.00 & 0.00 & 0.00 & 0.00 & \cellcolor{gray!20}1.00 \\
\hline
\end{tabular}
\end{adjustbox}
\end{table*}

\begin{table*}[htbp]
\centering
\caption{Cross-world dynamics of character status variables (mean value) in \textbf{\texttt{World Small}}.}
\label{tab:cross_world_dynamics_small}
\begin{adjustbox}{width=0.6\textwidth}
\begin{tabular}{|l|c|c|c|c|c|}
\hline
\textbf{World} & \textbf{health} & \textbf{hunger} & \textbf{morale} & \textbf{stamina} & \textbf{thirst} \\
\hline
worldsmall1 &   66.3 & 2.5 & 83.8 & 94.3 & 4.2 \\
worldsmall2 &   73.8 & 4.7 & 78.2 & 93.9 & 5.8 \\
worldsmall3 &   80.1 & 3.9 & 84.6 & 94.7 & 3.8 \\
worldsmall4 &   80.7 & 3.4 & 84.4 & 94.1 & 5.1 \\
worldsmall5 &   79.5 & 3.1 & 72.1 & 95.2 & 3.9 \\
worldsmall6 &   78.3 & 4.2 & 88.7 & 93.2 & 4.9 \\
worldsmall7 &   71.8 & 3.7 & 80.4 & 91.6 & 4.1 \\
worldsmall8 &   71.5 & 4.2 & 82.7 & 93.3 & 5.7 \\
worldsmall9 &  81.1 & 4.2 & 79.7 & 93.3 & 3.9 \\
\hline
\end{tabular}
\end{adjustbox}
\end{table*}

\begin{table*}[htbp]
\centering
\caption{Status $\times$ activity correlations (n=72). Each cell shows Pearson $r$ / Spearman $\rho$ in \textbf{\texttt{World Small}}.}
\label{tab:status_activity_corr_small}
\begin{adjustbox}{width=\textwidth}
\begin{tabular}{|l|c|c|c|c|c|c|c|c|c|c|c|}
\hline
\textbf{Status} & \textbf{Actions\_Total} & \textbf{Emotions\_Total} & \textbf{Goals\_Total} & \textbf{AI\_States\_Total} & \textbf{Distinct\_Emotions} & \textbf{Distinct\_Goals} & \textbf{Distinct\_AI\_States} & \textbf{Emotion\_Transitions} & \textbf{Emotion\_Entropy} & \textbf{Goal\_Entropy} & \textbf{AI\_State\_Entropy} \\
\hline
\texttt{Mean\_Energy} & +0.00/+0.00 & +0.00/+0.00 & +0.00/+0.00 & +0.00/+0.00 & +0.00/+0.00 & +0.00/+0.00 & +0.00/+0.00 & +0.00/+0.00 & +0.00/+0.00 & +0.00/+0.00 & +0.00/+0.00 \\
\texttt{Mean\_Health} & \cellcolor{orange!20}-0.36/-0.33 & +0.00/+0.00 & -0.06/-0.05 & +0.00/+0.00 & +0.00/+0.00 & -0.05/-0.04 & \cellcolor{red!20}-0.68/-0.69 & \cellcolor{red!20}+0.69/+0.61 & \cellcolor{yellow!20}-0.21/-0.46 & -0.06/-0.04 & \cellcolor{red!20}-0.67/-0.69 \\
\texttt{Mean\_Hunger} & +0.08/+0.06 & +0.00/+0.00 & +0.01/-0.01 & +0.00/+0.00 & +0.00/+0.00 & +0.07/+0.06 & \cellcolor{red!20}+0.51/+0.51 & \cellcolor{yellow!20}-0.20/-0.28 & \cellcolor{yellow!20}+0.26/+0.14 & +0.05/+0.11 & \cellcolor{red!20}+0.56/+0.55 \\
\texttt{Mean\_Loyalty} & +0.00/+0.00 & +0.00/+0.00 & +0.00/+0.00 & +0.00/+0.00 & +0.00/+0.00 & +0.00/+0.00 & +0.00/+0.00 & +0.00/+0.00 & +0.00/+0.00 & +0.00/+0.00 & +0.00/+0.00 \\
\texttt{Mean\_Morale} & +0.03/+0.10 & +0.00/+0.00 & \cellcolor{yellow!20}-0.11/-0.09 & +0.00/+0.00 & +0.00/+0.00 & -0.04/-0.07 & \cellcolor{yellow!20}-0.21/-0.14 & \cellcolor{yellow!20}+0.14/+0.05 & -0.05/-0.09 & -0.05/-0.07 & \cellcolor{yellow!20}-0.19/-0.11 \\
\texttt{Mean\_Stamina} & -0.07/-0.07 & +0.00/+0.00 & \cellcolor{yellow!20}-0.11/-0.09 & +0.00/+0.00 & +0.00/+0.00 & -0.05/-0.02 & \cellcolor{red!20}-0.54/-0.54 & \cellcolor{yellow!20}+0.28/+0.43 & \cellcolor{yellow!20}-0.29/-0.29 & -0.02/+0.00 & \cellcolor{red!20}-0.61/-0.58 \\
\texttt{Mean\_Thirst} & \cellcolor{yellow!20}+0.14/+0.13 & +0.00/+0.00 & +0.05/+0.10 & +0.00/+0.00 & +0.00/+0.00 & +0.01/+0.07 & \cellcolor{orange!20}+0.41/+0.42 & \cellcolor{yellow!20}-0.22/-0.34 & \cellcolor{yellow!20}+0.11/-0.02 & +0.03/+0.07 & \cellcolor{orange!20}+0.44/+0.43 \\
\texttt{Mean\_Trust} & +0.00/+0.00 & +0.00/+0.00 & +0.00/+0.00 & +0.00/+0.00 & +0.00/+0.00 & +0.00/+0.00 & +0.00/+0.00 & +0.00/+0.00 & +0.00/+0.00 & +0.00/+0.00 & +0.00/+0.00 \\
\hline
\end{tabular}
\end{adjustbox}
\end{table*}

\begin{table*}[htbp]
\centering
\caption{Pearson correlation matrix between per-character metrics in \textbf{\texttt{World Small}}.}
\label{tab:correlation_matrix}
\begin{adjustbox}{width=\textwidth}
\begin{tabular}{|l|c|c|c|c|c|c|c|c|c|c|c|c|c|c|c|c|}
\hline
\textbf{Variable} & \textbf{Mean\_Health} & \textbf{Mean\_Stamina} & \textbf{Mean\_Morale} & \textbf{Actions\_Total} & \textbf{Distinct\_Emotions} & \textbf{Distinct\_Goals} & \textbf{Distinct\_AI\_States} & \textbf{Emotion\_Transitions} & \textbf{Emotion\_Entropy} & \textbf{Goal\_Entropy} & \textbf{AI\_State\_Entropy} & \textbf{Mean\_Energy} & \textbf{Mean\_Hunger} & \textbf{Mean\_Loyalty} & \textbf{Mean\_Thirst} & \textbf{Mean\_Trust} \\
\hline
\texttt{Mean\_Health} & \cellcolor{gray!20}1.00 & \cellcolor{blue!10}0.34 & 0.10 & \cellcolor{red!10}-0.36 & 0.00 & -0.05 & \cellcolor{red!20}-0.68 & \cellcolor{blue!20}0.69 & \cellcolor{red!10}-0.21 & -0.06 & \cellcolor{red!20}-0.67 & 0.00 & \cellcolor{red!10}-0.37 & 0.00 & \cellcolor{red!10}-0.35 & 0.00 \\
\texttt{Mean\_Stamina} & \cellcolor{blue!10}0.34 & \cellcolor{gray!20}1.00 & -0.01 & -0.07 & 0.00 & -0.05 & \cellcolor{red!20}-0.54 & \cellcolor{blue!10}0.28 & \cellcolor{red!10}-0.29 & -0.02 & \cellcolor{red!20}-0.61 & 0.00 & \cellcolor{red!10}-0.38 & 0.00 & \cellcolor{red!10}-0.34 & 0.00 \\
\texttt{Mean\_Morale} & 0.10 & -0.01 & \cellcolor{gray!20}1.00 & 0.03 & 0.00 & -0.04 & \cellcolor{red!10}-0.21 & 0.14 & -0.05 & -0.05 & -0.19 & 0.00 & 0.03 & 0.00 & 0.08 & 0.00 \\
\texttt{Actions\_Total} & \cellcolor{red!10}-0.36 & -0.07 & 0.03 & \cellcolor{gray!20}1.00 & 0.00 & 0.13 & 0.06 & \cellcolor{red!10}-0.28 & -0.02 & 0.13 & 0.12 & 0.00 & 0.08 & 0.00 & 0.14 & 0.00 \\
\texttt{Distinct\_Emotions} & 0.00 & 0.00 & 0.00 & 0.00 & \cellcolor{gray!20}1.00 & 0.00 & 0.00 & 0.00 & 0.00 & 0.00 & 0.00 & 0.00 & 0.00 & 0.00 & 0.00 & 0.00 \\
\texttt{Distinct\_Goals} & -0.05 & -0.05 & -0.04 & 0.13 & 0.00 & \cellcolor{gray!20}1.00 & 0.09 & -0.15 & -0.09 & \cellcolor{blue!20}0.98 & 0.15 & 0.00 & 0.07 & 0.00 & 0.01 & 0.00 \\
\texttt{Distinct\_AI\_States} & \cellcolor{red!20}-0.68 & \cellcolor{red!20}-0.54 & \cellcolor{red!10}-0.21 & 0.06 & 0.00 & 0.09 & \cellcolor{gray!20}1.00 & \cellcolor{red!20}-0.55 & 0.15 & 0.09 & \cellcolor{blue!20}0.94 & 0.00 & \cellcolor{blue!20}0.51 & 0.00 & \cellcolor{blue!10}0.41 & 0.00 \\
\texttt{Emotion\_Transitions} & \cellcolor{blue!20}0.69 & \cellcolor{blue!10}0.28 & 0.14 & \cellcolor{red!10}-0.28 & 0.00 & -0.15 & \cellcolor{red!20}-0.55 & \cellcolor{gray!20}1.00 & \cellcolor{blue!10}0.38 & -0.15 & \cellcolor{red!10}-0.49 & 0.00 & \cellcolor{red!10}-0.20 & 0.00 & \cellcolor{red!10}-0.22 & 0.00 \\
\texttt{Emotion\_Entropy} & \cellcolor{red!10}-0.21 & \cellcolor{red!10}-0.29 & -0.05 & -0.02 & 0.00 & -0.09 & 0.15 & \cellcolor{blue!10}0.38 & \cellcolor{gray!20}1.00 & -0.11 & \cellcolor{blue!10}0.23 & 0.00 & \cellcolor{blue!10}0.26 & 0.00 & 0.11 & 0.00 \\
\texttt{Goal\_Entropy} & -0.06 & -0.02 & -0.05 & 0.13 & 0.00 & \cellcolor{blue!20}0.98 & 0.09 & -0.15 & -0.11 & \cellcolor{gray!20}1.00 & 0.14 & 0.00 & 0.05 & 0.00 & 0.03 & 0.00 \\
\texttt{AI\_State\_Entropy} & \cellcolor{red!20}-0.67 & \cellcolor{red!20}-0.61 & -0.19 & 0.12 & 0.00 & 0.15 & \cellcolor{blue!20}0.94 & \cellcolor{red!10}-0.49 & \cellcolor{blue!10}0.23 & 0.14 & \cellcolor{gray!20}1.00 & 0.00 & \cellcolor{blue!20}0.56 & 0.00 & \cellcolor{blue!10}0.44 & 0.00 \\
\texttt{Mean\_Energy} & 0.00 & 0.00 & 0.00 & 0.00 & 0.00 & 0.00 & 0.00 & 0.00 & 0.00 & 0.00 & 0.00 & \cellcolor{gray!20}1.00 & 0.00 & 0.00 & 0.00 & 0.00 \\
\texttt{Mean\_Hunger} & \cellcolor{red!10}-0.37 & \cellcolor{red!10}-0.38 & 0.03 & 0.08 & 0.00 & 0.07 & \cellcolor{blue!20}0.51 & \cellcolor{red!10}-0.20 & \cellcolor{blue!10}0.26 & 0.05 & \cellcolor{blue!20}0.56 & 0.00 & \cellcolor{gray!20}1.00 & 0.00 & \cellcolor{blue!10}0.45 & 0.00 \\
\texttt{Mean\_Loyalty} & 0.00 & 0.00 & 0.00 & 0.00 & 0.00 & 0.00 & 0.00 & 0.00 & 0.00 & 0.00 & 0.00 & 0.00 & 0.00 & \cellcolor{gray!20}1.00 & 0.00 & 0.00 \\
\texttt{Mean\_Thirst} & \cellcolor{red!10}-0.35 & \cellcolor{red!10}-0.34 & 0.08 & 0.14 & 0.00 & 0.01 & \cellcolor{blue!10}0.41 & \cellcolor{red!10}-0.22 & 0.11 & 0.03 & \cellcolor{blue!10}0.44 & 0.00 & \cellcolor{blue!10}0.45 & 0.00 & \cellcolor{gray!20}1.00 & 0.00 \\
\texttt{Mean\_Trust} & 0.00 & 0.00 & 0.00 & 0.00 & 0.00 & 0.00 & 0.00 & 0.00 & 0.00 & 0.00 & 0.00 & 0.00 & 0.00 & 0.00 & 0.00 & \cellcolor{gray!20}1.00 \\
\hline
\end{tabular}
\end{adjustbox}
\end{table*}

\begin{table*}[htbp]
\centering
\caption{Cross-world dynamics of character status variables (mean value) in \textbf{\texttt{World Medium}}.}
\label{tab:cross_world_dynamics_medium}
\begin{adjustbox}{width=0.6\textwidth}
\begin{tabular}{|l|c|c|c|c|c|}
\hline
\textbf{World} & \textbf{health} & \textbf{hunger} & \textbf{morale} & \textbf{stamina} & \textbf{thirst} \\
\hline
worldmedium1 &  74.5 & 3.5 & 81.5 & 92.1 & 5.6 \\
worldmedium2 &  70.1 & 4.7 & 78.4 & 92.4 & 6.1 \\
worldmedium3 &  65.7 & 4.2 & 78.6 & 93.6 & 5.4 \\
worldmedium4 & 71.8 & 4.1 & 81.1 & 90.9 & 5.1 \\
worldmedium5 & 71.3 & 3.6 & 83.9 & 92.3 & 5.4 \\
worldmedium6 &  70.2 & 5.1 & 81.3 & 92.9 & 5.7 \\
worldmedium7 & 75.2 & 3.5 & 85.2 & 94.4 & 5.5 \\
worldmedium8 & 74.0 & 4.5 & 78.0 & 91.6 & 5.5 \\
worldmedium9 & 71.1 & 4.4 & 81.2 & 91.3 & 6.8 \\
\hline
\end{tabular}
\end{adjustbox}
\end{table*}

\begin{table*}[htbp]
\centering
\caption{Status $\times$ activity correlations (n=144). Each cell shows Pearson $r$ / Spearman $\rho$ in \textbf{\texttt{World Medium}}.}
\label{tab:status_activity_corr_medium}
\begin{adjustbox}{width=\textwidth}
\begin{tabular}{|l|c|c|c|c|c|c|c|c|c|c|c|}
\hline
\textbf{Status} & \textbf{Actions\_Total} & \textbf{Emotions\_Total} & \textbf{Goals\_Total} & \textbf{AI\_States\_Total} & \textbf{Distinct\_Emotions} & \textbf{Distinct\_Goals} & \textbf{Distinct\_AI\_States} & \textbf{Emotion\_Transitions} & \textbf{Emotion\_Entropy} & \textbf{Goal\_Entropy} & \textbf{AI\_State\_Entropy} \\
\hline
\texttt{Mean\_Energy} & +0.00/+0.00 & +0.00/+0.00 & +0.00/+0.00 & +0.00/+0.00 & +0.00/+0.00 & +0.00/+0.00 & +0.00/+0.00 & +0.00/+0.00 & +0.00/+0.00 & +0.00/+0.00 & +0.00/+0.00 \\
\texttt{Mean\_Health} & +0.07/+0.08 & +0.00/+0.00 & +0.02/+0.02 & +0.00/+0.00 & +0.00/+0.00 & +0.04/+0.05 & \cellcolor{red!20}-0.65/-0.65 & \cellcolor{red!20}+0.80/+0.75 & +0.01/-0.25 & +0.02/+0.07 & \cellcolor{red!20}-0.64/-0.66 \\
\texttt{Mean\_Hunger} & +0.01/-0.03 & +0.00/+0.00 & \cellcolor{yellow!20}+0.13/+0.16 & +0.00/+0.00 & +0.00/+0.00 & \cellcolor{yellow!20}+0.13/+0.15 & \cellcolor{orange!20}+0.47/+0.49 & \cellcolor{yellow!20}-0.14/-0.19 & \cellcolor{orange!20}+0.33/+0.43 & \cellcolor{yellow!20}+0.13/+0.14 & \cellcolor{orange!20}+0.50/+0.57 \\
\texttt{Mean\_Loyalty} & +0.00/+0.00 & +0.00/+0.00 & +0.00/+0.00 & +0.00/+0.00 & +0.00/+0.00 & +0.00/+0.00 & +0.00/+0.00 & +0.00/+0.00 & +0.00/+0.00 & +0.00/+0.00 & +0.00/+0.00 \\
\texttt{Mean\_Morale} & +0.04/-0.03 & +0.00/+0.00 & -0.01/-0.04 & +0.00/+0.00 & +0.00/+0.00 & -0.02/-0.05 & \cellcolor{yellow!20}-0.12/-0.12 & \cellcolor{yellow!20}+0.19/+0.19 & +0.02/-0.01 & -0.00/-0.03 & \cellcolor{yellow!20}-0.19/-0.16 \\
\texttt{Mean\_Stamina} & -0.02/-0.04 & +0.00/+0.00 & -0.06/-0.03 & +0.00/+0.00 & +0.00/+0.00 & -0.07/-0.03 & \cellcolor{red!20}-0.54/-0.51 & \cellcolor{yellow!20}+0.10/+0.17 & \cellcolor{orange!20}-0.42/-0.51 & -0.04/-0.02 & \cellcolor{red!20}-0.65/-0.64 \\
\texttt{Mean\_Thirst} & +0.03/+0.12 & +0.00/+0.00 & \cellcolor{yellow!20}+0.13/+0.09 & +0.00/+0.00 & +0.00/+0.00 & \cellcolor{yellow!20}+0.17/+0.10 & \cellcolor{orange!20}+0.44/+0.44 & -0.08/-0.13 & \cellcolor{yellow!20}+0.29/+0.34 & \cellcolor{yellow!20}+0.15/+0.10 & \cellcolor{red!20}+0.51/+0.53 \\
\texttt{Mean\_Trust} & +0.00/+0.00 & +0.00/+0.00 & +0.00/+0.00 & +0.00/+0.00 & +0.00/+0.00 & +0.00/+0.00 & +0.00/+0.00 & +0.00/+0.00 & +0.00/+0.00 & +0.00/+0.00 & +0.00/+0.00 \\
\hline
\end{tabular}
\end{adjustbox}
\end{table*}

\begin{table*}[htbp]
\centering
\caption{Pearson correlation matrix between per-character metrics in \textbf{\texttt{World Medium}}.}
\label{tab:correlation_matrix_medium}
\begin{adjustbox}{width=\textwidth}
\begin{tabular}{|l|c|c|c|c|c|c|c|c|c|c|c|c|c|c|c|c|}
\hline
\textbf{Variable} & \textbf{Mean\_Health} & \textbf{Mean\_Stamina} & \textbf{Mean\_Morale} & \textbf{Actions\_Total} & \textbf{Distinct\_Emotions} & \textbf{Distinct\_Goals} & \textbf{Distinct\_AI\_States} & \textbf{Emotion\_Transitions} & \textbf{Emotion\_Entropy} & \textbf{Goal\_Entropy} & \textbf{AI\_State\_Entropy} & \textbf{Mean\_Energy} & \textbf{Mean\_Hunger} & \textbf{Mean\_Loyalty} & \textbf{Mean\_Thirst} & \textbf{Mean\_Trust} \\
\hline
\texttt{Mean\_Health} & \cellcolor{gray!20}1.00 & \cellcolor{blue!10}0.27 & 0.16 & 0.07 & 0.00 & 0.04 & \cellcolor{red!20}-0.65 & \cellcolor{blue!20}0.80 & 0.01 & 0.02 & \cellcolor{red!20}-0.64 & 0.00 & \cellcolor{red!10}-0.33 & 0.00 & \cellcolor{red!10}-0.24 & 0.00 \\
\texttt{Mean\_Stamina} & \cellcolor{blue!10}0.27 & \cellcolor{gray!20}1.00 & \cellcolor{blue!10}0.20 & -0.02 & 0.00 & -0.07 & \cellcolor{red!20}-0.54 & 0.10 & \cellcolor{red!10}-0.42 & -0.04 & \cellcolor{red!20}-0.65 & 0.00 & \cellcolor{red!10}-0.41 & 0.00 & \cellcolor{red!20}-0.55 & 0.00 \\
\texttt{Mean\_Morale} & 0.16 & \cellcolor{blue!10}0.20 & \cellcolor{gray!20}1.00 & 0.04 & 0.00 & -0.02 & -0.12 & 0.19 & 0.02 & -0.00 & -0.19 & 0.00 & -0.17 & 0.00 & -0.18 & 0.00 \\
\texttt{Actions\_Total} & 0.07 & -0.02 & 0.04 & \cellcolor{gray!20}1.00 & 0.00 & 0.04 & 0.00 & 0.11 & 0.08 & 0.04 & -0.04 & 0.00 & 0.01 & 0.00 & 0.03 & 0.00 \\
\texttt{Distinct\_Emotions} & 0.00 & 0.00 & 0.00 & 0.00 & \cellcolor{gray!20}1.00 & 0.00 & 0.00 & 0.00 & 0.00 & 0.00 & 0.00 & 0.00 & 0.00 & 0.00 & 0.00 & 0.00 \\
\texttt{Distinct\_Goals} & 0.04 & -0.07 & -0.02 & 0.04 & 0.00 & \cellcolor{gray!20}1.00 & 0.05 & 0.01 & -0.04 & \cellcolor{blue!20}0.97 & 0.04 & 0.00 & 0.13 & 0.00 & 0.17 & 0.00 \\
\texttt{Distinct\_AI\_States} & \cellcolor{red!20}-0.65 & \cellcolor{red!20}-0.54 & -0.12 & 0.00 & 0.00 & 0.05 & \cellcolor{gray!20}1.00 & \cellcolor{red!10}-0.45 & \cellcolor{blue!10}0.24 & 0.03 & \cellcolor{blue!20}0.92 & 0.00 & \cellcolor{blue!10}0.47 & 0.00 & \cellcolor{blue!10}0.44 & 0.00 \\
\texttt{Emotion\_Transitions} & \cellcolor{blue!20}0.80 & 0.10 & 0.19 & 0.11 & 0.00 & 0.01 & \cellcolor{red!10}-0.45 & \cellcolor{gray!20}1.00 & \cellcolor{blue!10}0.43 & -0.00 & \cellcolor{red!10}-0.43 & 0.00 & -0.14 & 0.00 & -0.08 & 0.00 \\
\texttt{Emotion\_Entropy} & 0.01 & \cellcolor{red!10}-0.42 & 0.02 & 0.08 & 0.00 & -0.04 & \cellcolor{blue!10}0.24 & \cellcolor{blue!10}0.43 & \cellcolor{gray!20}1.00 & -0.07 & \cellcolor{blue!10}0.29 & 0.00 & \cellcolor{blue!10}0.33 & 0.00 & \cellcolor{blue!10}0.29 & 0.00 \\
\texttt{Goal\_Entropy} & 0.02 & -0.04 & -0.00 & 0.04 & 0.00 & \cellcolor{blue!20}0.97 & 0.03 & -0.00 & -0.07 & \cellcolor{gray!20}1.00 & 0.01 & 0.00 & 0.13 & 0.00 & 0.15 & 0.00 \\
\texttt{AI\_State\_Entropy} & \cellcolor{red!20}-0.64 & \cellcolor{red!20}-0.65 & -0.19 & -0.04 & 0.00 & 0.04 & \cellcolor{blue!20}0.92 & \cellcolor{red!10}-0.43 & \cellcolor{blue!10}0.29 & 0.01 & \cellcolor{gray!20}1.00 & 0.00 & \cellcolor{blue!10}0.50 & 0.00 & \cellcolor{blue!20}0.51 & 0.00 \\
\texttt{Mean\_Energy} & 0.00 & 0.00 & 0.00 & 0.00 & 0.00 & 0.00 & 0.00 & 0.00 & 0.00 & 0.00 & 0.00 & \cellcolor{gray!20}1.00 & 0.00 & 0.00 & 0.00 & 0.00 \\
\texttt{Mean\_Hunger} & \cellcolor{red!10}-0.33 & \cellcolor{red!10}-0.41 & -0.17 & 0.01 & 0.00 & 0.13 & \cellcolor{blue!10}0.47 & -0.14 & \cellcolor{blue!10}0.33 & 0.13 & \cellcolor{blue!10}0.50 & 0.00 & \cellcolor{gray!20}1.00 & 0.00 & \cellcolor{blue!10}0.34 & 0.00 \\
\texttt{Mean\_Loyalty} & 0.00 & 0.00 & 0.00 & 0.00 & 0.00 & 0.00 & 0.00 & 0.00 & 0.00 & 0.00 & 0.00 & 0.00 & 0.00 & \cellcolor{gray!20}1.00 & 0.00 & 0.00 \\
\texttt{Mean\_Thirst} & \cellcolor{red!10}-0.24 & \cellcolor{red!20}-0.55 & -0.18 & 0.03 & 0.00 & 0.17 & \cellcolor{blue!10}0.44 & -0.08 & \cellcolor{blue!10}0.29 & 0.15 & \cellcolor{blue!20}0.51 & 0.00 & \cellcolor{blue!10}0.34 & 0.00 & \cellcolor{gray!20}1.00 & 0.00 \\
\texttt{Mean\_Trust} & 0.00 & 0.00 & 0.00 & 0.00 & 0.00 & 0.00 & 0.00 & 0.00 & 0.00 & 0.00 & 0.00 & 0.00 & 0.00 & 0.00 & 0.00 & \cellcolor{gray!20}1.00 \\
\hline
\end{tabular}
\end{adjustbox}
\end{table*}

\begin{table*}[htbp]
\centering
\caption{Cross-world dynamics of character status variables (mean value) in \textbf{\texttt{World Large}}.}
\label{tab:cross_world_dynamics_large}
\begin{adjustbox}{width=0.6\textwidth}
\begin{tabular}{|l|c|c|c|c|c|}
\hline
\textbf{World} &  \textbf{health} & \textbf{hunger} & \textbf{morale} & \textbf{stamina} & \textbf{thirst} \\
\hline
worldlarge1 &  70.0 & 4.1 & 80.3 & 94.3 & 4.9 \\
worldlarge2 &  70.0 & 3.5 & 78.4 & 93.5 & 5.7 \\
worldlarge3 &  71.9 & 3.8 & 81.3 & 93.3 & 5.6 \\
worldlarge4 &  77.7 & 3.6 & 83.0 & 93.2 & 4.8 \\
worldlarge5 &  73.5 & 3.4 & 80.3 & 93.2 & 5.6 \\
worldlarge6 &  75.3 & 3.1 & 79.8 & 94.6 & 4.9 \\
worldlarge7 &  71.5 & 4.2 & 81.4 & 93.5 & 5.2 \\
worldlarge8 &  67.8 & 4.2 & 81.5 & 92.5 & 6.5 \\
worldlarge9 &  73.5 & 4.3 & 80.8 & 93.0 & 5.5 \\
\hline
\end{tabular}
\end{adjustbox}
\end{table*}

\begin{table*}[htbp]
\centering
\caption{Status $\times$ activity correlations (n=288). Each cell shows Pearson $r$ / Spearman $\rho$ in \textbf{\texttt{World Large}}.}
\label{tab:status_activity_corr_large}
\begin{adjustbox}{width=\textwidth}
\begin{tabular}{|l|c|c|c|c|c|c|c|c|c|c|c|}
\hline
\textbf{Status} & \textbf{Actions\_Total} & \textbf{Emotions\_Total} & \textbf{Goals\_Total} & \textbf{AI\_States\_Total} & \textbf{Distinct\_Emotions} & \textbf{Distinct\_Goals} & \textbf{Distinct\_AI\_States} & \textbf{Emotion\_Transitions} & \textbf{Emotion\_Entropy} & \textbf{Goal\_Entropy} & \textbf{AI\_State\_Entropy} \\
\hline
\texttt{Mean\_Energy} & +0.00/+0.00 & +0.00/+0.00 & +0.00/+0.00 & +0.00/+0.00 & +0.00/+0.00 & +0.00/+0.00 & +0.00/+0.00 & +0.00/+0.00 & +0.00/+0.00 & +0.00/+0.00 & +0.00/+0.00 \\
\texttt{Mean\_Health} & +0.01/-0.01 & +0.00/+0.00 & -0.01/-0.01 & +0.00/+0.00 & +0.00/+0.00 & +0.02/+0.02 & \cellcolor{red!20}-0.67/-0.68 & \cellcolor{red!20}+0.77/+0.74 & \cellcolor{yellow!20}-0.20/-0.36 & +0.03/+0.04 & \cellcolor{red!20}-0.66/-0.69 \\
\texttt{Mean\_Hunger} & +0.03/+0.02 & +0.00/+0.00 & +0.04/+0.05 & +0.00/+0.00 & +0.00/+0.00 & -0.01/+0.00 & \cellcolor{red!20}+0.50/+0.53 & \cellcolor{orange!20}-0.30/-0.30 & \cellcolor{yellow!20}+0.20/+0.27 & -0.01/-0.01 & \cellcolor{red!20}+0.54/+0.60 \\
\texttt{Mean\_Loyalty} & +0.00/+0.00 & +0.00/+0.00 & +0.00/+0.00 & +0.00/+0.00 & +0.00/+0.00 & +0.00/+0.00 & +0.00/+0.00 & +0.00/+0.00 & +0.00/+0.00 & +0.00/+0.00 & +0.00/+0.00 \\
\texttt{Mean\_Morale} & +0.08/+0.11 & +0.00/+0.00 & +0.00/+0.01 & +0.00/+0.00 & +0.00/+0.00 & +0.07/+0.06 & -0.07/-0.12 & \cellcolor{yellow!20}+0.11/+0.15 & -0.03/-0.02 & +0.09/+0.09 & -0.07/-0.12 \\
\texttt{Mean\_Stamina} & -0.05/-0.06 & +0.00/+0.00 & +0.04/-0.01 & +0.00/+0.00 & +0.00/+0.00 & +0.04/-0.00 & \cellcolor{red!20}-0.50/-0.50 & \cellcolor{yellow!20}+0.25/+0.29 & \cellcolor{orange!20}-0.38/-0.38 & +0.04/+0.03 & \cellcolor{red!20}-0.62/-0.60 \\
\texttt{Mean\_Thirst} & +0.02/+0.02 & +0.00/+0.00 & -0.03/-0.03 & +0.00/+0.00 & +0.00/+0.00 & -0.08/-0.06 & \cellcolor{orange!20}+0.41/+0.41 & \cellcolor{yellow!20}-0.24/-0.19 & \cellcolor{yellow!20}+0.26/+0.27 & -0.08/-0.08 & \cellcolor{orange!20}+0.47/+0.46 \\
\texttt{Mean\_Trust} & +0.00/+0.00 & +0.00/+0.00 & +0.00/+0.00 & +0.00/+0.00 & +0.00/+0.00 & +0.00/+0.00 & +0.00/+0.00 & +0.00/+0.00 & +0.00/+0.00 & +0.00/+0.00 & +0.00/+0.00 \\
\hline
\end{tabular}
\end{adjustbox}
\end{table*}

\begin{table*}[htbp]
\centering
\caption{Pearson correlation matrix between per-character metrics in \textbf{\texttt{World Large}}.}
\label{tab:correlation_matrix_large}
\begin{adjustbox}{width=\textwidth}
\begin{tabular}{|l|c|c|c|c|c|c|c|c|c|c|c|c|c|c|c|c|}
\hline
\textbf{Variable} & \textbf{Mean\_Health} & \textbf{Mean\_Stamina} & \textbf{Mean\_Morale} & \textbf{Actions\_Total} & \textbf{Distinct\_Emotions} & \textbf{Distinct\_Goals} & \textbf{Distinct\_AI\_States} & \textbf{Emotion\_Transitions} & \textbf{Emotion\_Entropy} & \textbf{Goal\_Entropy} & \textbf{AI\_State\_Entropy} & \textbf{Mean\_Energy} & \textbf{Mean\_Hunger} & \textbf{Mean\_Loyalty} & \textbf{Mean\_Thirst} & \textbf{Mean\_Trust} \\
\hline
\texttt{Mean\_Health} & \cellcolor{gray!20}1.00 & \cellcolor{blue!10}0.33 & 0.12 & 0.01 & 0.00 & 0.02 & \cellcolor{red!20}-0.67 & \cellcolor{blue!20}0.77 & \cellcolor{red!10}-0.20 & 0.03 & \cellcolor{red!20}-0.66 & 0.00 & \cellcolor{red!10}-0.43 & 0.00 & \cellcolor{red!10}-0.30 & 0.00 \\
\texttt{Mean\_Stamina} & \cellcolor{blue!10}0.33 & \cellcolor{gray!20}1.00 & 0.02 & -0.05 & 0.00 & 0.04 & \cellcolor{red!20}-0.50 & \cellcolor{blue!10}0.25 & \cellcolor{red!10}-0.38 & 0.04 & \cellcolor{red!20}-0.62 & 0.00 & \cellcolor{red!10}-0.43 & 0.00 & \cellcolor{red!10}-0.40 & 0.00 \\
\texttt{Mean\_Morale} & 0.12 & 0.02 & \cellcolor{gray!20}1.00 & 0.08 & 0.00 & 0.07 & -0.07 & 0.11 & -0.03 & 0.09 & -0.07 & 0.00 & -0.05 & 0.00 & 0.00 & 0.00 \\
\texttt{Actions\_Total} & 0.01 & -0.05 & 0.08 & \cellcolor{gray!20}1.00 & 0.00 & 0.00 & -0.08 & 0.03 & 0.08 & 0.02 & -0.04 & 0.00 & 0.03 & 0.00 & 0.02 & 0.00 \\
\texttt{Distinct\_Emotions} & 0.00 & 0.00 & 0.00 & 0.00 & \cellcolor{gray!20}1.00 & 0.00 & 0.00 & 0.00 & 0.00 & 0.00 & 0.00 & 0.00 & 0.00 & 0.00 & 0.00 & 0.00 \\
\texttt{Distinct\_Goals} & 0.02 & 0.04 & 0.07 & 0.00 & 0.00 & \cellcolor{gray!20}1.00 & 0.02 & -0.02 & -0.04 & \cellcolor{blue!20}0.97 & 0.02 & 0.00 & -0.01 & 0.00 & -0.08 & 0.00 \\
\texttt{Distinct\_AI\_States} & \cellcolor{red!20}-0.67 & \cellcolor{red!20}-0.50 & -0.07 & -0.08 & 0.00 & 0.02 & \cellcolor{gray!20}1.00 & \cellcolor{red!10}-0.50 & \cellcolor{blue!10}0.27 & 0.01 & \cellcolor{blue!20}0.93 & 0.00 & \cellcolor{blue!20}0.50 & 0.00 & \cellcolor{blue!10}0.41 & 0.00 \\
\texttt{Emotion\_Transitions} & \cellcolor{blue!20}0.77 & \cellcolor{blue!10}0.25 & 0.11 & 0.03 & 0.00 & -0.02 & \cellcolor{red!10}-0.50 & \cellcolor{gray!20}1.00 & \cellcolor{blue!10}0.26 & 0.01 & \cellcolor{red!10}-0.48 & 0.00 & \cellcolor{red!10}-0.30 & 0.00 & \cellcolor{red!10}-0.24 & 0.00 \\
\texttt{Emotion\_Entropy} & \cellcolor{red!10}-0.20 & \cellcolor{red!10}-0.38 & -0.03 & 0.08 & 0.00 & -0.04 & \cellcolor{blue!10}0.27 & \cellcolor{blue!10}0.26 & \cellcolor{gray!20}1.00 & -0.03 & \cellcolor{blue!10}0.31 & 0.00 & 0.20 & 0.00 & \cellcolor{blue!10}0.26 & 0.00 \\
\texttt{Goal\_Entropy} & 0.03 & 0.04 & 0.09 & 0.02 & 0.00 & \cellcolor{blue!20}0.97 & 0.01 & 0.01 & -0.03 & \cellcolor{gray!20}1.00 & 0.02 & 0.00 & -0.01 & 0.00 & -0.08 & 0.00 \\
\texttt{AI\_State\_Entropy} & \cellcolor{red!20}-0.66 & \cellcolor{red!20}-0.62 & -0.07 & -0.04 & 0.00 & 0.02 & \cellcolor{blue!20}0.93 & \cellcolor{red!10}-0.48 & \cellcolor{blue!10}0.31 & 0.02 & \cellcolor{gray!20}1.00 & 0.00 & \cellcolor{blue!20}0.54 & 0.00 & \cellcolor{blue!10}0.47 & 0.00 \\
\texttt{Mean\_Energy} & 0.00 & 0.00 & 0.00 & 0.00 & 0.00 & 0.00 & 0.00 & 0.00 & 0.00 & 0.00 & 0.00 & \cellcolor{gray!20}1.00 & 0.00 & 0.00 & 0.00 & 0.00 \\
\texttt{Mean\_Hunger} & \cellcolor{red!10}-0.43 & \cellcolor{red!10}-0.43 & -0.05 & 0.03 & 0.00 & -0.01 & \cellcolor{blue!20}0.50 & \cellcolor{red!10}-0.30 & 0.20 & -0.01 & \cellcolor{blue!20}0.54 & 0.00 & \cellcolor{gray!20}1.00 & 0.00 & \cellcolor{blue!10}0.31 & 0.00 \\
\texttt{Mean\_Loyalty} & 0.00 & 0.00 & 0.00 & 0.00 & 0.00 & 0.00 & 0.00 & 0.00 & 0.00 & 0.00 & 0.00 & 0.00 & 0.00 & \cellcolor{gray!20}1.00 & 0.00 & 0.00 \\
\texttt{Mean\_Thirst} & \cellcolor{red!10}-0.30 & \cellcolor{red!10}-0.40 & 0.00 & 0.02 & 0.00 & -0.08 & \cellcolor{blue!10}0.41 & \cellcolor{red!10}-0.24 & \cellcolor{blue!10}0.26 & -0.08 & \cellcolor{blue!10}0.47 & 0.00 & \cellcolor{blue!10}0.31 & 0.00 & \cellcolor{gray!20}1.00 & 0.00 \\
\texttt{Mean\_Trust} & 0.00 & 0.00 & 0.00 & 0.00 & 0.00 & 0.00 & 0.00 & 0.00 & 0.00 & 0.00 & 0.00 & 0.00 & 0.00 & 0.00 & 0.00 & \cellcolor{gray!20}1.00 \\
\hline
\end{tabular}
\end{adjustbox}
\end{table*}

\begin{table*}[htbp]
\centering
\caption{Cross-world dynamics of character status variables (mean value) in \textbf{\texttt{World Big}}.}
\label{tab:cross_world_dynamics_big}
\begin{adjustbox}{width=0.6\textwidth}
\begin{tabular}{|l|c|c|c|c|c|}
\hline
\textbf{World} &  \textbf{health} & \textbf{hunger} & \textbf{morale} & \textbf{stamina} & \textbf{thirst} \\
\hline
worldbig1 & 73.3 & 3.7 & 80.2 & 93.9 & 5.0 \\
worldbig2 &  73.7 & 4.1 & 80.4 & 93.1 & 4.7 \\
worldbig3 &  75.4 & 3.8 & 81.6 & 93.4 & 5.2 \\
worldbig4 &  75.1 & 3.7 & 80.2 & 92.9 & 5.2 \\
worldbig5 &  74.7 & 3.6 & 80.9 & 93.3 & 5.0 \\
worldbig6 &  75.2 & 4.0 & 80.4 & 92.9 & 5.4 \\
worldbig7 &  73.6 & 3.8 & 82.4 & 93.3 & 5.1 \\
worldbig8 &  71.0 & 4.0 & 81.1 & 92.6 & 5.3 \\
worldbig9 &  75.3 & 3.9 & 81.1 & 93.0 & 5.0 \\
\hline
\end{tabular}
\end{adjustbox}
\end{table*}

\begin{table*}[htbp]
\centering
\caption{Status $\times$ activity correlations (n=1152). Each cell shows Pearson $r$ / Spearman $\rho$ in \textbf{\texttt{World Big}}.}
\label{tab:status_activity_corr_big}
\begin{adjustbox}{width=\textwidth}
\begin{tabular}{|l|c|c|c|c|c|c|c|c|c|c|c|}
\hline
\textbf{Status} & \textbf{Actions\_Total} & \textbf{Emotions\_Total} & \textbf{Goals\_Total} & \textbf{AI\_States\_Total} & \textbf{Distinct\_Emotions} & \textbf{Distinct\_Goals} & \textbf{Distinct\_AI\_States} & \textbf{Emotion\_Transitions} & \textbf{Emotion\_Entropy} & \textbf{Goal\_Entropy} & \textbf{AI\_State\_Entropy} \\
\hline
\texttt{Mean\_Energy} & +0.00/+0.00 & +0.00/+0.00 & +0.00/+0.00 & +0.00/+0.00 & +0.00/+0.00 & +0.00/+0.00 & +0.00/+0.00 & +0.00/+0.00 & +0.00/+0.00 & +0.00/+0.00 & +0.00/+0.00 \\
\texttt{Mean\_Health} & +0.02/+0.01 & +0.00/+0.00 & -0.00/-0.01 & +0.00/+0.00 & +0.00/+0.00 & -0.01/-0.02 & \cellcolor{red!20}-0.68/-0.69 & \cellcolor{red!20}+0.73/+0.68 & \cellcolor{yellow!20}-0.26/-0.42 & -0.01/-0.02 & \cellcolor{red!20}-0.67/-0.69 \\
\texttt{Mean\_Hunger} & +0.02/+0.02 & +0.00/+0.00 & -0.01/+0.01 & +0.00/+0.00 & +0.00/+0.00 & +0.01/+0.02 & \cellcolor{red!20}+0.50/+0.52 & \cellcolor{yellow!20}-0.22/-0.27 & \cellcolor{orange!20}+0.30/+0.31 & +0.00/+0.02 & \cellcolor{red!20}+0.56/+0.59 \\
\texttt{Mean\_Loyalty} & +0.00/+0.00 & +0.00/+0.00 & +0.00/+0.00 & +0.00/+0.00 & +0.00/+0.00 & +0.00/+0.00 & +0.00/+0.00 & +0.00/+0.00 & +0.00/+0.00 & +0.00/+0.00 & +0.00/+0.00 \\
\texttt{Mean\_Morale} & -0.03/-0.07 & +0.00/+0.00 & +0.04/+0.04 & +0.00/+0.00 & +0.00/+0.00 & +0.03/+0.04 & \cellcolor{yellow!20}-0.16/-0.16 & +0.04/+0.05 & -0.06/-0.07 & +0.04/+0.04 & \cellcolor{yellow!20}-0.18/-0.17 \\
\texttt{Mean\_Stamina} & -0.04/-0.04 & +0.00/+0.00 & +0.05/+0.04 & +0.00/+0.00 & +0.00/+0.00 & +0.03/+0.02 & \cellcolor{red!20}-0.55/-0.55 & \cellcolor{yellow!20}+0.22/+0.29 & \cellcolor{orange!20}-0.40/-0.40 & +0.04/+0.02 & \cellcolor{red!20}-0.65/-0.63 \\
\texttt{Mean\_Thirst} & +0.00/-0.00 & +0.00/+0.00 & -0.05/-0.04 & +0.00/+0.00 & +0.00/+0.00 & -0.04/-0.03 & \cellcolor{orange!20}+0.40/+0.41 & \cellcolor{yellow!20}-0.19/-0.17 & \cellcolor{yellow!20}+0.15/+0.19 & -0.05/-0.03 & \cellcolor{orange!20}+0.45/+0.46 \\
\texttt{Mean\_Trust} & +0.00/+0.00 & +0.00/+0.00 & +0.00/+0.00 & +0.00/+0.00 & +0.00/+0.00 & +0.00/+0.00 & +0.00/+0.00 & +0.00/+0.00 & +0.00/+0.00 & +0.00/+0.00 & +0.00/+0.00 \\
\hline
\end{tabular}
\end{adjustbox}
\end{table*}

\begin{table*}[htbp]
\centering
\caption{Pearson correlation matrix between per-character metrics in \textbf{\texttt{World Big}}.}
\label{tab:correlation_matrix_big}
\begin{adjustbox}{width=\textwidth}
\begin{tabular}{|l|c|c|c|c|c|c|c|c|c|c|c|c|c|c|c|c|}
\hline
\textbf{Variable} & \textbf{Mean\_Health} & \textbf{Mean\_Stamina} & \textbf{Mean\_Morale} & \textbf{Actions\_Total} & \textbf{Distinct\_Emotions} & \textbf{Distinct\_Goals} & \textbf{Distinct\_AI\_States} & \textbf{Emotion\_Transitions} & \textbf{Emotion\_Entropy} & \textbf{Goal\_Entropy} & \textbf{AI\_State\_Entropy} & \textbf{Mean\_Energy} & \textbf{Mean\_Hunger} & \textbf{Mean\_Loyalty} & \textbf{Mean\_Thirst} & \textbf{Mean\_Trust} \\
\hline
\texttt{Mean\_Health} & \cellcolor{gray!20}1.00 & \cellcolor{blue!10}0.36 & 0.11 & 0.02 & 0.00 & -0.01 & \cellcolor{red!20}-0.68 & \cellcolor{blue!20}0.73 & \cellcolor{red!10}-0.26 & -0.01 & \cellcolor{red!20}-0.67 & 0.00 & \cellcolor{red!10}-0.38 & 0.00 & \cellcolor{red!10}-0.26 & 0.00 \\
\texttt{Mean\_Stamina} & \cellcolor{blue!10}0.36 & \cellcolor{gray!20}1.00 & 0.15 & -0.04 & 0.00 & 0.03 & \cellcolor{red!20}-0.55 & \cellcolor{blue!10}0.22 & \cellcolor{red!10}-0.40 & 0.04 & \cellcolor{red!20}-0.65 & 0.00 & \cellcolor{red!20}-0.51 & 0.00 & \cellcolor{red!10}-0.35 & 0.00 \\
\texttt{Mean\_Morale} & 0.11 & 0.15 & \cellcolor{gray!20}1.00 & -0.03 & 0.00 & 0.03 & -0.16 & 0.04 & -0.06 & 0.04 & -0.18 & 0.00 & -0.16 & 0.00 & -0.12 & 0.00 \\
\texttt{Actions\_Total} & 0.02 & -0.04 & -0.03 & \cellcolor{gray!20}1.00 & 0.00 & -0.07 & -0.01 & 0.02 & 0.01 & -0.07 & -0.01 & 0.00 & 0.02 & 0.00 & 0.00 & 0.00 \\
\texttt{Distinct\_Emotions} & 0.00 & 0.00 & 0.00 & 0.00 & \cellcolor{gray!20}1.00 & 0.00 & 0.00 & 0.00 & 0.00 & 0.00 & 0.00 & 0.00 & 0.00 & 0.00 & 0.00 & 0.00 \\
\texttt{Distinct\_Goals} & -0.01 & 0.03 & 0.03 & -0.07 & 0.00 & \cellcolor{gray!20}1.00 & -0.01 & 0.02 & 0.04 & \cellcolor{blue!20}0.98 & -0.01 & 0.00 & 0.01 & 0.00 & -0.04 & 0.00 \\
\texttt{Distinct\_AI\_States} & \cellcolor{red!20}-0.68 & \cellcolor{red!20}-0.55 & -0.16 & -0.01 & 0.00 & -0.01 & \cellcolor{gray!20}1.00 & \cellcolor{red!10}-0.49 & \cellcolor{blue!10}0.30 & -0.02 & \cellcolor{blue!20}0.93 & 0.00 & \cellcolor{blue!20}0.50 & 0.00 & \cellcolor{blue!10}0.40 & 0.00 \\
\texttt{Emotion\_Transitions} & \cellcolor{blue!20}0.73 & \cellcolor{blue!10}0.22 & 0.04 & 0.02 & 0.00 & 0.02 & \cellcolor{red!10}-0.49 & \cellcolor{gray!20}1.00 & \cellcolor{blue!10}0.26 & 0.02 & \cellcolor{red!10}-0.48 & 0.00 & \cellcolor{red!10}-0.22 & 0.00 & -0.19 & 0.00 \\
\texttt{Emotion\_Entropy} & \cellcolor{red!10}-0.26 & \cellcolor{red!10}-0.40 & -0.06 & 0.01 & 0.00 & 0.04 & \cellcolor{blue!10}0.30 & \cellcolor{blue!10}0.26 & \cellcolor{gray!20}1.00 & 0.03 & \cellcolor{blue!10}0.34 & 0.00 & \cellcolor{blue!10}0.30 & 0.00 & 0.15 & 0.00 \\
\texttt{Goal\_Entropy} & -0.01 & 0.04 & 0.04 & -0.07 & 0.00 & \cellcolor{blue!20}0.98 & -0.02 & 0.02 & 0.03 & \cellcolor{gray!20}1.00 & -0.02 & 0.00 & 0.00 & 0.00 & -0.05 & 0.00 \\
\texttt{AI\_State\_Entropy} & \cellcolor{red!20}-0.67 & \cellcolor{red!20}-0.65 & -0.18 & -0.01 & 0.00 & -0.01 & \cellcolor{blue!20}0.93 & \cellcolor{red!10}-0.48 & \cellcolor{blue!10}0.34 & -0.02 & \cellcolor{gray!20}1.00 & 0.00 & \cellcolor{blue!20}0.56 & 0.00 & \cellcolor{blue!10}0.45 & 0.00 \\
\texttt{Mean\_Energy} & 0.00 & 0.00 & 0.00 & 0.00 & 0.00 & 0.00 & 0.00 & 0.00 & 0.00 & 0.00 & 0.00 & \cellcolor{gray!20}1.00 & 0.00 & 0.00 & 0.00 & 0.00 \\
\texttt{Mean\_Hunger} & \cellcolor{red!10}-0.38 & \cellcolor{red!20}-0.51 & -0.16 & 0.02 & 0.00 & 0.01 & \cellcolor{blue!20}0.50 & \cellcolor{red!10}-0.22 & \cellcolor{blue!10}0.30 & 0.00 & \cellcolor{blue!20}0.56 & 0.00 & \cellcolor{gray!20}1.00 & 0.00 & \cellcolor{blue!10}0.30 & 0.00 \\
\texttt{Mean\_Loyalty} & 0.00 & 0.00 & 0.00 & 0.00 & 0.00 & 0.00 & 0.00 & 0.00 & 0.00 & 0.00 & 0.00 & 0.00 & 0.00 & \cellcolor{gray!20}1.00 & 0.00 & 0.00 \\
\texttt{Mean\_Thirst} & \cellcolor{red!10}-0.26 & \cellcolor{red!10}-0.35 & -0.12 & 0.00 & 0.00 & -0.04 & \cellcolor{blue!10}0.40 & -0.19 & 0.15 & -0.05 & \cellcolor{blue!10}0.45 & 0.00 & \cellcolor{blue!10}0.30 & 0.00 & \cellcolor{gray!20}1.00 & 0.00 \\
\texttt{Mean\_Trust} & 0.00 & 0.00 & 0.00 & 0.00 & 0.00 & 0.00 & 0.00 & 0.00 & 0.00 & 0.00 & 0.00 & 0.00 & 0.00 & 0.00 & 0.00 & \cellcolor{gray!20}1.00 \\
\hline
\end{tabular}
\end{adjustbox}
\end{table*}

\section{Discussion}

In the current implementation of \varalias' Simulated World Module, we have implemented MUD functionality through text interaction, an integration that could be extended to multimodal interaction (e.g. through 2D/3D audiovisual virtual experiences). To integrate \varalias with real-time applications (videogames and virtual experiences) would consist on developing an API/Plugin of the aforementioned socio-affective factor and behavior computation classes and functionality in C++ (for Unreal Engine), C\# (for Godot and Unity).

The tests perfomed in Sections \ref{sec:results_reasoning} to \ref{sec:results_simulation} assess model inference on human ToM reasoning and emotion terms as a proof of concept in autonomous MAS and LLM-based agents. Further extension for \varalias' ToM include strategic interaction games such as the Prisoner's Dilemma, Nash's Equilibrium, Cooperation vs Competition strategies and other game theory test scenarios \cite{Busoniu2008, Wellman2025, Sun2025, Trencsenyi2025, wang2026deep, Hao2026}, also including social factor simulations such as social homophily and gender dynamics \cite{Frolova2002,Fadaei2026}. All these theories can be implemented as \varalias templates with conditions/rules in \texttt{json} graphs. 

 The \varalias framework contributes to a rapidly expanding body of literature focused on the embodiment of Large Language Models within persistent, multi-agent virtual spaces. The fundamental necessity for breaking away from monolithic, hardcoded architectures is highlighted by recent foundational works like \textit{Generative Agents} \cite{Park2023}, which successfully demonstrated that integrating reflective memory streams allows LLMs to simulate believable, autonomous human behavior. However, while this focused heavily on offline, agent-to-agent socialization computed over simulated days, \varalias extends this paradigm by prioritizing real-time, multimodal \textit{human-in-the-loop} interaction by design. Latter integrations of complex biometric, spatial, and auditive and visual triggers would ensure that the agents react not only to isolated textual prompts, but to embodied, physical presence of the human within mixed-reality spaces \cite{Oviatt1999}.

Similarly, Wang et al.'s \textit{Voyager} \cite{wang2023voyager} showcases the impressive capacity for embodied agents to perform lifelong learning and procedural skill acquisition within complex 3D environments like Minecraft. While Voyager excels at autonomous task completion through dynamic code generation, it inherently lacks structural frameworks for social governance, interpersonal negotiation, and multi-actor fairness. \varalias directly addresses this critical gap in the literature. By explicitly embedding formal game theory mechanisms—such as Shapley value evaluations and Ostrom's design principles—into the rigid \textit{System Rules} and \textit{Rule Interpreters}, \varalias ensures that as these virtual ecosystems scale in population, they do not merely act efficiently, but negotiate, cooperate, and govern themselves ethically. 

Furthermore, while modern frameworks like \textit{AutoGen} \cite{wu2024autogen} have successfully standardized LLM multi-agent conversation topologies, they are primarily text-based and lack the physical network synchronization techniques necessary for integration into real-time game engines. \varalias bridges this engineering gap by explicitly defining the network layers (Centralized vs. P2P semantic delta transmission) necessary to keep distributed spatial simulation states coherent across multiple machines \cite{Liu2022}. The adoption of MCP services allows multi-tool usage and further standardizes how these reasoning agents interface with distinct physical engines, vastly reducing vendor lock-in and development friction. Future iterations of the \varalias framework will able experimenting with dynamic offloading of cognitive tasks by shifting processing workloads from local quantized language models to centralized macroscopic reasoning models on the fly, optimizing for network bandwidth, hardware capability, and the complexity of the required social negotiation.

Through a set of world \texttt{json} templates (i.e. \texttt{world\_} \texttt{vocabulary.json}, \texttt{generation\_}\texttt{config.json}, \texttt{world\_}\texttt{dynamics.json}) and entity names (i.e. \texttt{objects\_}\texttt{names.txt}, \texttt{characters\_}\texttt{names.txt} and \texttt{scenes\_}\texttt{names.txt}), \varalias can generate any type of World Simulation under explicit rules, scalable with Multi-User text interaction. This implies \varalias' SWM can explicitly define entities, rules and conditions, to specific objects and scenes, enabling behavioral tests through a socio-affective computation regime while computing complex LLM reasoning models. These tests may enable evaluation of evolutionary/swarm intelligence through large simulations of micro- and macro-economic societies by parameterizing system and organizational rules, object resources (e.g. scarce resources in populated worlds vs abundant resources in small populations), historical entities as well as fictional worlds.

\subsection*{Author Contributions} 
David Berga contributed to the research, data analysis, experimentation, writing and reviewing of this article.

\subsection*{Conflicts of Interest}
The authors declare that they have no known competing financial interests or personal relationships that could have appeared to influence the work reported in this article.

\section{Supplementary Material} \label{sec:materials}

\begin{itemize}
    \item AGIMUD Software: \href{https://github.com/dberga/AGIMUD}{(https://github.com/dberga/AGIMUD)}
    \item LLM Module (using LM Studio Python SDK): \href{https://github.com/dberga/lmstudio-python-wrapper}{(https://github.com/dberga/lmstudio-python-wrapper)}
\end{itemize}

\bibliographystyle{IEEEtran}
\bibliography{library}

\end{document}